\documentclass{article}

\usepackage[main, final]{neurips_2026}
\usepackage[utf8]{inputenc} % allow utf-8 input
\usepackage[T1]{fontenc}    % use 8-bit T1 fonts
\usepackage{hyperref}       % hyperlinks
\usepackage{url}            % simple URL typesetting
\usepackage{booktabs}       % professional-quality tables
\usepackage{amsfonts}       % blackboard math symbols
\usepackage{nicefrac}       % compact symbols for 1/2, etc.
\usepackage{microtype}      % microtypography
\usepackage{xcolor}         % colors
\usepackage{stmaryrd}
\usepackage{pdfpages}
\usepackage[most]{tcolorbox}
\newtcolorbox{mybox}[1][]{
    colframe=black!70,   % couleur du cadre
    colback=black!3,           % couleur de fond
    arc=2mm,                  % coins arrondis
    boxrule=0.2mm,              % épaisseur du cadre
    left=1.1mm, right=1.1mm,
    top=0mm, bottom=0mm,
    enhanced,
    #1                        % permet d'ajouter des options supplémentaires à l'appel
}

\usepackage{algorithm}
\usepackage{algpseudocode}

\usepackage{amsmath}
\usepackage{amssymb}
\usepackage{mathtools}
\usepackage{amsthm}

\usepackage{float}
\usepackage{multirow}
\usepackage{subcaption}
\usepackage{makecell}
\usepackage{xspace}

\usepackage[dvipsnames]{xcolor}
\newcommand{\ABCPearson}{\textbf{\color{red}ABC-Pearson}\xspace}
\newcommand{\ABCCNN}{\textbf{\color{blue}ABC-CNN}\xspace}
\definecolor{gold}{RGB}{218,165,32}
\newcommand{\ABCTransf}{\textbf{\color{gold}ABC-Transf}\xspace}
\newcommand{\NPECNN}{\textbf{\color{ForestGreen}NPE-CNN}\xspace}
\definecolor{purple}{RGB}{147,112,219}
\newcommand{\NPETransf}{\textbf{\color{purple}NPE-Transf}\xspace}

\definecolor{C0}{HTML}{3182ce}
\definecolor{C1}{HTML}{1F4E79}
\hypersetup{colorlinks,citecolor=C1,linkcolor=C0,urlcolor=C1}

\title{\textit{BlockFormer}:
Transformer-based inference \\
from genomic contact maps} %interaction maps} %with spot-like structures}

\author{%
  Eloïse Touron \\
        Univ. Grenoble Alpes, Inria \\
CNRS, Grenoble INP, LJK, France \\
        \texttt{eloise.touron@inria.fr} \\
      \And Pedro L. C. Rodrigues\\
        Univ. Grenoble Alpes, Inria \\
CNRS, Grenoble INP, LJK, France \\
        \texttt{pedro.rodrigues@inria.fr} \\
    \And Julyan Arbel\\
        Univ. Grenoble Alpes, Inria \\
CNRS, Grenoble INP, LJK, France \\
        \texttt{julyan.arbel@inria.fr} \\
    \And Nelle Varoquaux\\
    TIMC, Univ. Grenoble Alpes\\
    CNRS, Grenoble INP, France \\
    \texttt{nelle.varoquaux@univ-grenoble-alpes.fr}
    \And Michael Arbel\\
        Univ. Grenoble Alpes, Inria \\
CNRS, Grenoble INP, LJK, France \\
        \texttt{michael.arbel@inria.fr} \\
}

\begin{document}

\maketitle

\begin{abstract}

Locating genomic features from genomic contact maps, such as centromere identification from genome-wide chromosome conformation capture techniques, notably Hi-C, can be formulated as an inverse problem: infer one parameter per entity given a map summarizing pairwise interactions through blocks of variable numbers and sizes. In this work, we introduce a data-driven approach that leverages shared structure between these contact maps, such as global alignment between localized patterns, while handling the variability in number and size of chromosomes arising in real-world data. Our approach relies on a transformer architecture capable of handling such variability and a custom simulator to generate abundant, yet computationally cheap synthetic data for training. Applied to the problem of centromere localization, the method recovers genomic positions at or below the map resolution across species with various genome sizes using a coarse-to-fine refinement step when chromosome sizes fall outside the training range. We concentrate our evaluation on Hi-C data, where the block structure is well characterized and ground truth is available.
\end{abstract}

\section{Introduction}
%\note{ajouter notion: pas de méthode existante + alignement de l'info sur plusieurs blocks pour meilleure estimation + presenter défis de la tache: nb et taille bloc variable, mais toujours meme structure  -> une seule methode flexible à tous + dependence on resolution seq depth, genome size, real examples}\\
Interaction maps summarize pairwise relationships between 
%arising from the interaction of several 
entities within a system. Inherently encoding the underlying system's structure, these maps can support inference of entity-level properties. 
%they enable inference of certain properties arising from the system.  
%They constitute a fundamental data structure arising from some experimental assays and from which properties about whole or parts of the system are to be inferred.
For example, protein-protein interaction maps are used to identify key regulatory proteins at the origin of abnormal gene expression \cite{protein}, modularity in species-species interaction maps help identifying critical species for ecosystem stability %, as losing one of them can lead to extinction cascades 
 \cite{ecology}. 
%functional Magnetic Resonance Imaging (fMRI) connectivity matrices are used to identify hub regions in the brain \cite{brain}; 
%and gene co-expression networks are employed to quantify module hubness per gene \cite{gene}.
The goal is then to solve the following inverse problem : infer per-entity parameters from a given interaction map.

In many such problems, these maps exhibit a block-wise structure, where each block represents interactions between two entities, with block number and size varying across maps. 
The maps can exhibit both localized patterns within each block and non-local information across multiple blocks, for instance, in the form of an alignment between per-block patterns. 
Several biological applications fit this framework, including Hi-C maps, where the interacting entities are chromosomes.
%Relevant biological applications follow this inference framework, with Hi-C maps as interaction maps and chromosomes as interacting entities. 
Hi-C maps summarize physical contact counts between genomic loci across a population of cells into a block-wise matrix %square, symmetric matrix 
and have become a central tool for studying DNA folding and some associated genetic diseases, notably through the identification of chromatin loops and topologically associated domains (TADs) \cite{tads, loops}. Beyond them, centromeres are key elements of genome organization due to their essential role in chromosome segregation and genome stability~\cite{kinetochore}. 
%In yeasts, centromeres are compact regions of approximately $125$ base pairs (bp) \cite{centrosize} that colocalize in the nucleus, producing characteristic peaks in Hi-C contact maps. 
While they have traditionally been annotated experimentally~\cite{FISH,CHIP}, these approaches can be imprecise or fail for some species~\cite{inferfail}. Methods such as \textit{Centurion} instead infer centromere positions directly from Hi-C data by fitting Gaussian profiles to interaction peaks~\cite{nelle}. This procedure, however, is non-amortized and computationally costly because it requires solving a non-convex optimization problem (see Appendix~\ref{sec:centurion}).
%Centromere localization, for instance, have  traditionally been annotated experimentally from Hi-C maps~\cite{FISH,CHIP}. These approaches can be imprecise or fail for some species \cite{inferfail}. 
%Other methods such as \textit{Centurion}, instead infer centromere positions directly from Hi-C data \cite{nelle} by fitting Gaussian profiles to interaction peaks, but this approach is non-amortized and time-consuming due to the non-convex nature of the optimization (see Appendix~\ref{sec:centurion}). 
%it relies on accurate pre-localization and is time-consuming (see Appendix~\ref{sec:centurion}). 

As a large number of Hi-C maps became recently available, there is a clear interest in leveraging learning-based approaches to automatize the inference of properties from a given interaction map, like centromeres location. This raises methodological challenges such as how to effectively design and learn models capable of handling maps with various block numbers and shapes while capturing consistent structural pattern across blocks.
Off-the shelf techniques, such as supervised deep learning, could be applied, but they would require manually annotating data, which is costly.
Bayesian inference approaches were proposed to estimate some DNA properties such as chromatin compaction and persistence length \cite{arbona}, but they require defining a tractable likelihood model of these interaction maps, which might be challenging to do, given the complexity and rich structure of these interactions.

Alternatively, leveraging simulated data appears as a promising way to bypass those limitations
%the need for annotated data or simplified tractable likelihood models 
when learning an inference model. 
%Such approaches have yielded promising results with approaches such as simulation-based inference (SBI) or prior-fitted networks (TabPFN).  
Simulation-based inference (SBI) was particularly useful in physics applications \cite{sbi_cosmo, sbi_cosmo_1} due to realistic simulators \cite{cosmo_simu, cosmo_simu_1}. Prior-fitted networks have shown excellent results on real world tabular data \cite{tabfn} while training is still performed on purely synthetic ones. In both settings, the ability to handle data of various sizes is still an active research topic.

\begin{figure*}[t]
  \centering
  \includegraphics[width=\linewidth]{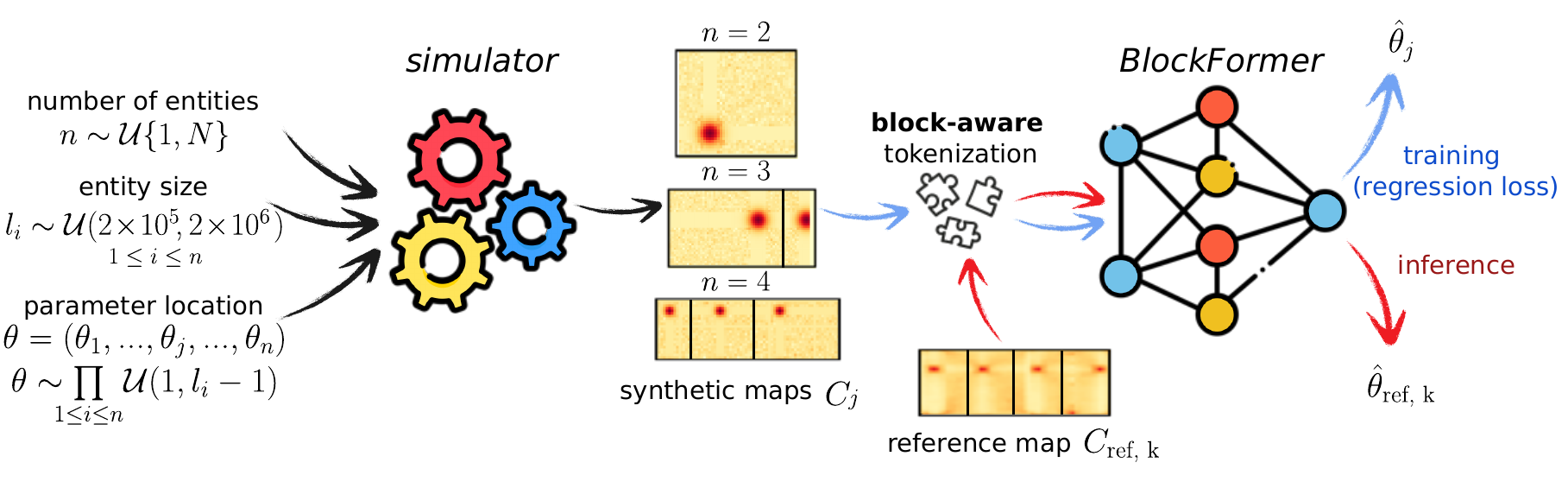}
  \caption{\small Inference from interaction maps using \textit{BlockFormer} (see architecture in Figure~\ref{fig:transformer}). The input $C_k$ is any sequence of blocks of interactions between entity $k$ and others, the output $\hat{\theta}_k$ is the parameter estimation. $N$ denotes the maximum number of entities considered.\label{fig:general_process}}
  % \label{fig:transformer}
\end{figure*}

In this work\footnote{This submission is based on an earlier version 
\cite{etouron}
%(included in Appendix~\ref{sec:workshop}) 
presented at a workshop without official proceedings. }, we propose \textit{BlockFormer}, a transformer-based model to infer per-entity properties from genomic contact maps, whose block-wise structure varies in both the number and the size of blocks. %with variable block-wise structure.
%train a transformer-based model, the \textit{BlockFormer}, capable of inferring 
Our architecture employs a three dimensional positional encoding that allows handling variable block sizes and numbers while capturing per-block patterns and aggregating  non-local information across multiple blocks. 
%In this work\footnote{This submission is based on an earlier version, included in Appendix~\ref{sec:workshop}, presented at a workshop without official proceedings.}, we propose a learning-based framework for inferring per-entity properties from block-structured interaction maps of variable variable block number and size. We train \textit{BlockFormer}, a transformer-based model, on simulated data to accommodate variable block numbers and sizes while capturing within-block patterns and aggregating non-local information across blocks.
%a transformer-based architecture for interaction maps that naturally handles  
We designed a simple simulator that reproduces patterns necessary for inference, enabling fast generation of synthetic contact maps. 
Pre-trained on synthetic data of variable structures generated via this simulator, \textit{BlockFormer} performs fast and accurate 
centromere localization from Hi-C maps of species with widely varying genome sizes 
%spanning yeasts, a parasite and a plant 
(see Figure~\ref{fig:general_process}). Without retraining, it also extends to chromatin loop localization on pre-filtered single-loop regions.

\section{Related work}

\begin{figure*}[t]
  \centering
  \includegraphics[width=\linewidth]{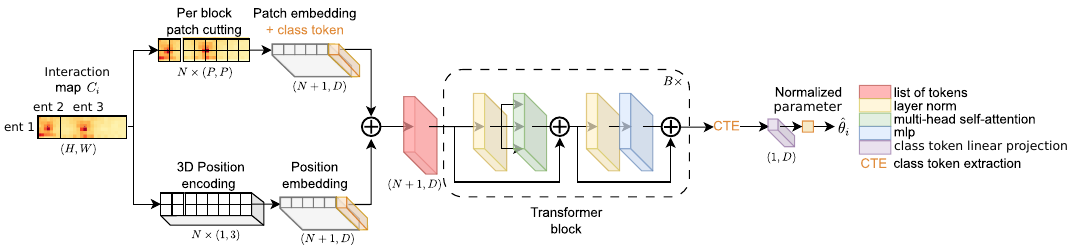}
  \caption{\small Architecture of \textit{BlockFormer}. The input $C_i$ is any sequence of blocks of interactions between entity $i$ and others (size of the map $(H,W)$) and the output is the parameter estimation $\hat{\theta}_i$. $N$ is the number of patches, $P$ the patch size, $D$ the embedding dimension and $B$ the number of transformer blocks.}
  \label{fig:transformer}
\end{figure*}

\textbf{Deep learning approaches in structural biology.} 
Prior works have leveraged deep learning methods to model biological structure and function at different scales. \cite{akita} and \cite{enformer} proposed CNN- and attention-based architectures to predict chromatin folding or gene expression directly from DNA sequence. However, these methods are not directly applicable to our setting, as they operate on single blocks and do not model interactions across multiple blocks.\\
In the context of chromatin structure, Hi-C–based graph approaches such as \cite{hicgnn, hicoex} model contact maps as graphs and apply graph neural networks (GNNs) or graph attention networks (GATs) to reconstruct 3D genome organization or infer functional relationships. These methods, however, typically operate on localized regions or treat the genome as a single homogeneous graph, relying on global message passing without explicitly capturing block-wise organization.
Transformer-based architectures have also been successfully applied in structural biology, most notably in \textit{Alphafold} \cite{alphafold}, which predicts the 3D structure of individual proteins from amino acid sequences using evolutionary constraints such as multiple sequence alignments (MSAs). However, such approaches are designed for single-molecule folding and do not extend to genome-scale chromatin organization.\\
In contrast, our method operates directly on genome-wide Hi-C contact maps and explicitly models their block-wise structure. By leveraging a block-aware transformer-based architecture, it enables the identification of large-scale structural features such as centromeres that are not addressed by prior sequence-based or graph-based methods. %\note{verifier si scale à nos matrices (tailles), MSA ? plutot dire que difficile à appliquer dans notre cas} Designed to handle inputs of varying sizes, its row- and column-wise attention mechanism motivates us to operate on submatrices of the interaction map, thereby decomposing the inference task into several simpler subproblems.
%\note{ajouter hic + transformer}
%Recently, transformer-based architectures have replaced convolutional neural networks (CNNs) in visual recognition tasks, such as classification or object detection. 
Inspired by Vision Transformers (ViT) \cite{transf_image} which tokenize input into fixed-length patches to remove the image size constraint, we adopt a token-based mechanism, with key adaptations to block-wise structures %to patch construction and positional encoding tailored to block-wise genome-wide contact map structure. The classification token proposed in ViT is repurposed in our model to encode the parameters of interest for the inference 
(see Section~\ref{sec:transf_archi} and Figure~\ref{fig:transformer}).

\textbf{Bio-physical simulators of DNA.} In the biological context, highly complex simulators are typically used \cite{simu_bio, simu_folding}. Based on molecular dynamics, they model DNA fragments as polymers or chains of beads and attempt to mimic the chromosome folding in the cell by solving biophysical equations. As many variables are involved and must satisfy a set of constraints, such simulators are extremely slow to produce only a single folding configuration. However, training our model requires lots of contact maps that are summary statistics over a population of chromosome folding, not just one. Consequently, using a biological simulator to construct thousands of contact maps is computationally impractical within a reasonable runtime. Moreover, Hi-C structure already encodes sufficient information to localize centromeres, making explicit reconstruction of full 3D chromatin folding unnecessary. We therefore build a simplified and lightweight contact map simulator that directly generates the map $C$ from the centromere positions $\theta$ without simulating any DNA folding (see Section~\ref{sec:gwcontact}).  
%\note{mettre plus de citations, couteux et on propose un simulateur lightweight}

\section{\textit{BlockFormer}: transformer-based inference from interaction maps}
%To address the inference problem, we propose a two-step approach. First, we obtain a point estimate $\hat{\theta}$ via the transformer-based architecture. We then adopt a stochastic inference scheme to approximate $p(\theta | C_{\text{ref}})$, enabling principled quantification of uncertainty in the parameter estimates. This approach is novel in this framework, and we will compare several Bayesian inference methods.
%\subsection{Transformer-based architecture for point estimates}\label{sec:archi}
%The first step is to give a point estimate of the parameter $\theta$. 
The goal is to give a point estimate of the per-entity parameter $\theta$ (in $\mathbb{R}^n$ for $n$ entities) given any interaction map $C$ (in $\mathbb{R}^{k \times k}$ where $k = \sum_{1 \leq i \leq n} l_i$, $l_i$ size of entity $i$) that shows an enrichment of interactions at the location of the parameters.
Two entities $(i, j)$ of size $l_i$ and $l_j$ represent a block of interaction in $C$ of size $(l_i, l_j)$. Depending on the studied system, both the number and the size of interacting entities vary, leading to interaction blocks with heterogeneous sizes and numbers. Consequently, a model applicable to any set of entities should generalize to any number and dimension of blocks. Transformers provide a natural architecture to handle these challenges: by mapping any input image into a sequence of tokens, they eliminate the image size constraint. However, in contrast to standard settings, our problem involves block-wise maps, for which no existing architectures are directly designed. In particular, we introduce a block-aware transformer-based architecture tailored to interaction maps: \textit{BlockFormer}. Along with a proper training strategy, our architecture handles contact maps across varying resolutions, sequencing depths and genome sizes, as shown in Sections~\ref{sec:appli} and \ref{sec:ablations}.
%is flexible to a wide range of interaction maps.\\
We provide comparisons of \textit{BlockFormer} with other architectures as well as ablation studies of per-block modeling and the training strategy in Appendix~\ref{sec:ablations_appendix}.

\subsection{A block-aware transformer-based architecture }\label{sec:transf_archi}
Inferring the full parameter $\theta$ from an interaction map $C$ can be challenging because the dimensionality of $\theta$ varies with the number of entities. Assuming $n$ entities, we therefore decompose the inference problem into $n$ subproblems, where the goal is to infer the $i^{\text{th}}$ component $\theta_i$ from a corresponding sub-map $C_i$, that describes interactions of entity $i$ with others (see Appendix~\ref{sec:indep} for decomposition justification). This reformulation reduces the task to learning a shared architecture \textit{BlockFormer} that maps $C_i$ to $\theta_i$. The same model can then be applied independently across entities, enabling parameter inference for arbitrary numbers of entities (see Figure~\ref{fig:transformer} for architecture details).
%Since the dimensionality of the parameter $\theta$ depends on the number of entities, we reformulate the problem as a collection of one-dimensional inference tasks. For each entity $i$, the network takes as input an interaction map $C_i$ describing its interactions with all other entities.
%: $C_i$ is a set of trans-blocks where each block shows an enrichment of interaction at the position $(\theta_i, \theta_k)$ with $k$ the number of other interacting entities. 
%The model then infers the $i^{\text{th}}$ component of $\theta$, denoted $\theta_i$ (see Figure~\ref{fig:transformer} for architecture details).

\textbf{Per-block patching.}
The input interaction map $C_i$ is first cut into squared patches of size $(P,P)$. The partitioning process must respect the block-wise structure of the map: to avoid patches overlapping multiple blocks, we first apply an asymmetric zero-padding on right and bottom of each block of $C_i$ to a multiple of the patch size. After padding, the map is of size $(H_p, W_p)$ resulting in $N = H_pW_p/P^2$ patches. We define a constant latent vector size $D$ used across all transformer layers and referred to as the embedding dimension. All the patches are then projected to $D$ dimensions, resulting in the patch embeddings of size $(N, D)$. We prepend a learnable embedding to the previous sequence (a class token of size $(1,D)$) that will encode the estimate of the parameter $\theta_i$.

\textbf{Per-block positional encoding.}
To retain the position of each patch in the original map, position embeddings of size $(N+1, D)$ are added to the patch embeddings. To enable the network to handle varying numbers of blocks and thus varying number of interacting entities, this information must be incorporated into the positional encoding. For each patch, we define a 3D-position vector $(i,j,k)$ where $i$ is the block index, and $(j,k)$ the relative position of the patch in the block $i$. This vector is then projected to $D$ dimensions via a fixed per-coordinate sine-cosine positional encoding. 
The resulting sequence of embedding vectors (tokens) serves as input to the transformer. 
The transformer adopts the classical series of $B$ blocks alternating Multi-head self-attention, Layernorm, and MLP presented in vision transformer (ViT) \cite{transf_image}.

\textbf{Class token projection.}
The target parameter $\theta_i$ often represents relative position within each entity and is therefore a non-negative real scalar. Consequently, at the final layer output, the class token is extracted and projected to a real scalar using a linear layer followed by a sigmoid activation, ensuring an output between $0$ and $1$. This normalized estimate is then rescaled to a real entity position to recover the actual parameter $\theta_i$.
%We provide in Appendix~\ref{sec:archi_comparaison} a comparison with others learning-based approaches.
\subsection{Training strategy for flexibility to various size and number of blocks.}
For computational efficiency, we propose training the model on small interaction maps while allowing it to generalize to larger ones, that have more blocks (i.e. varying amounts of parameter information) or larger blocks (i.e. different map resolutions or entity sizes). 
Specifically, we rely on simulated training data consisting of map/parameter pairs $(C_i,\theta_i)$, indexed by their corresponding entity $i$. To enable the transformer to generalize to various numbers and sizes of blocks, we ensure that training maps are as broadly representative as possible by choosing, per batch, a number of blocks, the size of each block, and the entity to test. Each map is normalized between $0$ and $1$ to account for differences in their value scales, ensuring more consistent comparisons across maps. Since \textit{BlockFormer} ($\mathcal{BF}_\phi$) acts as a projection of $C_i$ onto $\theta_i$, the natural training loss is the regression one: \begin{equation}
\label{eq:loss}
\mathcal{L}(\phi)=\frac 1M \sum_{1\leq m\leq M}  \Vert \mathcal{BF}_\phi(C_i^m) - \tilde{\theta}_i^m \Vert ^2_ 2,\ \text{where $\tilde{\theta}_i$ is any normalized parameter.}
\end{equation}% where $C_i^n$ are simulated from $\theta_i^n$.
%We are now interested in the whole posterior estimation $p(\theta | C_{\text{ref}})$.
%We show in Appendix~\ref{sec:train_various_blocks_appendix} the efficiency of such strategy.

\section{Genomic application: inferring DNA-fragment locations from contact maps}\label{sec:appli}

We validate \textit{BlockFormer}’s performance through a biological application where entities are chromosomes interacting within the cell nucleus. The resulting interaction map $C$ is a contact map. 
%and our principal goal is to accurately infer centromere positions from it. 

%\subsection{Genome-wide contact maps}\label{sec:gwcontact}

\textbf{Generic contact maps structure.}\label{sec:gwcontact}
The genome-wide contact map $C$ projects the information contained in a population of 3D chromatin foldings into a 2D square and symmetric matrix made of cis- (or intra-chromosomal) and trans- (or inter-chromosomal) blocks of interactions between pairs of chromosomes. To construct it, we cut each chromosome into genomic windows of a given length (called resolution, e.g. $32$ kilobases (kb)) and each matrix entry is called the contact count, representing the number of times a given window was in contact with another one over the population (see Appendix \ref{sec:hic_appendix} and Figure \ref{fig:contactmap_schema}). 
In our setting, a cis-block mainly contains a diagonal of enrichment (i.e. high contact frequency region), whereas a trans-block between chromosomes $i$ and $j$ shows an enrichment of interactions at the location of both centromeres $(\theta_i, \theta_j)$. Thus, the main informative part about centromeres relies only on the trans-blocks.\\
To infer the centromere $\theta_i$, only the $i^{\text{th}}$ row of trans-blocks of $C$ (denoted $C_i$) is considered, reflecting chromosome $i$’s interactions with all the others. As the number of chromosomes can vary across species, from now on, $C_i$ will be any sequence of multiple trans-blocks.

\textbf{Hi-C map specificity.}
During inference, we use a real Hi-C map $C_{\text{ref}}$ and simulate synthetic contact maps $C$. Hi-C contact maps have many biases due to sequencing and mapping errors or to the inherent structure of the chromatin. Therefore, $C_{\text{ref}}$ is actually a normalized Hi-C map using ICE
%where the normalization corrects those biases, iteratively forcing all rows and columns to sum up to one
\cite{hicnorm} (see Appendix~\ref{sec:hic_norm_appendix}). 
Contact map quality depends on both resolution and sequencing depth. Greater sequencing depth increases the number of detected chromatin contacts, improving the signal-to-noise ratio and enabling analysis at finer resolutions. 
The resolution affects the precision of the parameter's inference. Indeed, a pixel in the contact map represents a fragment of DNA of length the resolution (e.g. $40$~kb) that is much larger than the centromere length (e.g. $100$ bp). 
%and centromere appear as few brighter pixels in each trans-block of the map. Very often, the resolution (e.g. $40$~kb) is much larger than the centromere length (e.g. $100$ bp). 
Achieving such precision is therefore challenging and a reasonable goal is to estimate centromeres with sub-resolution precision. Accordingly, the main pre-processing step is to select an appropriate resolution to generate smooth contact maps.
%Depending on the species, we will set the reference contact maps at a resolution between $20$~kb to $70$~kb. 

\textbf{Simulator.}
We exploit the structure of yeast contact maps to design a very efficient simulator that directly {creates the upper trans-blocks} given its centromere positions $\theta$. The simulated maps are simplified compared to real biological ones but still capture minimal sufficient structure (spot-like interactions) required for inference (see Appendix~\ref{sec:mismatch} for a misspecification analysis).
%\note{citations} 
%This way, we can have many simulated contact maps $C$ very rapidly. 
At the centromere positions, the chromatin has a brush-like organization: chromosomal regions near the centromeres often enter in contact over the population, whereas the further we move away from the centromeres, the rarer the contacts become. To mimic this effect, we simulate a Gaussian spot at the position ($\theta_i, \theta_j$) for each trans-contact block.
The depletion of contacts between centromeres and other loci is then simulated via a cross of non-interaction passing by ($\theta_i, \theta_j$).
Between chromosomes, we also observe rare interactions over the population that we reproduce by adding Gaussian noise to all the trans-blocks up to $10\%$ of the maximal contact count (see Appendix~\ref{sec:simulator_appendix}).
%with Algorithm~\ref{alg:simu} and Figure~\ref{fig:C_simu}).

\textbf{Training procedure. \label{sec:train_details}}
We aim to train \textit{BlockFormer} on small contact maps that can generalize to larger ones while limiting the training data budget. Accordingly, we adopt a small transformer with $B=4$ blocks, $4$ heads of attention, a patch size $P=4$ (smaller than the typical size of interaction blocks that can be $8$), and an embedding dimension of $D = 24$ (a multiple of the patch size). We design a training set of $50\ 000$ examples, grouped by batch of size $200$. Each synthetic map is a sequence of trans-blocks at resolution $32$~kb where the number and size of blocks vary. To avoid overfitting to any fixed simulator parameter, each simulated trans-block has a Gaussian spot that varies in size and location. Training data generation details are provided in Appendix~\ref{sec:training_details_appendix}. The split between train and validation set is $90\%$ - $10\%$ and the learning rate is fixed to $5 \times 10^{-4}$. The transformer was trained for over $200$ epochs on an NVIDIA TITAN X (Pascal) GPU for $10$ h, and the model with the lowest validation loss over those epochs was retained. We did not find any accuracy improvement when using a scheduler. The efficiency of such strategy is shown in Appendices~\ref{sec:train_various_blocks_appendix} and~\ref{sec:simu_sensitivity}.
%\note{add a global diagram}

\section{Results on real-world contact maps}\label{sec:appli}
We evaluate the performance of \textit{BlockFormer} on two real-world tasks. First, we assess its accuracy on the centromere prediction task, demonstrating its generalization to a wide range of species. We also provide a thorough analysis on the reference organism \textit{S. cerevisiae}. We further show that our model extends to other tasks characterized by spot-like interaction patterns, such as loop localization. In all the cases, the parameter estimate $\hat{\theta}$ is computed as in Appendix~\ref{sec:parameter_construction}.

%\subsection{A model that generalizes to any species}
\paragraph{Centromere identification across diverse genome sizes and heterogeneous Hi-C maps.} % \label{sec:other_species}
%\note{mettre en valeur diversité + genome size}
To quantitatively evaluate prediction accuracy, we estimate centromere positions across species with known ground truth, although \textit{BlockFormer} can also estimate centromeres in species where they are unknown. Our analysis includes datasets from seven yeast species with varying numbers of chromosomes, the parasite \textit{P. falciparum} at three distinct lifecycle stages, and the plant \textit{A. thaliana} (see Table~\ref{tab:other_species_small},~\ref{tab:other_species_big} and Appendix~\ref{sec:other_species_appendix} %and Figure~\ref{fig:abc_sbi_sm}
for results). 
%As the number of trans-blocks considered does not really impact the performance of our model, 
The results are obtained by averaging the predictions of the model over several sub-sampled $2$ trans-blocks for fast runtime\footnote{Using larger number of blocks did not substantially affect the performance (see Table \ref{tab:comparison_sc}).}. 
Species details about full names and genomes are given in  Appendix~\ref{sec:other_species_appendix}.
\begin{table}[t]
\centering
\caption{\small Comparison of normalized error (the smaller the better) and time across species. (5 random seeds)}
\label{tab:other_species_small}
\renewcommand{\arraystretch}{1.1}
\setlength{\tabcolsep}{2.5pt}
\resizebox{\linewidth}{!}{%
\begin{tabular}{cc cccccccc}
\toprule
\multicolumn{2}{c}{\multirow{2}{*}{\textbf{Method}}} 
 & \multicolumn{2}{c}{\textbf{S.C.}} 
& \multicolumn{2}{c}{\textbf{L.K.}} 
& \multicolumn{2}{c}{\textbf{L.T.}} 
& \multicolumn{2}{c}{\textbf{S.M.}} \\
\cmidrule(r){3-4} \cmidrule(r){5-6} \cmidrule(r){7-8} \cmidrule(r){9-10}
&  & Err & Time & Err & Time & Err & Time & Err & Time \\
\midrule

\multirow{2}{*}{\rotatebox[origin=c]{90}{Init.}} 
& \textit{BlockFormer}
& \textbf{0.28 $\pm$ 0.02} & \textbf{0.36 $\pm$ 0.03} 
& 1.35 $\pm$ 0.09 & \textbf{0.25 $\pm$ 0.02} 
& \textbf{1.13 $\pm$ 0.04} & \textbf{0.31 $\pm$ 0.1} 
& \textbf{0.59 $\pm$ 0.05} & \textbf{0.40 $\pm$ 0.03} \\

& \textit{Centurion} 
& 0.59 $\pm$ 0.0 & 2.26 $\pm$ 0.06 & \textbf{0.48 $\pm$ 0.0} & 0.68 $\pm$ 0.01 & 1.28 $\pm$ 0.03 & 0.98 $\pm$ 0.29 & 0.70 $\pm$ 0.0 & 3.05 $\pm$ 0.09  \\

\midrule

\multirow{2}{*}{\rotatebox[origin=c]{90}{Fitting}} 
& \textit{BlockFormer}
& \textbf{0.08 $\pm$ 0.0} & \textbf{2.75 $\pm$ 0.1} 
& \textbf{0.17 $\pm$ 0.0} & \textbf{1.75 $\pm$ 0.04} 
& \textbf{0.33 $\pm$ 0.04} & \textbf{1.93 $\pm$ 0.45} 
& \textbf{0.17 $\pm$ 0.0} & \textbf{3.51 $\pm$ 0.21 } \\

& \textit{Centurion} 
& \textbf{0.08 $\pm$ 0.0} & 11.73 $\pm$ 0.27 & \textbf{0.17 $\pm$ 0.0} & 2.14 $\pm$ 0.09 & \textbf{0.33 $\pm$ 0.04} & 7.27 $\pm$ 1.17 & \textbf{0.17 $\pm$ 0.0} & 9.45 $\pm$ 0.32 \\

\bottomrule
\end{tabular}
}
\end{table}

\begin{table}[t]
\centering
\caption{\small Species with out-of-training chromosomes sizes, too noisy map or high resolution: we apply the refinement step as in Appendix ~\ref{sec:refine}. P.F.r. stands for the stage rings of the parasite.  Comparison of normalized error (the smaller the better) and time. (5 random seeds)}
\label{tab:other_species_big}
\renewcommand{\arraystretch}{1.2}
\setlength{\tabcolsep}{2pt}
\resizebox{\linewidth}{!}{%
\begin{tabular}{cc cccccccc}
\toprule
\multicolumn{2}{c}{\multirow{2}{*}{\textbf{Method}}} 
& \multicolumn{2}{c}{\textbf{S.P.}} 
& \multicolumn{2}{c}{\textbf{A.T.}}
& \multicolumn{2}{c}{\textbf{P.F.r.}}
& \multicolumn{2}{c}{\textbf{Z.R.}} \\
\cmidrule(r){3-4}
\cmidrule(r){5-6}
\cmidrule(r){7-8}
\cmidrule(r){9-10}
& & Err & Time & Err & Time & Err & Time & Err & Time \\
\midrule

\multirow{2}{*}{\rotatebox[origin=c]{90}{Init.}} 
& \textit{BlockFormer}
& \textbf{2.39 $\pm$ 0.14} & \textbf{0.17 $\pm$ 0.02}
& \textbf{4.19 $\pm$ 0.33} & \textbf{0.94 $\pm$ 0.27}
& \textbf{0.40 $\pm$ 0.02} & \textbf{0.91 $\pm$ 0.07}
& \textbf{0.81 $\pm$ 0.14} & \textbf{0.40 $\pm$ 0.03} \\

& \textit{Centurion}
& 26.10 $\pm$ 0.0 & 0.06 $\pm$ 0.01
& 162.92 $\pm$ 0.0 & 35.87 $\pm$ 4.66
& 1.54 $\pm$ 0.0 & 13.11 $\pm$ 0.03
& 3.31 $\pm$ 0.0 & 0.61 $\pm$ 0.01 \\

\midrule

\multirow{2}{*}{\rotatebox[origin=c]{90}{Fitting}} 
& \textit{BlockFormer}
& \textbf{0.93 $\pm$ 0.0} & \textbf{1.0 $\pm$ 0.1}
& \textbf{6.12 $\pm$ 1.55} & \textbf{56.66 $\pm$ 48
54}
& \textbf{0.18 $\pm$ 0.0} & \textbf{34.81 $\pm$ 0.74}
& \textbf{0.27 $\pm$ 0.0} & \textbf{2.15 $\pm$ 0.04} \\

& \textit{Centurion}
& 23.92 $\pm$ 10.04 & 10.5 $\pm$ 0.22
& 34.82 $\pm$ 0.0 & 13103.07 $\pm$ 1795.76
& \textbf{0.18 $\pm$ 0.0} & 1064.73 $\pm$ 15.24
& \textbf{0.27 $\pm$ 0.0} & 5.85 $\pm$ 0.10 \\

\bottomrule
\end{tabular}%
}
\end{table}

Across Table~\ref{tab:other_species_small}, the network consistently achieves near-resolution accuracy (e.g. $0.59$ for S.M. (16 chr.), $1.35$ for L.K. (8 chr.)), showing robustness to varying block sizes and numbers.\\
%(e.g. L.K. has $8$ chromosomes ranging from $9.5 \times 10^5$ to $2.3 \times 10^6$ bp).\\
Across Tables~\ref{tab:other_species_small}, {\ref{tab:other_species_big} and \ref{tab:other_species_big_1}, \textit{BlockFormer} is consistently faster than \textit{Centurion} in both initialization and fitting (e.g. $0.18$ error in $34.81$ s versus $1064.73$ s for P.F.r.). Unlike \textit{Centurion}, which must be rerun for each species, our model is amortized: trained once during $10$h on GPU, it requires only forward passes for inference. For species with out-of-training chromosomes sizes, \textit{Centurion} is slow and often less accurate (e.g. for A.T., $34.82$ error in $\approx$ 3.5 h versus $6.12$ in $56.66$ s). Overall, even with training cost, after processing such three species, our model becomes more cost-efficient.
In most of the cases, \textit{BlockFormer} provides a substantially more accurate and efficient initialization for the fitting step, but can also offer a standalone sub-resolution estimate when fitting is unnecessary or computationally impractical (e.g. for species such as A.T., S.P. or  P.F.).\\
All real Hi-C maps considered in this study are normalized using ICE. We show in Appendix~\ref{sec:hic_norm_appendix} that \textit{BlockFormer} is robust to the absence of map normalization. 

%\note{ajout texte explicatif qui presente plan}
%\subsection{Centromere identification}
\paragraph{Focus on a reference case: \textit{Saccharomyces cerevisiae} (S.C.).}
We analyze the performance of our model on a reference case: the yeast \textit{S. cerevisiae}. This organism has known centromere positions, and deep sequencing produced a low-noise genome-wide contact matrix at resolution $32$~kb.

\textit{Point estimate. \label{sec:point_estimate}}
We compute a point estimate of each centromere using \textit{BlockFormer}. The parameter can be further refined using the fitting procedure from \cite{nelle}, which enforces both horizontal and vertical alignment between centromere positions. 
Table~\ref{tab:comparison_sc} reports the mean absolute error over the number of chromosomes normalized by the resolution, as well as the runtime on a CPU, for various block numbers $k$ and compares them with \textit{Centurion}.
% As our architecture is flexible to the number of blocks, we compare the number of trans-blocks that must be taken. For each number of blocks $k$, $10$ random choices of $k$ blocks are sampled leading to $10$ forward passes through the network and thus to $10$ candidates. The estimated $\theta$ for $k$ blocks is then the mean over those $10$ candidates. We report the accuracy (mean absolute error over the number of chromosomes) as well as the runtime on a CPU in the Table~\ref{tab:comparison_sc} and compare with Centurion method \cite{nelle}. 
\begin{table}[t]
\centering
\small
\setlength{\tabcolsep}{4pt}
\caption{\small Comparison for \textit{S. cerevisiae} at $32$~kb resolution. We report the mean absolute error divided by the resolution (the smaller the better). The number of blocks is indicated in brackets.}
\label{tab:comparison_sc}
\begin{minipage}{0.48\textwidth}
\centering
\textbf{Pre-localization}\\[0.3em]
\begin{tabular}{lcc}
\toprule
Method & Norm. error & Time (s) \\
\midrule
\textbf{\textit{BlockFormer} (1)}  & \textbf{0.26 $\pm$ 0.01} & \textbf{0.46 $\pm$ 0.12} \\
\textit{BlockFormer} (3) & 0.30 $\pm$ 0.01 & 0.68 $\pm$ 0.05 \\
\textit{BlockFormer} (5) & 0.37 $\pm$ 0.01 & 0.53 $\pm$ 0.01 \\
\textit{BlockFormer} (10) & 0.43 $\pm$ 0.01 & 1.08 $\pm$ 0.01 \\
\textit{BlockFormer} (15) & 0.58 $\pm$ 0.0 & 2.02 $\pm$ 0.02 \\
Initialization \textit{Centurion} & 0.62 $\pm$ 0.0 & 2.99 $\pm$ 0.16 \\
\bottomrule
\end{tabular}
\end{minipage}
\hfill
\begin{minipage}{0.48\textwidth}
\centering
\textbf{Full approach}\\[0.2em]
\begin{tabular}{lcc}
\toprule
Method & Norm. error & Time (s) \\
\midrule
\textit{BlockFormer} (1) + fitting  & 0.08 $\pm$ 0.0 & 3.54 $\pm$ 0.54 \\
\textit{BlockFormer} (3) + fitting  & 0.08 $\pm$ 0.0 & 4.03 $\pm$ 0.46 \\
\textbf{\textit{BlockFormer} (5) + fitting}  & \textbf{0.08 $\pm$ 0.0} & \textbf{2.94 $\pm$ 0.15} \\
\textit{BlockFormer} (10) + fitting  & 0.08 $\pm$ 0.0 & 3.31 $\pm$ 0.03 \\
\textit{BlockFormer} (15) + fitting  & 0.08 $\pm$ 0.0 & 4.29 $\pm$ 0.07 \\
\textit{Centurion} & 0.08 $\pm$ 0.0 & 16.40 $\pm$ 1.56 \\
\bottomrule
\end{tabular}
\end{minipage}
\end{table}
%ajouter init + fitting all blocks
Overall, the number of chosen blocks has little impact on performance, as errors remain below the resolution. Moreover, the network outputs a more accurate initial candidate ($0.34$ for $3$ blocks versus $0.58$ for \textit{Centurion}) in significantly less time. 
Applying the fitting step further improves accuracy across all methods, yielding highly precise estimates ($0.08$ error or approximately $2$~kb), well below the data resolution ($32$~kb), while achieving a substantial speedup with \textit{BlockFormer} compared to \textit{Centurion} ($2.64$~s versus $12.37$~s). 
%%Applying the refinement step further improves the accuracy for all methods, yielding a very precise estimate (around $2$~kb) well below the data resolution ($32$~kb), with a significant speed up using the network compared to Centurion ($2.64$~s versus $12.37$~s). 
%However, the optimization phase of \textit{Centurion} leads to very precise estimate (around $2$~kb). If we want to achieve this level of precision using our network, we run \textit{Centurion} with our estimate as initialization and we end up with the same accuracy in less time ($2.64$~s versus $12.37$~s).
\begin{figure*}[!t]
  \centering
  
  \includegraphics[width=\textwidth]{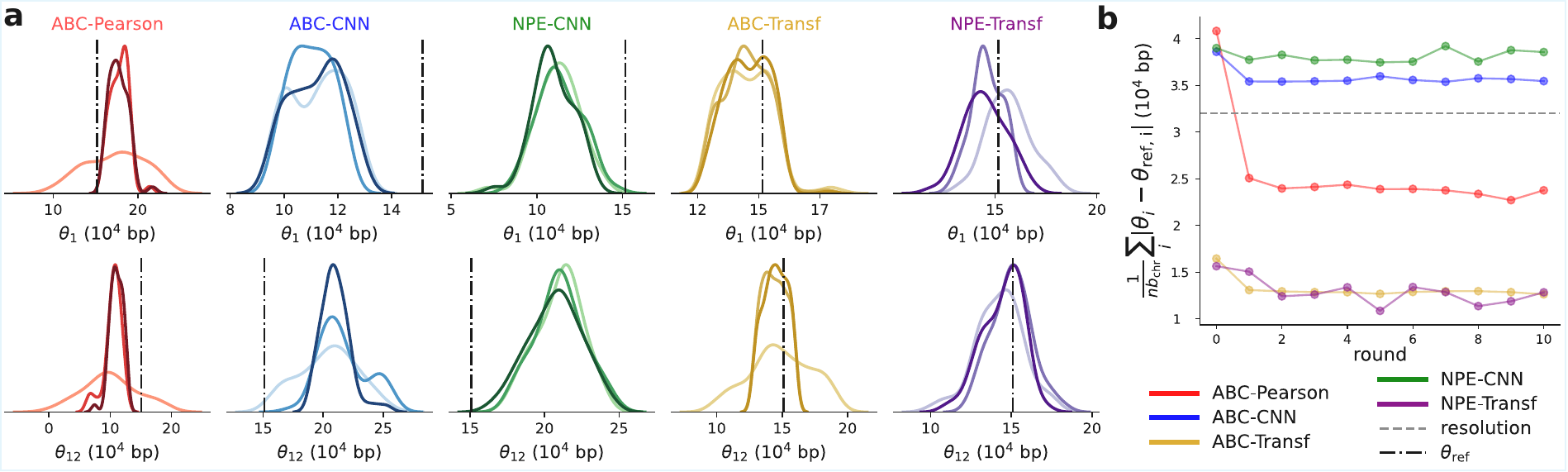}%

  \caption{\small Inference using ABC-Pearson (see Appendix~\ref{sec:SMCABCPearson}), ABC-CNN, ABC-Transf, (see Appendix~\ref{sec:ABCCNN_appendix}) NPE-CNN, and NPE-Transf (\textbf{a}) (see Appendix~\ref{sec:snpe_appendix}). Color shades increase from lightest to darkest across rounds. Densities are estimated with the $5\%$ best $\theta$ according to the ABC criterion or sampled from the flow. In some dimensions, only the \textit{BlockFormer}-based approaches ABC-Transf and NPE-Transf are accurate (e.g. the densities for the centromere of chromosome $1$ and $12$), the CNN-based methods lead to biased densities. (\textbf{b}) Mean absolute error distance between $\theta$ and $\theta_\text{ref}$, computed over the $5\%$ best-performing samples. The horizontal gray dashed line stands for the resolution of the contact map $C_\text{ref}$ (in bp).
    %%In some dimensions, all the methods are accurate: the densities are very peaky and centered around $\theta_i$ (e.g. chromosome 1, 13, 15) but in others, 
    %%Overall, transformer-based approaches outperform all other methods.  
    }
  \label{fig:abc_sbi_densities}
\end{figure*}

\textit{Posterior estimation.}
%\note{reduire}
\textit{BlockFormer} and \textit{Centurion} output only a point estimate of each centromere whereas it is actually a whole segment of chromosome. To quantify uncertainty about parameter estimations, we instead target the posterior density $p(\theta | C_{\text{ref}})$ using a Bayesian approach. To reduce dimensionality of the input map, the pre-trained model \textit{BlockFormer} serves as summary statistic within the inference approaches. Since our model infers $\theta_i$ from $C_i$, we actually target in parallel each marginal. The inference framework strategy is detailed in Appendix~\ref{sec:sbi_presentation}.
Our choice of summary statistic is motivated by the fact that $\mathcal{BF}_\phi(C_i)$ trained with Equation~\ref{eq:loss} approximates the conditional expectation $\mathbb{E}\left[\theta_i|C_i\right]$ preserving first-order information.
We consider two simulation-based inference approaches: approximate Bayesian computation (method referred as ABC-Transf, see Appendix~\ref{sec:ABCCNN_appendix}) and neural posterior estimation (method referred as NPE-Transf, see Appendix~\ref{sec:snpe_appendix}).\\
%either via {approximate Bayesian computation} (Sequential Monte-Carlo ABC: SMC-ABC) or with a conditional normalizing flow (Sequential neural posterior estimation: SNPE) \cite{snpea, snpec}.\\
%To tackle inference problem where parameters $\theta$ have various sizes, we target in parallel each marginal such that $p(\theta | C_\text{ref}) = \prod_{i=1}^I p(\theta_i | C_{\text{ref}, i})$ where $C_i$ is the parameter-related part of the interaction maps (see Appendix ~\ref{sec:sbi_presentation} for details about the inference framework strategy).\\
%We consider two simulation-based inference approaches: approximate Bayesian computation (ABC) and neural posterior estimation (NPE).\\
%To reduce dimensionality of the input map, the pre-trained model \textit{BlockFormer} serves as summary statistic within both inference approaches -- ABC-Transf and NPE-Transf (see Appendix \ref{sec:ABCCNN_appendix} and \ref{sec:snpe_appendix}). Our choice of summary statistic is motivated by the fact that $\mathcal{T}_\phi(C)$ trained with (\ref{eq:loss}) approximates the conditional expectation $\mathbb{E}\left[\theta|C\right]$ preserving first-order information.\\
%We compare ABC-Transf and NPE-Transf to baselines namely ABC-Pearson, ABC-CNN and NPE-CNN (see Appendix~\ref{sec:SMCABCPearson}, \ref{sec:ABCCNN_appendix} and \ref{sec:snpe_appendix} respectively).\\
Figure~\ref{fig:abc_sbi_densities}~(\textbf{a}) shows the marginal densities for two components of $\theta$. \textit{BlockFormer}-based methods produce sharper and better-calibrated posteriors, while CNN-based approaches exhibit bias. Benchmark of metrics in Figure~\ref{fig:abc_sbi_densities}~(\textbf{b}) and Figure~\ref{fig:abc_sbi_metrics} in Appendix~\ref{sec:sc_post} as well as calibration diagnostics confirm this trend. For instance, the Wasserstein-2 distance between inferred posteriors and posterior ground truth is lowest for \textit{BlockFormer}-based approaches, indicating an error of twice the resolution between samples and ground truth.  

\paragraph{Loop localization.}
\textit{BlockFormer} is applicable to parameter inference using spot-like patterns in interaction maps. Among real applications, loop localization is of particular interest. Loops often connect two functional elements thanks to a DNA-binding regulatory protein CTCF \cite{loop_protein}. In Hi-C maps from eukaryotes, they appear as multiple bright and isolated dots away from the main diagonal in cis-blocks, their total number and position being unknown. We evaluate our approach using Hi-C data from the human cell line IMR90 at resolution $5$ kb. However, because \textit{BlockFormer} is designed to produce a single parameter estimate per entity using multiple trans-blocks as input, the standard multi loops detection framework is not directly compatible. To ensure consistency with the training setup, we restrict analysis to chromosomal regions containing a single loop and report results in Appendix~\ref{sec:loop_appendix}. Especially, we achieve better performance than \textit{Centurion}, that is not well-adapted to inference from single block.

\section{Ablations and synthetic experiments \label{sec:ablations}}
%\note{ablation + synthetics}
%\note{1 sous-section avec effect of nb/size seq depth et une sous section ablations sur CNN + une sur patterns}

\paragraph{Ablations of block-aware modeling.}
One of our main contributions is the design of a block-aware architecture consisting in per-block padding and per-block 3D positional encoding that preserves the block-wise structure of the map and that cannot be naturally included in CNN-based architectures. We consider several positional encoding strategies including no positional encoding and a few 2D encoding alternatives.  
All the transformers are trained in identical conditions on $1$ to $9$ blocks (see Appendix~\ref{sec:training_details_appendix}). To assess the impact of block-aware modeling, we consider synthetic maps that are sequences of $1$ to $14$ trans-blocks (see details in Appendix~\ref{sec:position_encoding_appendix}). According to Tables~\ref{tab:ablation_pos_main} and~\ref{tab:ablation_pos}, the per-block 3D positional encoding along with the per-block padding method (BF : 3D pos. per block) is the best overall method: it has lowest or near-lowest mean error most consistently with competitive or best median performance and is in general more stable (lower std in most settings).
\begin{table}[t]
\centering
\caption{\small Normalized absolute error comparison across methods for low/extreme regime (1, 14 blocks). 3D pos. per-block outperforms others methods being the most stable across the number of blocks.}
\label{tab:ablation_pos_main}
\small
\begin{tabular}{c ccc ccc}
\toprule
& \multicolumn{3}{c}{\textbf{1 Block}} 
& \multicolumn{3}{c}{\textbf{14 Blocks}} \\
\textbf{Method} 
& Mean$\pm$Std & Median & 95\% CI
& Mean$\pm$Std & Median & 95\% CI \\
\midrule

\textbf{BF : 3D pos. per block} 
& \textbf{0.43 $\pm$ 0.32} & 0.39 & [0.02, 1.22]
& \textbf{0.36 $\pm$ 0.29} & 0.29 & [0.01, 0.93] \\

\textbf{2D pos. per block} 
& 0.73 $\pm$ 2.21 & 0.40 & [0.01, 1.99]
& 0.38 $\pm$ 0.29 & 0.38 & [0.02, 1.17] \\

\textbf{2D pos.} 
& 0.65 $\pm$ 2.01 & 0.39 & [0.04, 1.19]
& 0.55 $\pm$ 0.46 & 0.43 & [0.02, 1.82] \\

\textbf{2D pos. pad.} 
& 0.64 $\pm$ 2.41 & 0.33 & [0.02, 1.20]
& 0.53 $\pm$ 0.45 & 0.41 & [0.02, 1.66] \\

\textbf{1D pos.} 
& 1.04 $\pm$ 0.92 & 0.73 & [0.04, 3.41]
& 2.21 $\pm$ 5.09 & 0.59 & [0.03, 19.61] \\

\textbf{no pos.} 
& 4.16 $\pm$ 4.75 & 2.11 & [0.17, 16.83]
& 3.49 $\pm$ 4.82 & 1.06 & [0.03, 17.39] \\

\bottomrule
\end{tabular}
\end{table}

% \subsection{Robustness to heterogeneous interaction map structure and coverage}
%\subsection{A model that generalizes to various resolutions/ sizes of blocks}

We now evaluate the robustness of \textit{BlockFormer}, trained on contact maps at $32$ kb resolution, to varying numbers and sizes of blocks, different sequencing depths, and diverse spot-like patterns.

\textbf{Robustness to various sizes of blocks.}
The size of each block in a Hi-C map depends on the species or the selected resolution. For instance, in S.C., the largest chromosome is $1.5 $ Mbp long, while it is  nearly $30$ times larger in the plant A.T.. 
%measures 30 427 671 bp. 
At $40$~kb resolution, this translates respectively to blocks of height $38$ and $760$. When the resolution is doubled ($20$~kb), the heights are also doubled.\\
We show in Figure~\ref{fig:simu_variable} that \textit{BlockFormer} generalizes to various sizes of blocks:  
%We simulate $1\ 000$ synthetic maps at resolution $32$~kb from synthetic genomes of $2$ to $11$ chromosomes, whose sizes vary from $2 \times 10^{5}$ bp to $2$ Mbp. 
%Per number of chromosomes $k+1$, each map consists of $k$ trans-blocks of various sizes in which the position and size of centromere spots also vary. We report the absolute error between the estimated centromere and the real one for each number of trans-blocks (see Figure~\ref{fig:simu_variable}).
most of the errors remain below the resolution, whatever the number of blocks. However, high dispersion occurs in some cases (e.g., $1$, $7$ blocks), with estimates accuracy ranging from $100$ bp to $0.5$ Mbp.\\
We then show in Figure~\ref{fig:mean_simu_resolution} that \textit{BlockFormer} generalizes to different resolutions. For this, we simulate synthetic maps from the S.C. genome and downsample the maps to resolutions from $20$~kb to $70$~kb (see Appendix ~\ref{sec:downsample_appendix} for the downsampling procedure). \textit{BlockFormer} has issues for some chromosomes at high resolution $20$ kb because block sizes can be far from the training distribution.

%\subsection{A model that generalizes to various number of blocks}
\textbf{Robustness to various numbers of blocks.}
The number of chromosomes varies across species: e.g., S.C. has $16$ chromosomes, whereas A.T. only has $5$. Since a trans-block in a Hi-C map represents interactions between $2$ chromosomes, the total number of blocks differs across species. We show in Figure~\ref{fig:mean_simu_resolution} that \textit{BlockFormer} generalizes to varying numbers of blocks. 
%, we generate $100$ synthetic maps from the S.C. genome and consider $1$ to $15$ trans-blocks (see ). 
The parameter estimate $\hat{\theta_i}$ is constructed as in Appendix~\ref{sec:parameter_construction}.
%For each number of blocks $k$, we select 10 random groups of $k$ blocks and output $10$ candidates. The final estimation for $k$ blocks is the average of those candidates.
\begin{figure*}[t]
   \centering
   \includegraphics[width=\textwidth]{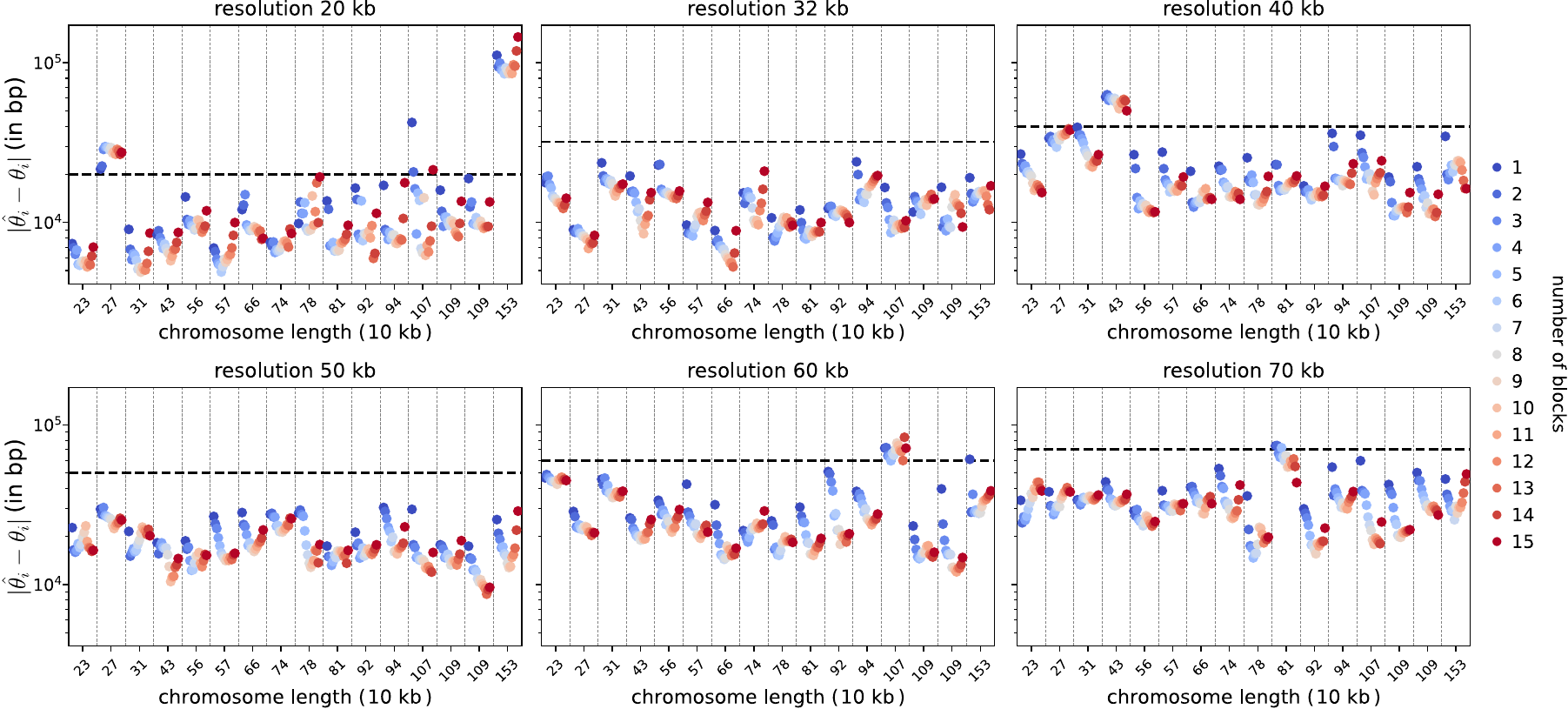}
   \caption{\small Absolute error per centromere over $100$ synthetic maps generated from the S.C. genome. For each number of blocks $k$ and each chromosome $i$, we report the absolute error $\mathrm{err}_i^k$ (see details in \ref{sec:seq_depth_appendix}). Target chromosomes $i$ on the x-axis are sorted by length (bp). Color shades range from blue to red as the number of blocks $k$ increases from $1$ to $15$. Across resolutions, the centromeres are estimated with a precision lower than the resolution. Neither the size of the chromosome nor the number of blocks deteriorate \textit{BlockFormer}'s accuracy.
   %and can even improve it.
   }
   %Absolute error per centromere over 100 synthetic contact maps generated from the S.C. genome. For each number of blocks $k$ and each chromosome $i$, we report the absolute error $err_i^k$ (see details in \ref{sec:seq_depth_appendix}). 
   %Per number of blocks $k$, we randomly select 10 sets of $k$ blocks of various sizes from the map (e.g. for $k=2$, we choose blocks $(2,9)$ (choice $l=1$), blocks $(1,7)$ (choice $l=2$), ...). Then, for target chromosome $i$, $k$ blocks, and blocks selection $l$ , we compute $e^{k,l}_i = \frac{1}{100}\sum_{n=1}^{100} |\hat{\theta}_i^{n,k,l} - \theta_i|$.  We report $err^k_i = \frac{1}{10} \sum_{l=1}^{10} e^{k,l}_i$ . 
   %Target chromosomes on the x-axis are sorted by length (bp). Color shades range from blue to red as the number of blocks $k$ increases from 1 to 15.}
\label{fig:mean_simu_resolution}
\end{figure*}
Neither the size of the chromosome nor the number of blocks seems to impact  accuracy: most of the estimations are under the resolution, and there are no large performance changes when the number of blocks increases.
%\subsection{A model that generalizes to various sequencing depth/ quality of the map}

\textbf{Robustness to various sequencing depths.}
The quality of a Hi-C map is directly influenced by sequencing depth: a higher sequencing depth yields higher resolution maps and reduces noise, but at an increased cost. For example, the S.C. map at $30$~kb is remarkably clean in contrast to the S.K. map. We show in Appendix~\ref{sec:seq_depth_appendix} and Figure~\ref{fig:seq_depth_synth_ref} that \textit{BlockFormer} generalizes well to different sequencing depth. We provide results on both synthetic but also reference map of S.C., keeping between $10\%$ and $50\%$ of the sequencing depth.  Most of the errors remain below the resolution in both settings. Lower sequencing depth ($10\%$) is more difficult since the maps become very noisy. 
%, we generate $100$ synthetic maps based on the S.C. genome and downsample them to different sequencing depths. We do the same on the reference map of S.C., keeping between $10\%$ and $50\%$ of the sequencing depth (see Appendix~\ref{sec:seq_depth_appendix} and Figure~\ref{fig:seq_depth_synth_ref}). Overall, \textit{BlockFormer} generalizes well to different sequencing depths, with the estimate below the resolution most of the time.

\textbf{Robustness to diverse spot pattern structures.}
We evaluate the robustness of \textit{BlockFormer} on contact maps with spots of varying shapes, comparing it to \textit{Centurion}. 
%We report the normalized mean absolute error between $\hat{\theta}$ and $\theta$ along with the runtime 
The estimate $\hat{\theta}$ is constructed as in Appendix~\ref{sec:parameter_construction}. 
%In each setting, we generate $100$ synthetic maps at resolution $30$~kb based on the reference genome of S.C..

%\textit{Centurion}'s strength lies in its optimization step, which forces the alignment of all the Gaussian spots in the map. However, it highly relies on good pre-localization of candidates. The initialization of \textit{Centurion} consists of heuristic filtering of candidates: as centromeric regions represent enrichment in trans-blocks, the algorithm first seeks a few local maxima in each block to form a set of candidates and then reduce it via few steps of optimization.\\
\textit{Spots corrupted by noise.} The basic simulated map presents only one visible spot per block, but in real data, due to other genetic elements clustered together, the map can present multiple spots per interaction block or the spot can be affected by noise. Those confounders can affect the initialization of \textit{Centurion}, leading to inaccurate estimation. To test \textit{BlockFormer}'s robustness, we simulate maps with additional random bright pixels in each trans-block (see Fig.~\ref{fig:simu_trap} and~\ref{fig:comparison_nelle_simu_trap} in Appendix~\ref{sec:pirate_appendix}) or with an additional Gaussian spot randomly positioned in each trans-block (see Fig.~\ref{fig:comparison_nelle_simu}\textbf{a} and Fig.~\ref{fig:simu_multiple_spots} in Appendix~\ref{sec:multiple_spots_appendix}). In both settings, \textit{BlockFormer} is more accurate and faster than \textit{Centurion}.

\begin{figure}[!t]
\centering
    \includegraphics[width=\linewidth]
{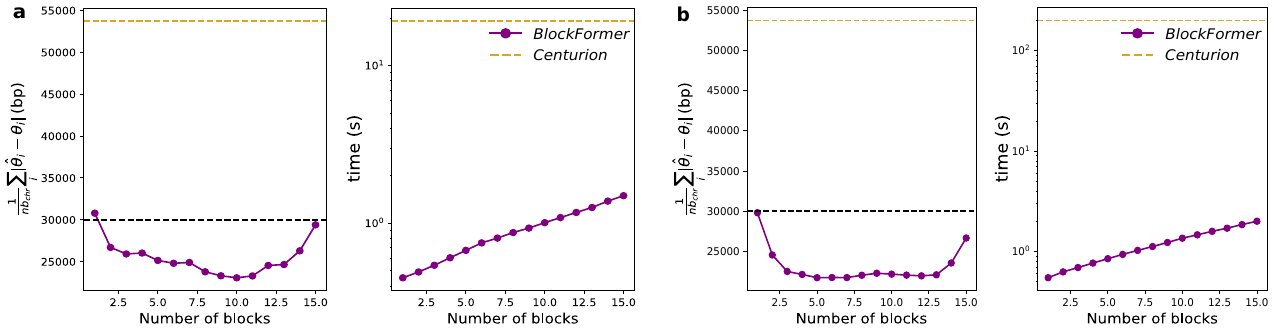}
\caption{\small Mean absolute error (in bp, 1$^{\text{st}}$ and 3$^{\text{rd}}$ panels) and runtime (in seconds, 2$^{\text{nd}}$ and 4$^{\text{th}}$ panels) over $100$ synthetic maps generated from the S.C. genome at resolution $30$~kb. \textbf{a}: per trans-block, one major Gaussian spot and one auxiliary Gaussian spot, smaller and less bright. \textbf{b}: Square spot in each trans-block. The black dotted line stands for the resolution. \label{fig:comparison_nelle_simu}}
\vspace{-2em}
\end{figure}

\textit{Square spots.} If large regions of DNA interact, leading to aggregation of multiple contacts (e.g. centromeres clustering of Drosophila but also enhancer–promoter contacts close together), the spots can be square. For such non-Gaussian spots, the optimization process of \textit{Centurion} based on Gaussian fitting can be challenging. To investigate this situation, we generate maps with square spots (see Fig.~\ref{fig:comparison_nelle_simu}\textbf{b} and Fig.~\ref{fig:simu_carre} in Appendix~\ref{sec:square_appendix}).
\textit{Centurion} is less accurate and slower than \textit{BlockFormer}. Moreover, our model estimates the parameter with sub-resolution precision for any number of blocks.

Other tests on different spot patterns inspired by real-world data such as Gaussian, ring, or elliptical spots are provided in Appendix~\ref{sec:gaussian_appendix}, ~\ref{sec:ring_appendix}, and ~\ref{sec:ellipse_appendix}. In nearly all settings, \textit{BlockFormer} achieves sub-resolution precision with faster runtimes than \textit{Centurion}. 

%\begin{figure}[!t]
%\centering
%\begin{subfigure}{0.49\linewidth}
%    \centering
%    \includegraphics[width=\linewidth]
%    %{comparison_nelle_simu_modif_cropped.pdf}
%    {comparison_simu_nelle_multiple_spots_init_1_cropped.pdf}
%\end{subfigure}\hfill
%\begin{subfigure}{0.49\linewidth}
%    \centering
%    \includegraphics[width=\linewidth]{comparison_nelle_simu_carre_modif_cropped.pdf}
%\end{subfigure}
%\caption{\small Mean absolute error (bp, left) and runtime (s, right) over $100$ synthetic maps generated from the S.C. genome at resolution $30$~kb. \textbf{a}: per trans-block, one major Gaussian spot and one auxiliary Gaussian spot, smaller and less bright. \textbf{b}: Square spot in each trans-block.\label{fig:comparison_nelle_simu}}
%\end{figure}

% \begin{figure}[!t]
%     \centering
%     \caption{\small Mean absolute error (bp, left) and runtime (s, right) over $100$ synthetic maps generated from the S.C. genome at resolution $30$~kb. The spot in each trans-block are squared. For the transformer-based network, we consider 1 to 15 trans-blocks. Per number of blocks $k$, 10 random subsets of $k$ blocks are sampled. Parameter estimation $\hat{\theta}_i^k$ is performed in parallel for each chromosome $i$ with $\hat{\theta}_i^k = \frac{1}{10} \sum_{l=1}^{10} \hat{\theta}_i^{l, k}$. We report the mean absolute error between $\hat{\theta}$ and $\theta$.\label{fig:comparison_nelle_simu_carre}}
% \end{figure}

\section{Conclusion}
We presented \textit{BlockFormer}, a novel architecture to infer per-entity parameters from genomic contact maps with localized patterns.
%given an interaction map with localized patterns. 
We designed a block-aware transformer-based model able to handle various sizes and numbers of interaction blocks, rendering inference flexible to many sets of entities. The network enables amortized accurate point estimation of the parameter in many synthetic or real-world scenarios with various spot patterns. It can also serve as an informative summary statistic within a Bayesian framework, enabling a principled quantification of uncertainty in parameter estimates.\\
We evaluate our approach in a biological setting where entities are chromosomes, interaction maps are Hi-C contact maps and parameters are centromere positions. 
%On this task, we show that our process is fully amortized: once the model is trained, centromere locations of any species can be inferred without re-training. 
On this task, our method is robust as it does not rely on any initialization or pre-localization: it uses an uninformative prior, randomly setting each centromere in the range of its chromosome. Despite its generality to various spot patterns, \textit{BlockFormer} matches or outperforms the state-of-the-art \textit{Centurion} in centromere identification.\\ 
However, our model always outputs a single parameter per entity which limits its applicability when the number of parameters is unknown or when some maps contain no parameter information. In particular, loop localization using \textit{BlockFormer} relies on  pre-filtered single-loop regions.\\  
%The probabilistic framework that we use in a second step allows us to quantify the uncertainty about the parameter estimates via the posterior densities. 
Our entire inference pipeline is based on a large number of simulations: to mitigate computing bottlenecks, we designed a {simplified} but efficient contact maps {simulator}. While introducing a mismatch between real and synthetic data, it still yields very convincing results for {inferences on real experimental data}, requiring minimal preprocessing. Future work could consider applying \textit{BlockFormer} to other inference tasks from interaction maps such as inferring the 3D configuration of the genome, with the model trained accordingly with other biologically-inspired simulators. 

%\textit{BlockFormer} can be applied as is to other inference tasks involving interaction maps but needs to be retrained on appropriate simulated data.

\section*{Acknowledgments and Disclosure of Funding}
This work was supported by the ANR project BONSAI (grant ANR-23-CE23-0012-01), the ANR project Bayes-Duality (grant ANR-21-JSTM-0001) and by the ANR project SBI4C of the MIAI AI Cluster (grant ANR-23-IACL-0006). This work was granted access to the HPC resources of IDRIS under the allocation AD010616970 made by GENCI.

\newpage
\bibliographystyle{plain} % apalike
\bibliography{biblio}
%%%%%%%%%%%%%%%%%%%%%%%%%%%%%%%%%%%%%%%%%%%%%%%%%%%%%%%%%%%%
\newpage
\appendix

\tableofcontents
\newpage

\section{Contact maps \label{sec:hic_appendix}}
%\subsection{Generic structure. }
A genome-wide contact map summarizes all the chromatin contacts observed over a population of DNA configurations. To construct it, we define the resolution of the map, which is the length of the chromosome fragment that is represented by one pixel in the map. Each chromosome is cut into fragments and each entry of the map represents the contact counts of any fragment with another one over the population of DNA. This creates a matrix by blocks of interactions between chromosomes: for instance, the $i^{\text{th}}$ line of blocks in the map summarizes the interactions of chromosome $i$ with all the other chromosomes. Usually, we represent them by a heatmap as in Figure~\ref{fig:contactmap_schema}.
\begin{figure}[H]
    \centering
    \includegraphics[width=9cm, height=3.5cm]{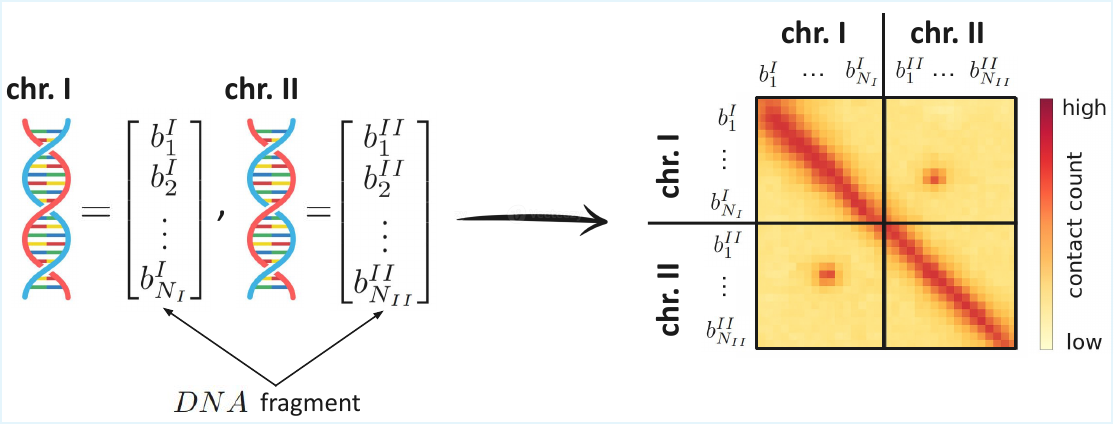}
    \caption{\small Process to construct a contact map in the case of $2$ chromosomes. \label{fig:contactmap_schema}}
\end{figure}

%\subsection{Hi-C map normalization. \label{sec:hic_norm_appendix}}

\section{A state-of-the-art method for centromere identification: \textit{Centurion} \label{sec:centurion}}
\cite{nelle} tackled the problem of centromere identification based on Hi-C data with an algorithm called \textit{Centurion}. It starts with a normalized Hi-C map and tries to identify peaks of interactions in the trans- (or inter-chromosomal) blocks for centromere initial candidates.\\
After heuristic filtering, one candidate is chosen for each chromosome, and the set of candidates serves as initialization of a joint optimization procedure to refine the estimated positions.\\
The interaction between any chromosome $k$ and $l$ in a window is modeled by a 2D Gaussian centered at $(\theta_k, \theta_l)$. To refine centromere positions, it performs a least-squares fit over all pairs of windows under the constraint that a centromere lies in its chromosome range. \textit{Centurion}'s strength lies in this optimization step, which forces the alignment of all the Gaussian spots in the map. However, it highly relies on good pre-localization of candidates.\\
This method is deterministic as it outputs only the mean position of each centromere, and the non-convex optimization process can be time-consuming for large matrices. In addition, its accuracy strongly depends on accurate pre-localization of the centromeres. Because the optimization relies on a specific Hi-C map, changing the species requires restarting the entire procedure from scratch. On the contrary, \textit{BlockFormer} is amortized: once trained, the network can infer the centromere of other species (see Section~\ref{sec:appli} and Appendix~\ref{sec:other_species_appendix}).\\
Moreover, \textit{Centurion} is specifically fine-tuned for centromere identification and needs adaptations before being used for other inference tasks (e.g. loop localization). 
%Building on the observations that Hi-C data contain inherently stochasticity and that centromeres are actually fragments of DNA, \cite{etouron} \note{mettre papier en appendix} proposed a method to infer centromere positions from simulated data while accounting for this stochasticity. However, their approach has notable limitations: the described architecture is tailored to a specific organism and lacks the flexibility to handle contact maps of varying sizes. In contrast, the present work extends this idea by introducing a transformer-based architecture that is more flexible (see Appendix~\ref{sec:archi_comparaison}), enabling the analysis of contact maps with arbitrary sizes, varying numbers of blocks, and different resolutions.  

\newpage
\section{The simulator}
\subsection{Simulation process \label{sec:simulator_appendix}}
The goal of the simulator is to create a contact map $C$ rapidly given the centromere positions $\theta$. As $C$ is symmetric, we only simulate the upper trans-blocks. We want to mimic the peak of interactions that appears in those blocks, as well as some rare interactions that can occur among the population of DNA. Given the $L$ chromosome lengths in bp $\{l_i \}_{1 \leq i \leq L}$, the centromere positions $\theta  =(\theta_1, ..., \theta_L)$ are sampled from the prior $\mathcal{U}(\underset{1 \leq i \leq L}{\prod} [1, l_i-1])$. To create each contact map $C$, the process is described in Algorithm~\ref{alg:simu}. 

\begin{algorithm}[H]
\caption{Simulator of contact maps \label{alg:simu}}
\begin{algorithmic}
\State \textbf{Input}: $L$ chromosome lengths in bp $\{l_i \}_{1 \leq i \leq L}$, resolution of the contact map in bp $r$ (e.g. $r = 32$~kb), centromere positions $\theta = (\theta_1, ..., \theta_L)$
\State \textbf{Return}: the upper trans-blocks of a simulated contact map $C$ at the resolution $r$ bp.
\State
\State choose the size of the peaks of interaction: sample $\sigma^2$ from $\mathcal{U}(0.1,10)$
\State choose the intensity of interaction $\alpha$ to simulate the DNA population size: sample $\alpha$ from $\mathcal{U}(\llbracket 1, 1\ 000\rrbracket)$
\State construct the upper trans-blocks of $C$ denoted $C^\text{upper}$ as:
\For{each chromosome pair $(i,j)$, $j>i$}
\State define a block of interaction $C^\text{upper}_{ij}$ of size ($\frac{l_i}{r}, \frac{l_j}{r}$)
\State define the center of the peak $(\theta_i, \theta_j)$ 
\State apply Gaussian density $\mathcal{N}( (\theta_i/r, \theta_j/r) , \sigma^2 )$ to the pixels of the block $C^\text{upper}_{ij}$ 
\State multiply each pixel of $C^\text{upper}_{ij}$  by the intensity factor $\alpha$
\State add Gaussian noise up to $10\%$ of the maximal value of $C^\text{upper}_{ij}$ to mimic the rare contacts:
\State construct a random matrix $M_{ij}$ of size ($\frac{l_i}{r}, \frac{l_j}{r}$) where each pixel is sampled from 
\State $\mathcal{N}(\max(C^\text{upper}_{ij}) \times 0.05, (\max(C^\text{upper}_{ij}) \times 0.05)^2)$, then add $M_{ij}$ to $C^\text{upper}_{ij}$
\State draw a cross of non interaction (set values to $0$) of width $\sigma$ passing through $(\theta_i, \theta_j)$
\EndFor
\State $C = C^\text{upper} + C^\text{upper, T}$
\State \textbf{return} a simulated contact map $C$ at resolution $r$ bp
\end{algorithmic}
\end{algorithm}

\begin{figure}[H]
    \centering
    \includegraphics[width=0.9\textwidth]{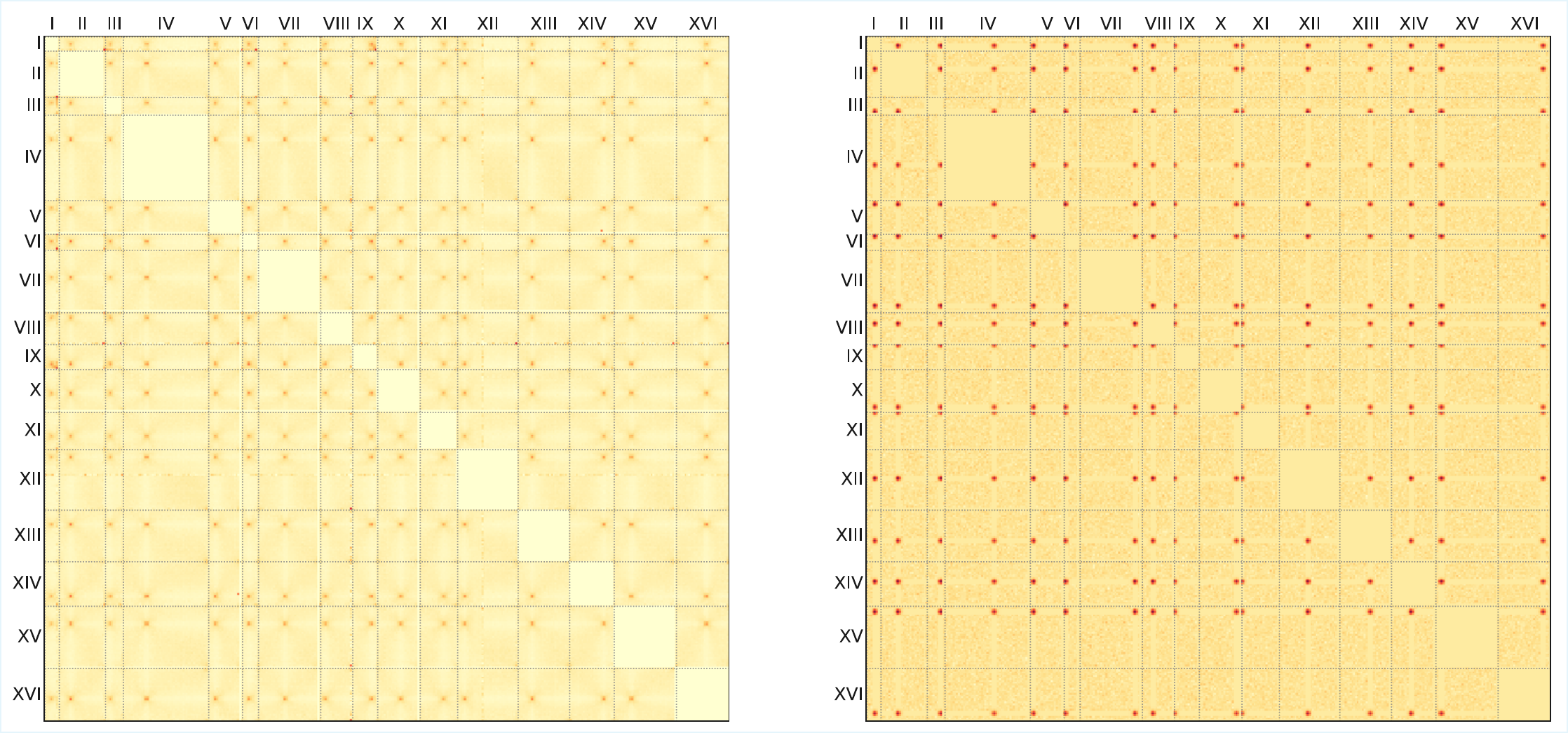}
    \caption{\small reference Hi-C map (left) and a simulated map (right) generated from the S.C. reference genome (resolution $32$~kb). \label{fig:C_simu}}
\end{figure}
%Our simulator outputs contact maps that present some dissimilarities with Hi-C maps. If we compute the row-based averaged Pearson correlation between $C_\text{ref}$ and $C$ simulated from $\theta_\text{ref}$ as in \cite{tjong}, we get a correlation of and $0.12$, which is quite low.\\
%However, concerning the inference task, our simulator estimates the centromere positions $\theta$ nearly as well on synthetic data (Figure~\ref{fig:abc_sbi_synthetic}) as on Hi-C data (Figure~\ref{fig:abc_sbi_densities}).  

%\begin{figure}[H]
%    \centering    
%    \includegraphics[width=0.7\textwidth]{abc_sbi_synthetic_cropped.pdf}
%    \caption{\label{fig:abc_sbi_synthetic}{\small Inference using \ABCPearson, \ABCTransf and \NPETransf from \emph{synthetic data}~(\textbf{a}). Color shades increase from lightest to darkest across rounds. Densities are estimated with the $5\%$ best $\theta$ according to the ABC criterion or sampled from the flow. 
    %%(For visibility, we select evenly spaced rounds for each method). 
    %We also report the mean absolute error between $\theta$ and $\theta_\text{ref}$, computed over the $5\%$ best-performing samples in the top right corner (\textbf{b}). The horizontal dashed line stands for the resolution of the contact map $C$ (in bp) in the top right figure. All approaches have errors smaller than the resolution of the contact maps and the densities are peaky around the ground truth in most of the dimensions.
%    }}
%\end{figure}

\newpage
\subsection{Misspecification analysis \label{sec:mismatch}}
Contact maps can vary substantially across species due to differences in genome folding organization and underlying physical constraints, as well as across experimental conditions including resolution, sequencing depth and protocol-specific biases. To mitigate these sources of heterogeneity, we intentionally designed a simplified simulator. It does not aim to reproduce all biological and technical complexities present in real Hi-C maps, but rather captures the common structural patterns shared across the different maps, necessary for inference. This approach reduces the risk of overfitting to a particular species or experimental protocol while enabling transferability of \textit{BlockFormer} to various organisms.\\ %the generation of contact maps that captures only the necessary information for inference, sometimes far from real contact maps.\\ 
The mismatch of the simulator can come from $3$ factors: \begin{itemize}
    \item the structure of the spot: to test it, we extract the real spots and add simulated noise to create simulated map with real spot (referred as $C_\text{spot}$). 
    \item the structure of the noise: to test it, we transport the Centurion pre-loc reference map to the closest simulated one via quantile mapping (referred as $C_\text{trans}$).
    \item the pirate pixels (e.g. telomeres interactions): in the cell, telomeres are tethered to the nuclear envelope making appear clustering patterns at the border of some interaction blocks in the contact maps. To test it, we use the reference map used in \textit{Centurion} pre-localization (remove the borders of each trans-block to remove telomeres interactions) (referred as $C_\text{cent}$).
\end{itemize}
The raw reference map is referred as $C_\text{raw}$ and the close simulated one as $C_\text{simu}$.\\
To analyze qualitatively the mismatch introduced by our simulator, we present heatmaps as well as the histograms of pixels colors of those maps. For fair comparison, each map is normalized between 0 and 1, the cis-blocks are set to Nan.\\ 
The metric used to quantify the mismatch is based on the Pearson correlation, commonly used in the domain. We average the Pearson correlation between each row of one trans-block of $C_\text{simu}$ and $C_\text{ref}$ and then average all those correlations over the number of upper trans-blocks (the closer to 1 the better).
We also report the runtime as well as the inference performance of the model with the normalized error: mean absolute error divided by the resolution (the smaller the better). Any value below 1 is satisfying (meaning an error below the resolution).

\newpage

\begin{figure}[H]
\centering
\includegraphics[width=\textwidth]{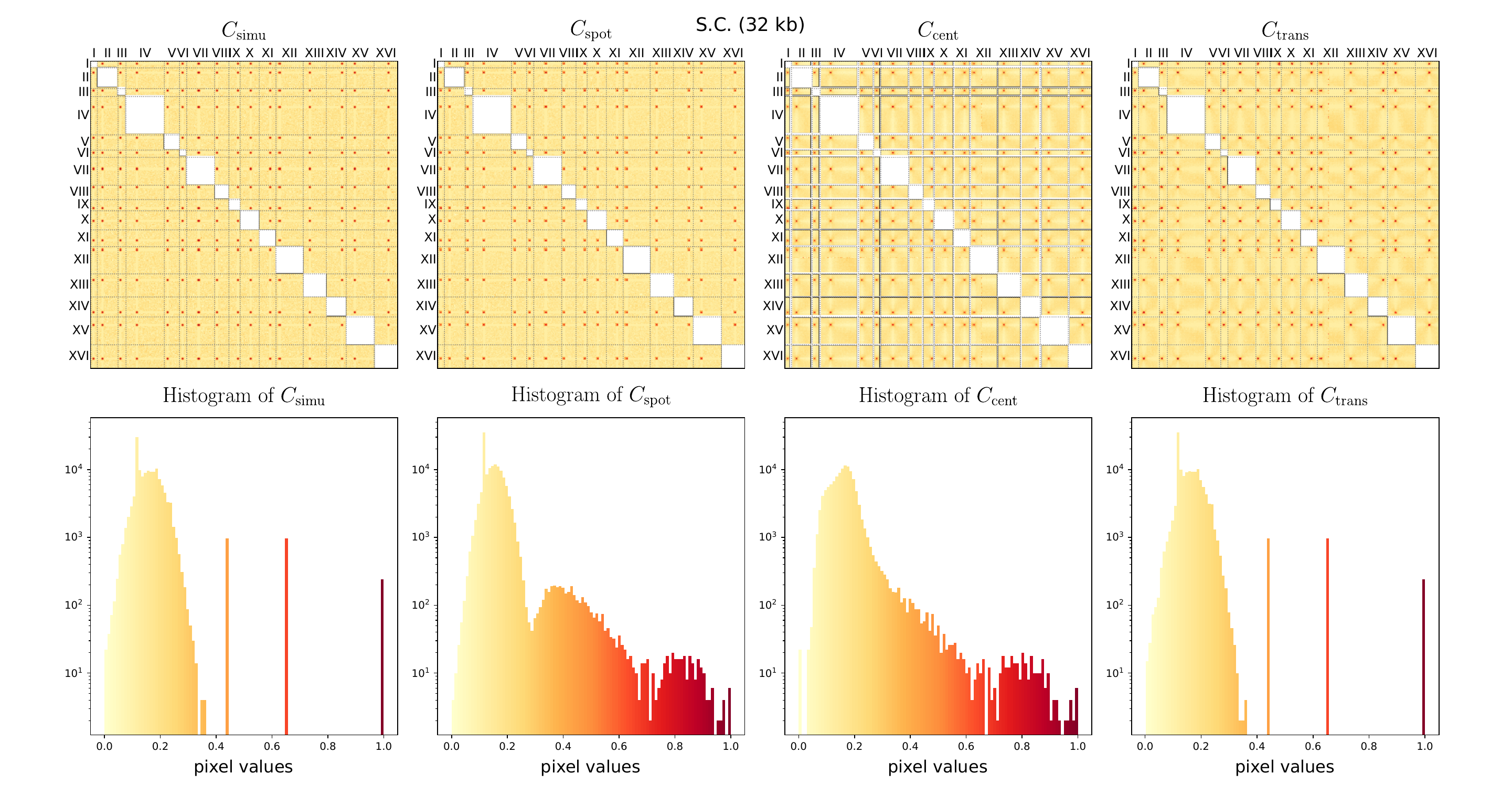}
\vspace{-4mm}
\includegraphics[width=\textwidth]{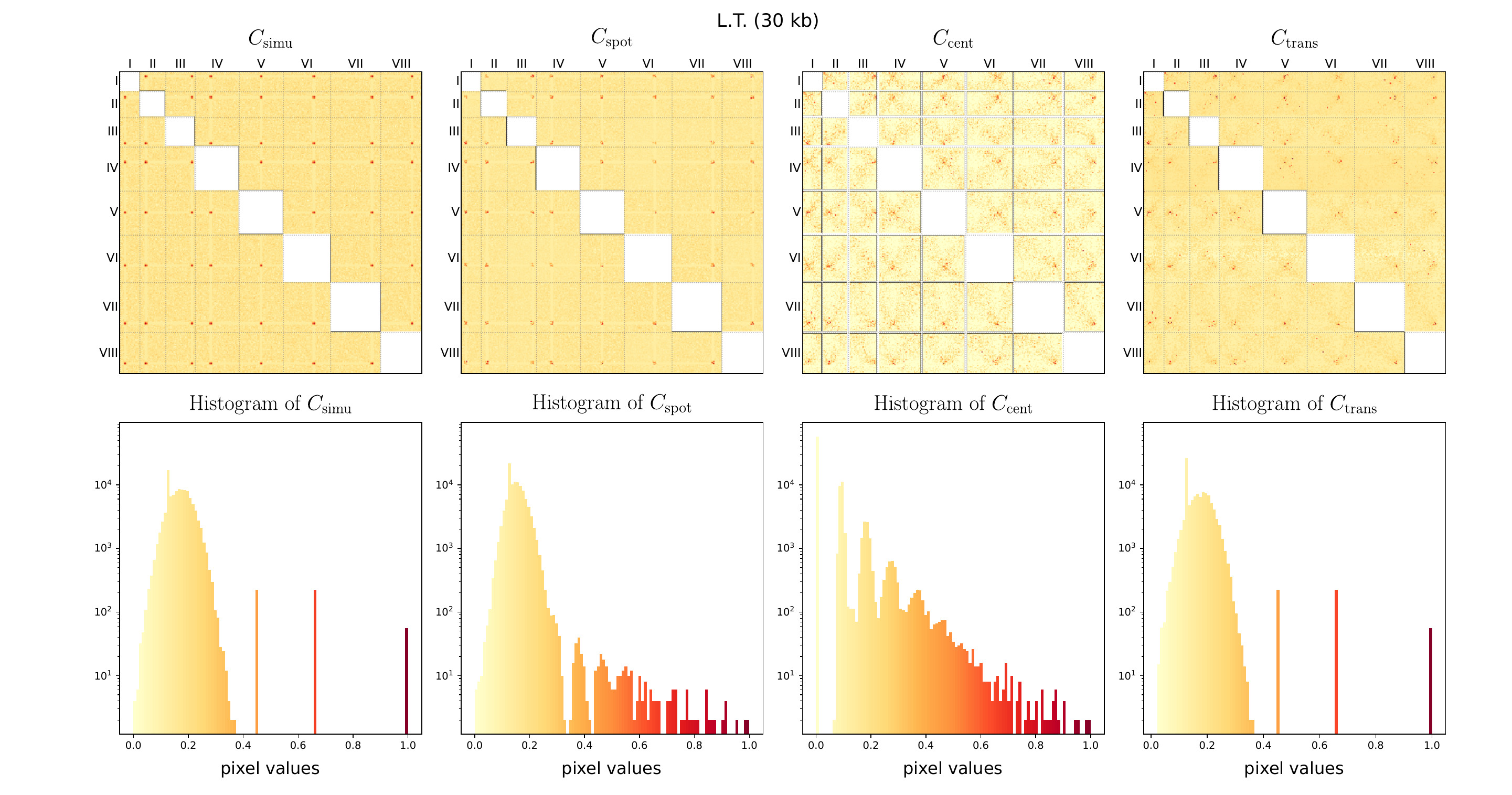}
\vspace{-4mm}
\includegraphics[width=\textwidth]{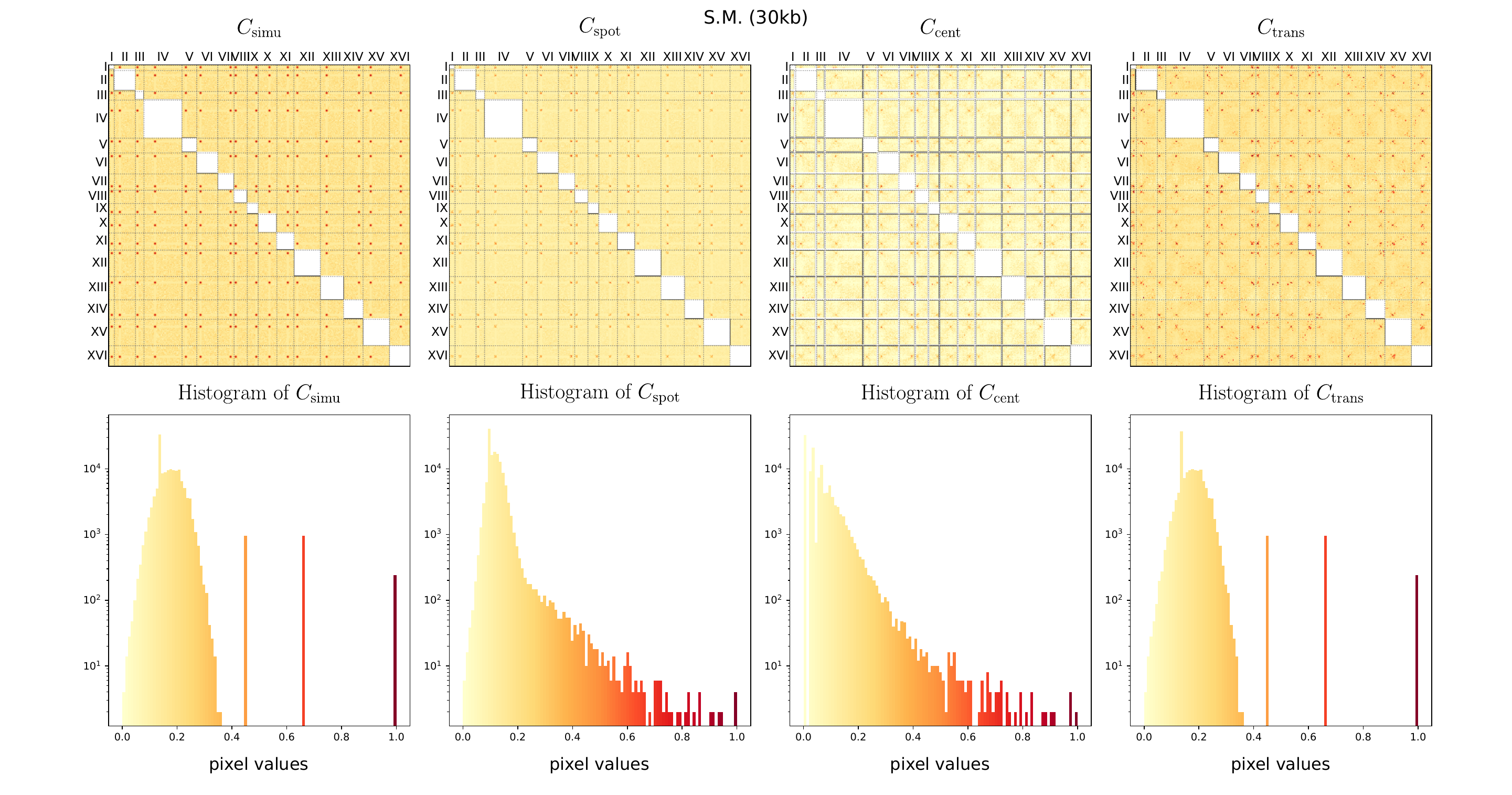}
\caption{\small
(top) Contact maps of yeasts S.C., L.T., and S.M., shown in the following order: simulated map $C_{\text{simu}}$, map with real spots and simulated noise $C_{\text{spot}}$, real map without telomere interactions $C_{\text{cent}}$, and real map transported to the closest simulated one $C_{\text{trans}}$. 
(bottom) Histograms of pixel-normalized values.
}
\label{fig:mismatch}
\end{figure}

\begin{table}[H]
\centering
\caption{\small 
Mismatch versus inference performance of \textit{BlockFormer} across three species of yeast. We report the normalized error, runtime (in seconds), and correlation with the simulated contact map $C_{\text{simu}}$. \textit{BlockFormer} achieves similar performance across settings, illustrating that simulator mismatch does not significantly impact results.
}
\label{tab:results}
\begin{tabular}{l l c c c}
\toprule
\textbf{Species} & \textbf{Method} & \textbf{Norm. error} & \textbf{Time (s)} & \textbf{Corr. with $C_\text{simu}$} \\
\midrule
\multirow{5}{*}{S.C. (16 chr.)}  
& $C_\text{simu}$  & 0.30 & 0.32 & -- \\
& $C_\text{raw}$  & 0.30 & 0.35 & 0.49 \\
& $C_\text{spot}$  & 0.41 & 0.40 & 0.49 \\
& $C_\text{cent}$  & 0.24 & 0.32 & 0.37 \\
& $C_\text{trans}$ & 0.30 & 0.30 & 0.58 \\
\midrule
\multirow{5}{*}{L.T. (8 chr.)}  
& $C_\text{simu}$  & 0.60 & 0.23 & -- \\
& $C_\text{raw}$  & 1.25 & 0.27 & 0.11 \\
& $C_\text{spot}$  & 0.66 & 0.28 & 0.11 \\
& $C_\text{cent}$  & 1.25 & 0.23 & 0.11 \\
& $C_\text{trans}$ & 0.84 & 0.27 & 0.37 \\
\midrule
\multirow{5}{*}{S.M. (16 chr.)} 
& $C_\text{simu}$  & 0.36 & 0.35 & -- \\
& $C_\text{raw}$  & 1.63 & 0.40 & 0.28 \\
& $C_\text{spot}$  & 0.43 & 0.31 & 0.27 \\
& $C_\text{cent}$  & 0.58 & 0.30 & 0.25 \\
& $C_\text{trans}$ & 0.57 & 0.47 & 0.46 \\
\bottomrule
\end{tabular}
\end{table}

The correlations are quite low (e.g. 0.11 for L.T.) and the distributions are very different (both the distribution of noise and the distribution of the spots) when analyzing the histograms: the spots values distribution in simulated maps are discrete whereas it is continuous in real maps. If we look at the inference performance on $C_\text{raw}$, we have a degradation of at most a factor 5, mostly due to high telomere interactions. If we remove those outliers as in $C_\text{cent}$ (which is done in \textit{Centurion}), we then have a degradation in the normalized error of at most a factor 2 compared to $C_\text{simu}$ (worst case: 1.25 versus 0.60 for L.T.). The mismatch degrades performance when working directly with the raw reference matrix: in particular, when there are strong telomeric interactions. However, since telomeres represent boundaries of interaction blocks, they can be removed (as done in \textit{Centurion}), and in such cases, the simulator mismatch no longer significantly impact inference accuracy.\\

Actually, the cases in which telomeric interaction patterns are prominent is a general challenge for centromere-localization methods rather than one specific to \textit{BlockFormer}. For example, \textit{Centurion} preprocesses contact maps by removing border regions to reduce potential telomeric effects. To further illustrate this challenge, we generate 100 synthetic maps based either on the L. thermotolerans (L.T.) or S. cerevisiae (S.C.) genomes and introduced telomeric interaction spots with sizes and intensities ranging from 0.25 to 2 times the size of the centromeric spots. We then evaluate \textit{BlockFormer} on both the original contact maps and maps in which the border regions were set to 0, following the pre-processing strategy used by \textit{Centurion}. The results are reported in the tables below.

\begin{table}[H]
    \centering
    \caption{\small Normalized absolute error averaged over 100 synthetic maps from the
    \textit{L. thermotolerans} (L.T.) genome.}
    \label{tab:synthetic_lt}
    \setlength{\tabcolsep}{4pt}
    \small
    \begin{tabular}{lccccccccc}
        \toprule
         &
        \multicolumn{9}{c}{\textbf{ratio telomere/centromere spot (size, intensity)}} \\
        & 0 & (0.25, 0.25) & (0.25, 0.5) & (0.25, 1)
        & (0.5, 0.25) & (0.5, 0.5) & (0.5, 1)
        & (1, 1) & (1, 2) \\
        \midrule
        Raw map &
        0.57 & 0.74 & 0.73 & 0.75 & 1.04 & 1.01 & 1.00 & 1.75 & 10.67 \\
        Border removed &
        -- & 0.56 & 0.55 & 0.55 & 0.53 & 0.52 & 0.52 & 0.61 & 2.20 \\
        \bottomrule
    \end{tabular}
\end{table}

\begin{table}[H]
    \centering
    \caption{\small Normalized absolute error averaged over 100 synthetic maps from the
    \textit{S. cerevisiae} (S.C.) genome.}
    \label{tab:synthetic_sc}
    \setlength{\tabcolsep}{4pt}
    \small
    \begin{tabular}{lccccccccc}
        \toprule
         &
        \multicolumn{9}{c}{\textbf{ratio telomere/centromere spot (size, intensity)}} \\
        & 0 & (0.25, 0.25) & (0.25, 0.5) & (0.25, 1)
        & (0.5, 0.25) & (0.5, 0.5) & (0.5, 1)
        & (1, 1) & (1, 2) \\
        \midrule
        Raw map &
        0.31 & 0.68 & 0.69 & 0.70 & 1.70 & 1.71 & 1.70 & 3.89 & 6.13 \\
        Border removed &
        -- & 0.34 & 0.34 & 0.34 & 0.49 & 0.48 & 0.50 & 1.13 & 3.44 \\
        \bottomrule
    \end{tabular}
\end{table}

The results show that prominent telomeric interactions can confound localization when their size approaches or exceeds that of the centromeric signal. However, when telomeric interactions are weaker than centromeric spots—as commonly observed in real datasets and represented by spot (size, intensity) ratios of 0.25-0.5 in our simulations—removing the border regions effectively mitigates this confounding effect.

Globally, our results indicate that accurate inference does not require highly realistic simulators, but rather capturing the key structural patterns shared across the considered interaction maps.

\subsection{Failure cases and solutions} 
\subsubsection{Too large chromosomes/high resolution: refinement step \label{sec:refine}}
\textit{BlockFormer} was trained on block size between $6$ and $62$ (see Appendix~\ref{sec:training_details_appendix}) and it has difficulties for block height far from this range. When species (e.g. A.T.) have too large chromosomes or when the resolution is too fine (e.g. $10$ kb), this leads to too big matrices and the inference fails. To address this problem, we propose an iterative refinement with multi-resolutions analysis: we start with a pre-localization step on a modified map at coarse resolution, ensuring that block sizes fall within the training range of block size. As in the pre-localization stage of \textit{Centurion}, the borders of each block in this map are set to $0$ to avoid bias caused by telomeres interactions. We then consider the raw map at a fine resolution where the block sizes are too big. We cut patches of maximal size $60 \times 60$ around the coarse parameter estimation in each trans-block and stack them for a refined estimation. The patches are centered around the coarse estimation in the fine map and the patch is clipped to the border of each trans-block (referred as $C_\text{refine}$).

We present refinement results on three species, for which the chromosome sizes are out of the training range. The yeast K.L. is studied at resolution $60$ kb then $30$ kb, the yeast S.P. is studied at resolution $90$ kb then $30$ kb and the plant A.T. is studied at resolution $600$ kb then $40$ kb.
We report the runtime as well as the normalized error: mean absolute error divided by the resolution (the smaller the better). Any value below 1 is satisfying (meaning an error below the resolution).
\begin{table}[H]
\centering
\caption{
Comparison between raw and refined interaction maps across species. We report the normalized error and runtime (in seconds). The refinement step substantially improves accuracy while maintaining low computational cost.
}
\label{tab:refine}
\begin{tabular}{l l c c}
\toprule
\textbf{Species} & \textbf{Method} & \textbf{Norm. error} & \textbf{Time (s)} \\
\midrule
\multirow{2}{*}{K.L. (6 chr.)}  
& $C_\text{raw}$  & 5.74 & 0.23 \\
& $C_\text{refine}$  & 1.60 & 0.31 \\
\midrule
\multirow{2}{*}{S.P. (3 chr.)}  
& $C_\text{raw}$  & 35.02 & 0.88\\
& $C_\text{refine}$  & 4.30 & 0.18 \\
\midrule
\multirow{2}{*}{A.T. (5 chr.)} 
& $C_\text{raw}$  & 125.94 & 718.2 \\
& $C_\text{refine}$  & 3.70 & 0.38 \\
\bottomrule
\end{tabular}
\end{table}

\subsubsection{Map too noisy: downsampling procedures \label{sec:downsample_appendix}}
 If the contact map deviates strongly from the simulator distribution (e.g. excessive noise obscuring spots), we propose two approaches: (i) reduce resolution via downsampling to smooth the map (the method is denoted $C_\text{coarse}$) or (ii) construct a map closer to simulated one by extracting the enrichment regions and introducing background simulated noise (method denoted $C_\text{spot}$).\\
 To change the resolution of the map, we apply a downsampling procedure. Suppose we want to go from a fine resolution $r$ bp to a coarser resolution $k \times r$ bp. Starting from the high-resolution map, each pixel in the coarse map is obtained by aggregating (e.g. summing) the values of the corresponding $k \times k$ neighborhood in the fine map. This aggregation is performed independently within each chromosome block, ensuring that the original block structure is preserved.\\
  We evaluate performance via the normalized error on a noisy contact map from yeast S.K. (the smaller the better). Both approaches improve performance, with $C_\text{spot}$ achieving the lowest error at low computational cost.
\begin{table}[H]
\centering
\caption{
Comparison between noisy raw and smoothed interaction maps for species S.K.. We report the normalized error and runtime (in seconds). Changing the resolution substantially improves accuracy while maintaining low computational cost.
}
\label{tab:downsample}
\begin{tabular}{l l c c}
\toprule
\textbf{Species} & \textbf{Method} & \textbf{Norm. error} & \textbf{Time (s)} \\
\midrule
\multirow{3}{*}{S.K. (16 chr.)}  
& $C_\text{raw}$ ($30$ kb)  & 1.90 & 0.44 \\
& $C_\text{spot}$ ($30$ kb)  & 0.59 & 0.31 \\
& $C_\text{coarse}$ ($60$ kb)  & 0.74 & 0.30 \\
\bottomrule
\end{tabular}
\end{table}

\newpage
\section{Training data generation \label{sec:training_details_appendix}}
Our model can generalize to unseen configurations because training samples span a wide range of block numbers and sizes, forcing the network to learn representations invariant to both the number and the size of blocks. We detail here the training data generation process.\\
Since the training set must include interaction maps of varying sizes, we organize the data into batches for computational efficiency. In each batch, we create a synthetic genome composed of $2$ to $10$ chromosomes with sizes ranging from $2 \times 10^5$ to $2 \times 10^6$ bp. Both the number of chromosomes and their individual sizes are sampled uniformly. The entity for which the parameter is to be inferred is also selected uniformly. The resolution of the maps is set to $r = 32$ kb.\\
Let L denote the number of chromosomes, with sizes $\{l_i \}_{1 \leq i \leq L}$ and let j be the selected entity. Within a batch, all training maps share the same structure derived from this synthetic genome: a sequence of $L-1$ trans-blocks with size $(\frac{l_j}{r}, \frac{l_i}{r})$ (namely varying from $6$ to $62$).\\
Inside a batch, the generation procedure follows Algorithm~\ref{alg:simu}: each parameter $\theta$ is sampled from $\prod_i \mathcal{U}(1, l_i-1)$, the spot size is sampled from $\mathcal{U}(0.1,10)$ and Gaussian noise is added to each map with amplitude up to $10\%$ of the maximum value in the map. 
%The intensity $\alpha$ is fixed to $100$. 
The map is normalized between $0$ and $1$ and each block in the map is then $0$-padded such that its size is a multiple of the patch size.

\begin{figure}[H]
    \centering
   % First row: two figures side by side
    
    \includegraphics[width=\textwidth]{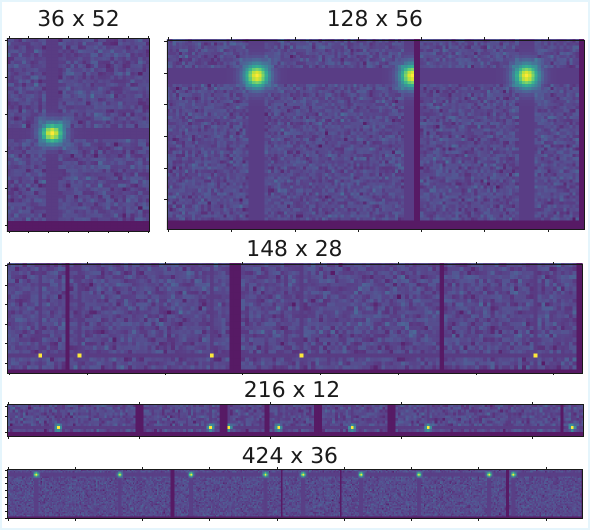}
    \caption{\small Examples of simulated maps, the number and the size of blocks vary. We provide sequences of trans-blocks created from synthetic genome of $2$, $4$, $6$, $8$ or $10$ chromosomes. The spots also vary in size ($\sigma^2$) and locations ($\theta$).
    }
\end{figure}

\newpage

\section{Ablations \label{sec:ablations_appendix}}
We demonstrate in that per-block modeling along with a highly various training set are necessary to obtain optimal performance in many settings. In the following ablations~\ref{sec:position_encoding_appendix} and~\ref{sec:various_train_appendix}, all the architectures are tested in the same setting: we consider maps that are sequences of $1$ to $14$ trans-blocks constructed from a synthetic genome made of $2$ to $15$ chromosomes of size ranging from $2 \times 10^5$ to $2 \times 10^6$ bp. For each number of blocks, we simulate $100$ maps as in Appendix~\ref{sec:training_details_appendix} where block sizes vary as well as parameter location and spot sizes . We evaluate performance by reporting the absolute error between the estimate $\hat{\theta}$ and $\theta$ normalized by the resolution.\\
We first provide comparison of \textit{BlockFormer} with other learning-based architectures.

\subsection{Comparison of \textit{BlockFormer} to others architectures \label{sec:archi_comparaison}}
\subsubsection{Comparison under architecture-specific training settings}
To the best of our knowledge, no modern baseline approaches such as graph-based or attention-based architectures are available for centromere inference from interaction maps except our prior work 
in 
\cite{etouron} 
%Appendix~\ref{sec:workshop} 
that proposes a probabilistic framework which uses a CNN-based architecture as summary statistic. We include this architecture in our study. We also tried several architectures including attention-based architectures to estimate the parameters. As some architectures are flexible to any maps and some are specific, we will compare them on a single setting with a synthetic genome made of the 3 first chromosomes of the species S.C. at resolution $32$ kb. The reference map of S.C. including only interactions between the three firsts chromosomes is denoted $C_\text{ref}$. We test the performance of each architecture with the normalized mean absolute error, either averaged over $1000$ synthetic maps $C_{\text{simu}}$ or on the reference map $C_{\text{ref}}$.\\
We designed architectures that takes as input: \begin{itemize}
    \item the entire map $C$ with only the upper trans-blocks and the lower symmetric part set to 0 but those methods are trained on maps with fixed number and size of blocks. Moreover, we rapidly face the curse of dimensionality when training those architectures on maps with increasing size (methods referred as entire $C$).\\
    Those models were trained on synthetic maps with same structure as $C_\text{ref}$.
    \item row of all blocks for each chromosome including the cis-block set to 0 but those methods are trained on fixed size row with a fixed number of blocks per chromosomes (methods referred as fixed row of $C$) .\\
    Those models were trained on rows with same structure as those of $C_\text{ref}$.
    
    \item row of all blocks for each chromosome including the cis-block set to 0 but those methods are trained on rows of variable sizes with a fixed number of blocks (methods referred as variable row of $C$).\\
    Those models were trained on rows of $3$ blocks of variable sizes.
    \item various sequences of trans-blocks but those methods are trained on various numbers of blocks with various sizes (methods referred as variable blocks of $C$).\\
    Those models were trained on sequences of $1$ to $9$ trans-blocks of variable sizes. 
\end{itemize} 
All the architectures that use a convolutional network (CNN) have the following backbone: 2 convolutional layers (1 to 6 to 12 channels with $3 \times 3$ kernels) with interleaved max-pooling (with $2 \times 2$ kernels).\\
The MLPs used have a factor-4 reduction to a single unit and a sigmoid activation to ensure a valid output.\\
The attention pooling (att-pool) is an MLP (12 to 64 to 1 channels).\\
The transformers (transf) used have 4 blocks, with 4 heads of attention, a patch size of 4 and an embedding dimension of 24.\\
All the architectures are trained on $5000$ training samples with batch size $200$ over $200$ epochs and a fixed learning rate of $5 \times 10^{-4}$ except the shared architectures based on transformers trained on $50000$ samples because of the variety in the training data and the number of model parameters.
\begin{table}[H]
\centering
\caption{Comparison of architectures across different input configuration. Normalized mean absolute error reported for simulated ($C_{\text{simu}}$) and reference ($C_{\text{ref}}$) maps.}
\label{tab:neurips_results}
\small
\resizebox{\textwidth}{!}{
\begin{tabular}{llccccc}
\toprule
\textbf{Input} 
& \textbf{Method} 
& \makecell{\textbf{Error on} \\ \textbf{$C_{\text{simu}}$}} 
& \makecell{\textbf{Error on} \\ \textbf{$C_{\text{ref}}$}} 
& \makecell{\textbf{Generalization}} \\
\midrule

\multirow{3}{*}{entire $C$}
& CNN+MLP                        & 0.35 & 0.37 & specific to map structure \\
& CNN+att-pool+MLP         &  0.95 & 3.11 & flexible to any $3 \times 3$ per-block map\\
& Transf+MLP                 & 0.44 & 0.26 & specific to map structure   \\

\midrule

\multirow{3}{*}{fixed row of $C$}
& CNN shared+MLP per chr.               & 0.29 & 0.49 & specific to map row structure\\
& Transf+MLP (per chr.)                & 0.41 & 0.60 & specific to map row structure\\
& Transf+cls token (per chr.)         & 0.46 & 1.06 & flexible to any row \\

\midrule

\multirow{2}{*}{variable row of $C$}
& CNN+att-pool+MLP (shared) & 0.58 & 2.54 & flexible to any row\\
& Transf+cls token (shared)  & 0.40 & 0.86 & flexible to any row\\

\midrule

\multirow{1}{*}{variable blocks of $C$}
& \textbf{BF : Transf+cls token (shared)}  & \textbf{0.52} & \textbf{1.34} & \textbf{flexible to any sequence of blocks}\\

\bottomrule
\end{tabular}
}
\end{table}
Architectures specific to map structure perform better than flexible architectures (e.g. CNN+MLP versus Transf+cls token). However, those architectures must be retrained when the map changes, rendering those architectures inefficient and the approach too costly compared to \textit{Centurion}. Among the flexible structures, the transformer-based approach outperforms the other attention-based approaches (Transf+cls token versus CNN+att-pool+MLP) (see below for a further study). Finally, presenting full rows of blocks containing the cis-block set to $0$ that could mislead inference seems less appropriate than variable sequences of trans-blocks. 
\subsubsection{Comparison of flexible architectures under a unified training setting}
In the previous section, to account for the differences in flexibility across architectures, we first compare \textit{BlockFormer} with other architectures where each model was trained using an appropriate configuration for its design and capabilities. We now compare three architectures designed to generalize to varying numbers and sizes of blocks :  \begin{itemize}
    \item the CNN with attention pooling (referred as CNN+att-pool+MLP)
    \item the classical Vision Transformer (referred as ViT) with standard 2D positional encoding
    \item \textit{BlockFormer} (referred as BF) with the per-block modeling strategy
\end{itemize}
All the architectures are trained under the same conditions described in Section~\ref{sec:train_details} and Appendix~\ref{sec:training_details_appendix}. We report their performance on the synthetic benchmark described at the beginning of the section and on real datasets (results are averaged over 5 random seeds) in the tables below.

\begin{table}[H]
\centering
\caption{\textbf{Synthetic case}: Normalized absolute error comparison across methods for low/mid regime (1, 4 blocks). Results are reported as mean $\pm$ std, median, and 95\% CI.}
\label{tab:ablation_split_1_5_norm}
\small
\begin{tabular}{c ccc ccc}
\toprule
& \multicolumn{3}{c}{\textbf{1 Block}} 
& \multicolumn{3}{c}{\textbf{4 Blocks}} \\
\textbf{Method} 
& Mean$\pm$Std & Median & 95\% CI
& Mean$\pm$Std & Median & 95\% CI \\
\midrule

\textbf{BF} 

& \textbf{0.43 $\pm$ 0.32} & 0.39 & [0.02, 1.22]
& \textbf{0.39 $\pm$ 0.30} & 0.33 & [0.02, 1.08]  \\

\textbf{ViT} 
& 0.64 $\pm$ 2.41 & 0.33 & [0.02, 1.20]
& 0.39 $\pm$ 0.34 & 0.31 & [0.01, 1.19] \\

\textbf{CNN+att-pool+MLP} 
& 3.68 $\pm$ 4.11 & 1.87 & [0.07, 13.63]
& 3.00 $\pm$ 3.69 & 1.57 & [0.08, 12.64] \\

\bottomrule
\end{tabular}
\end{table}

\begin{table}[H]
\centering
\caption{\textbf{Synthetic case}: Normalized absolute error comparison across methods for high, extreme regimes (10, 14 blocks). Results are reported as mean $\pm$ std, median, and 95\% CI.}
\label{tab:ablation_split_10_14_norm}
\small

\begin{tabular}{c ccc ccc}
\toprule
& \multicolumn{3}{c}{\textbf{10 Blocks}} 
& \multicolumn{3}{c}{\textbf{14 Blocks}} \\
\textbf{Method} 
& Mean$\pm$Std & Median & 95\% CI
& Mean$\pm$Std & Median & 95\% CI \\
\midrule
\textbf{BF} 
& \textbf{0.39 $\pm$ 0.30} & 0.35 & [0.03, 1.13]
& \textbf{0.36 $\pm$ 0.29} & 0.29 & [0.01, 0.93] \\

\textbf{ViT} 
& 0.42 $\pm$ 0.39 & 0.34 & [0.03, 1.36] 
& 0.53 $\pm$ 0.45 & 0.41 & [0.02, 1.66] \\

\textbf{CNN+att-pool+MLP} 
& 3.30 $\pm$ 4.21 & 1.48 & [0.07, 14.67]
& 2.83 $\pm$ 3.62 & 1.15 & [0.11, 13.06]  \\

\bottomrule
\end{tabular}
\end{table}

\begin{table}[H]
\centering
\caption{\textbf{Real case}: Normalized absolute error comparison across methods (5 random seeds) for maps with variable number of blocks.}
\small
\begin{tabular}{llcccc}
\toprule
\textbf{Specie} & \textbf{Method} & \multicolumn{1}{c}{\textbf{1 block}} & \multicolumn{1}{c}{\textbf{4 blocks}} & \multicolumn{1}{c}{\textbf{10 blocks}} & \multicolumn{1}{c}{\textbf{14 blocks}} \\
\midrule
S.C. & \textbf{BF} & \textbf{0.32 $\pm$ 0.03} & \textbf{0.35 $\pm$ 0.02} & \textbf{0.44 $\pm$ 0.03} & \textbf{0.46 $\pm$ 0.02} \\
 & \textbf{ViT} & 0.48 $\pm$ 0.03 & 0.59 $\pm$ 0.04 & 0.61 $\pm$ 0.03 & 0.59 $\pm$ 0.02 \\
 & \textbf{CNN+att-pool+MLP} & 1.72 $\pm$ 0.14 & 2.10 $\pm$ 0.11 & 2.47 $\pm$ 0.08 & 2.49 $\pm$ 0.03 \\
\midrule
S.M. & \textbf{BF} & \textbf{0.76 $\pm$ 0.05} & \textbf{0.64 $\pm$ 0.09} & \textbf{0.92 $\pm$ 0.07} & \textbf{2.18 $\pm$ 0.11} \\
 & \textbf{ViT} & 0.88 $\pm$ 0.16 & 0.80 $\pm$ 0.24 & 1.68 $\pm$ 0.17 & 2.35 $\pm$ 0.09 \\
 & \textbf{CNN+att-pool+MLP} & 3.35 $\pm$ 0.13 & 3.64 $\pm$ 0.10 & 3.52 $\pm$ 0.06 & 3.54 $\pm$ 0.07 \\
\midrule
L.K. & \textbf{BF} & \textbf{1.57 $\pm$ 0.07} & \textbf{1.83 $\pm$ 0.16} & -- & -- \\
 & \textbf{ViT} & 2.09 $\pm$ 0.38 & 1.89 $\pm$ 0.31 & -- & -- \\
 & \textbf{CNN+att-pool+MLP} & 7.93 $\pm$ 0.14 & 8.94 $\pm$ 0.12 & -- & -- \\
\midrule
S.P. & \textbf{BF} & \textbf{0.47 $\pm$ 0.17} & -- & -- & -- \\
 & \textbf{ViT} & 0.74 $\pm$ 0.19 & -- & -- & -- \\
 & \textbf{CNN+att-pool+MLP} & 13.50 $\pm$ 1.39 & -- & -- & -- \\
\midrule
K.L. & \textbf{BF} & \textbf{0.65 $\pm$ 0.05} & 0.96 $\pm$ 0.03 & -- & -- \\
 & \textbf{ViT} & 0.85 $\pm$ 0.05 & \textbf{0.64 $\pm$ 0.05} & -- & -- \\
 & \textbf{CNN+att-pool+MLP} & 1.97 $\pm$ 0.36 & 1.34 $\pm$ 0.16 & -- & -- \\
\midrule
P.F. (schizonts) & \textbf{BF} & \textbf{0.51 $\pm$ 0.04} & 0.67 $\pm$ 0.03 & \textbf{0.34 $\pm$ 0.03} & -- \\
 & \textbf{ViT} & 0.82 $\pm$ 0.03 & \textbf{0.52 $\pm$ 0.02} & 0.48 $\pm$ 0.01 & -- \\
 & \textbf{CNN+att-pool+MLP} & 1.78 $\pm$ 0.13 & 1.61 $\pm$ 0.09 & 1.76 $\pm$ 0.06 & -- \\
\midrule
\bottomrule
\end{tabular}
\end{table}

\textit{BlockFormer} consistently achieves the best overall performance, producing the lowest mean errors, narrowest confidence intervals and smallest standard deviations across nearly all datasets. The Vision Transformer is a reliable alternative although its performance is less stable when only a small number of blocks are used. In contrast, the attention-based CNN consistently exhibits the poorest performance, particularly on larger maps (e.g. S.P.) and shows limited robustness to changes in the number of input blocks.
\subsection{Ablations: importance of block-aware architecture \label{sec:position_encoding_appendix}}
One of our main contributions is the design of a block-aware architecture consisting in per-block padding and per-block 3D positional encoding that preserves the block-wise structure of the map and that cannot be naturally included in CNN-based architectures. The transformer architectures and training setup used in each case are identical to those described in Section ~\ref{sec:train_details} Training details and Appendix~\ref{sec:training_details_appendix}. We consider the following ablations: \begin{itemize}
    \item no positional encoding with per-block padding (referred as no pos.)
    \item 1D positional encoding with per-block padding (referred as 1D pos.)
    \item 2D positional encoding with bottom-right padding (referred as 2D pos. pad.)
    \item 2D positional encoding with per-block padding (referred as 2D pos.)
    \item 2D per-block positional encoding with per-block padding (referred as 2D pos. per block)
    \item 3D per-block positional encoding with block index and per-block padding (referred as BF : 3D pos. per block)
\end{itemize}

\begin{table}[H]
\centering
\caption{Performance comparison across methods for mid/high regime (4, 10 blocks). Results are reported as mean $\pm$ std, median, and 95\% CI.}
\label{tab:ablation_pos}
\small
\begin{tabular}{c ccc ccc}
\toprule
& \multicolumn{3}{c}{\textbf{4 Blocks}} 
& \multicolumn{3}{c}{\textbf{10 Blocks}} \\
\textbf{Method} 
& Mean$\pm$Std & Median & 95\% CI
& Mean$\pm$Std & Median & 95\% CI \\
\midrule

\textbf{BF : 3D pos. per block} 
& \textbf{0.39 $\pm$ 0.30} & 0.33 & [0.02, 1.08]
& \textbf{0.39 $\pm$ 0.30} & 0.35 & [0.03, 1.13]\\

\textbf{2D pos. per block} 
& 0.40 $\pm$ 0.30 & 0.34 & [0.02, 1.04]
& 0.41 $\pm$ 0.29 & 0.36 & [0.02, 0.96]\\

\textbf{2D pos.} 
& 0.42 $\pm$ 0.33 & 0.36 & [0.01, 1.19]
& 0.41 $\pm$ 0.32 & 0.34 & [0.01, 1.09]\\

\textbf{2D pos. pad.} 
& 0.39 $\pm$ 0.34 & 0.31 & [0.01, 1.19]
& 0.42 $\pm$ 0.39 & 0.34 & [0.03, 1.36]\\

\textbf{1D pos.} 
& 0.52 $\pm$ 0.48 & 0.43 & [0.01, 1.26]
& 1.04 $\pm$ 0.92 & 0.73 & [0.04, 3.41]\\

\textbf{no pos.} 
& 4.18 $\pm$ 5.07 & 1.70 & [0.08, 17.88]
& 4.06 $\pm$ 5.16 & 1.32 & [0.05, 17.21]\\
\bottomrule
\end{tabular}
\end{table}

Results for low ($1$ block) and extreme regimes (out-of-training range: $14$ blocks) are in Table~\ref{tab:ablation_pos_main}.\\
The 1D positional encoding (1D pos.) is generally unstable except in the medium regime ($4$ blocks), and the no positional encoding (no pos.) is always unstable. This strengthen the fact to consider 2D or 3D positional encoding. 
The bottom-right padding method (2D pos. pad.) is the worst method among 2D positional encoding methods: it is highly variable especially in low regime (1 blocks) and extreme regime (14 blocks) indicating no consistent improvement with scale and reinforcing the need of per-block padding.\\
The 2D pos. with per-block padding method (2D pos.) is more robust but still have instabilities with degradation in extreme regime (14 blocks) and do not show improvement with scale.\\
The per-block 2D pos. with per-block padding (2D pos. per block) is the closest method of 3D pos. and performs better than 2D pos.. It is unstable in low regime (1 block) but stabilizes with scale with no degradation in extreme regime (14 block). This strengthens the fact to consider per-block positional encoding.\\
The per-block 3D positional encoding along with the per-block padding method (3D pos. per block) is the best overall method: it has lowest or near-lowest mean error most consistently with competitive or best median performance and is in general more stable (lower std in most settings).\\
All methods that do not consider per-block information (2D pos. or 2D pos. pad.) are stable in medium regime but become unstable in low/extreme regime. Both methods that incorporate per-block positional information (2D pos. per block or 3D pos. per block) stay stable at extreme regime. Adding the block index in positional encoding as in 3D pos. helps stabilize the method in all regimes.\\

\subsection{Ablations: importance of variety in training strategy \label{sec:various_train_appendix}}
We design a training strategy that enables as much flexibility as possible: the parameter location, the spot size but also the entity to infer, the size and the number of blocks vary. We show that this design is necessary to have a flexible architecture to many interaction maps. 

\subsubsection{Necessity of various block numbers during training \label{sec:train_various_blocks_appendix}}
Presenting maps with various numbers of blocks to the model is necessary to have an architecture flexible to many numbers of blocks. We consider the following ablations in the training set: \begin{itemize}
    \item the model is trained on maps of $2$ blocks (referred as 3 chr.)
    \item the model is trained on maps of $4$ blocks (referred as 5 chr.)
    \item the model is trained on maps of $6$ blocks (referred as 7 chr.)
    \item the model is trained on maps of $2$, $4$ or $6$ blocks (referred as 3-5-7 chr.)
\end{itemize}
\begin{table}[H]
\centering
\caption{Normalized absolute error comparison across methods for low/mid regime (1, 4 blocks). Results are reported as mean $\pm$ std, median, and 95\% CI.}
\label{tab:ablation_split_1_5_norm}
\small
\begin{tabular}{c ccc ccc}
\toprule
& \multicolumn{3}{c}{\textbf{1 Block}} 
& \multicolumn{3}{c}{\textbf{4 Blocks}} \\
\textbf{Method} 
& Mean$\pm$Std & Median & 95\% CI
& Mean$\pm$Std & Median & 95\% CI \\
\midrule

\textbf{BlockFormer} 
& \textbf{0.43 $\pm$ 0.32} & 0.39 & [0.02, 1.22]
& 0.39 $\pm$ 0.30 & 0.33 & [0.02, 1.08] \\

\textbf{3-5-7 chr.} 
& 0.47 $\pm$ 0.43 & 0.38 & [0.02, 1.47]
& \textbf{0.37 $\pm$ 0.29} & 0.28 & [0.03, 1.05] \\

\textbf{7 chr.} 
& 0.62 $\pm$ 0.54 & 0.49 & [0.02, 1.74]
& 0.39 $\pm$ 0.28 & 0.36 & [0.03, 1.06] \\

\textbf{5 chr.} 
& 0.89 $\pm$ 0.68 & 0.74 & [0.02, 2.42]
& 0.44 $\pm$ 0.68 & 0.30 & [0.03, 1.06] \\

\textbf{3 chr.} 
& 0.45 $\pm$ 0.36 & 0.40 & [0.05, 1.32]
& 0.76 $\pm$ 0.69 & 0.52 & [0.04, 2.38] \\

\bottomrule
\end{tabular}
\end{table}
\begin{table}[H]
\centering
\caption{Normalized absolute error comparison across methods for high, extreme regimes (10, 14 blocks). Results are reported as mean $\pm$ std, median, and 95\% CI.}
\label{tab:ablation_split_10_14_norm}
\small

\begin{tabular}{c ccc ccc}
\toprule
& \multicolumn{3}{c}{\textbf{10 Blocks}} 
& \multicolumn{3}{c}{\textbf{14 Blocks}} \\
\textbf{Method} 
& Mean$\pm$Std & Median & 95\% CI
& Mean$\pm$Std & Median & 95\% CI \\
\midrule
\textbf{BlockFormer} 
& \textbf{0.39 $\pm$ 0.30} & 0.35 & [0.03, 1.13]
& \textbf{0.36 $\pm$ 0.29} & 0.29 & [0.01, 0.93] \\

\textbf{3-5-7 chr.} 
& 1.13 $\pm$ 1.89 & 0.80 & [0.01, 3.15] 
& 1.17 $\pm$ 1.02 & 0.88 & [0.01, 3.71] \\

\textbf{7 chr.} 
& 1.22 $\pm$ 2.40 & 0.61 & [0.01, 5.96]
& 1.82 $\pm$ 2.80 & 0.97 & [0.03, 10.58] \\

\textbf{5 chr.} 
& 5.66 $\pm$ 7.81 & 2.51 & [0.17, 27.56]
& 9.47 $\pm$ 10.77 & 4.53 & [0.47, 34.95] \\

\textbf{3 chr.} 
& 2.12 $\pm$ 1.82 & 1.66 & [0.07, 6.36]
& 6.40 $\pm$ 10.14 & 2.58 & [0.09, 38.50] \\
\bottomrule
\end{tabular}
\end{table}

\begin{figure}[H]
    \centering
    \includegraphics[width=0.7\textwidth]{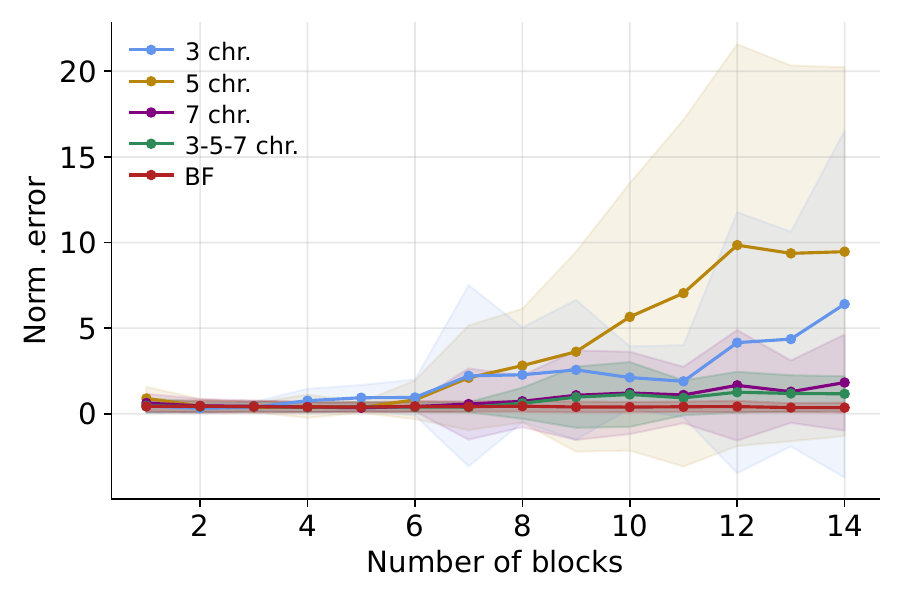}
    \caption{\small Mean normalized error with standard deviation. \textit{BlockFormer} (BF) outperforms all the others methods especially for high number of blocks, showing that maps with various number of blocks in the training set is necessary to maintain sub-resolution accuracy no matter the number of blocks (under 1).
    }
\end{figure}

\newpage

\subsubsection{Sensitivity to simulator parameters \label{sec:simu_sensitivity}}
The training set was designed to avoid any sensitivity to simulator parameters especially the parameter locations and spot sizes. We consider the following ablations in the training set:  \begin{itemize}
\item the spot size is fixed to $1.0$ in each map. (referred as fixed $\sigma$)
\item the maps have no background noise. (referred as no noise)
\end{itemize}
The training setting is identical as the one of \textit{BlockFormer} and detailed in Appendix~\ref{sec:training_details_appendix}.

\begin{table}[H]
\centering
\caption{Normalized absolute error comparison across methods for low/mid regime (1, 4 blocks). Results are reported as mean $\pm$ std, median, and 95\% CI.}
\label{tab:ablation_split_1_5_norm}
\small
\begin{tabular}{c ccc ccc}
\toprule
& \multicolumn{3}{c}{\textbf{1 Block}} 
& \multicolumn{3}{c}{\textbf{4 Blocks}} \\
\textbf{Method} 
& Mean$\pm$Std & Median & 95\% CI
& Mean$\pm$Std & Median & 95\% CI \\
\midrule

\textbf{BlockFormer} 
& \textbf{0.43 $\pm$ 0.32} & 0.39 & [0.02, 1.22]
& \textbf{0.39 $\pm$ 0.30} & 0.33 & [0.02, 1.08] \\

\textbf{fixed $\sigma$} 
& 1.39 $\pm$ 1.31 & 1.03 & [0.02, 4.54]
& 1.82 $\pm$ 2.41 & 1.10 & [0.07, 5.98] \\

\textbf{no noise} 
& 2.33 $\pm$ 2.64 & 1.29 & [0.03, 9.50]
& 3.08 $\pm$ 3.24 & 2.85 & [0.22, 7.02] \\

\bottomrule
\end{tabular}
\end{table}

\begin{table}[H]
\centering
\caption{Normalized absolute error comparison across methods for high, extreme regimes (10, 14 blocks). Results are reported as mean $\pm$ std, median, and 95\% CI.}
\label{tab:ablation_split_10_14_norm}
\small

\begin{tabular}{c ccc ccc}
\toprule
& \multicolumn{3}{c}{\textbf{10 Blocks}} 
& \multicolumn{3}{c}{\textbf{14 Blocks}} \\
\textbf{Method} 
& Mean$\pm$Std & Median & 95\% CI
& Mean$\pm$Std & Median & 95\% CI \\
\midrule
\textbf{BlockFormer} 
& \textbf{0.39 $\pm$ 0.30} & 0.35 & [0.03, 1.13]
& \textbf{0.36 $\pm$ 0.29} & 0.29 & [0.01, 0.93] \\

\textbf{fixed $\sigma$} 
& 2.05 $\pm$ 2.75 & 1.33 & [0.03, 9.74]
& 2.77 $\pm$ 4.02 & 1.67 & [0.07, 13.22] \\

\textbf{no noise} 
& 2.81 $\pm$ 2.71 & 2.13 & [0.09, 9.48]
& 3.63 $\pm$ 3.46 & 2.81 & [0.04, 11.74] \\

\bottomrule
\end{tabular}
\end{table}

\begin{figure}[H]
    \centering
    \includegraphics[width=0.7\textwidth]{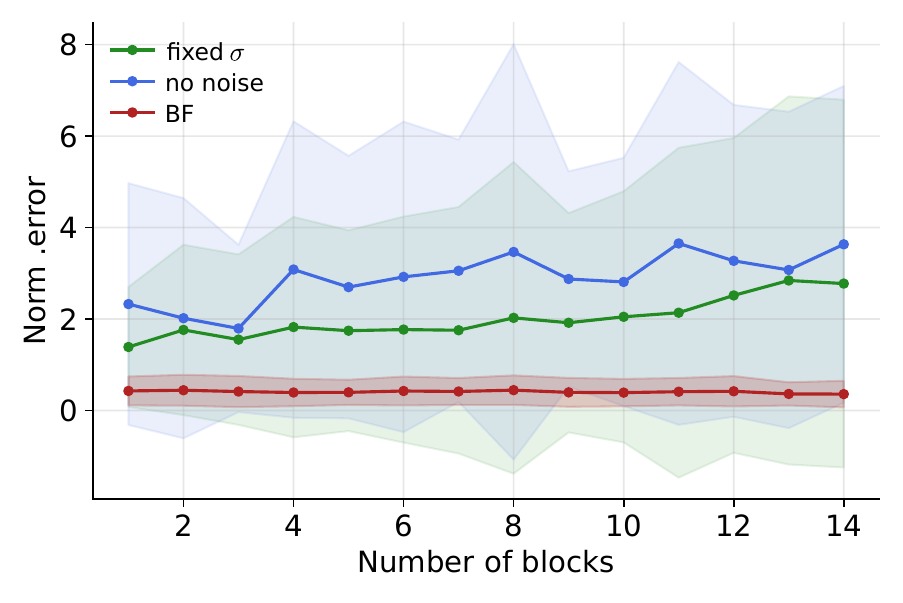}
    \caption{\small Mean normalized error with standard deviation. \textit{BlockFormer} (BF) outperforms all the others methods showing that adding noise and making spot size vary in the training set is necessary to have sub-resolution accuracy (under 1).
    }
\end{figure}
\newpage

\section{Experiments}

\subsection{A model flexible to various sizes of blocks}

\begin{figure}[H]
  \centering
  \includegraphics[width=0.5\linewidth]{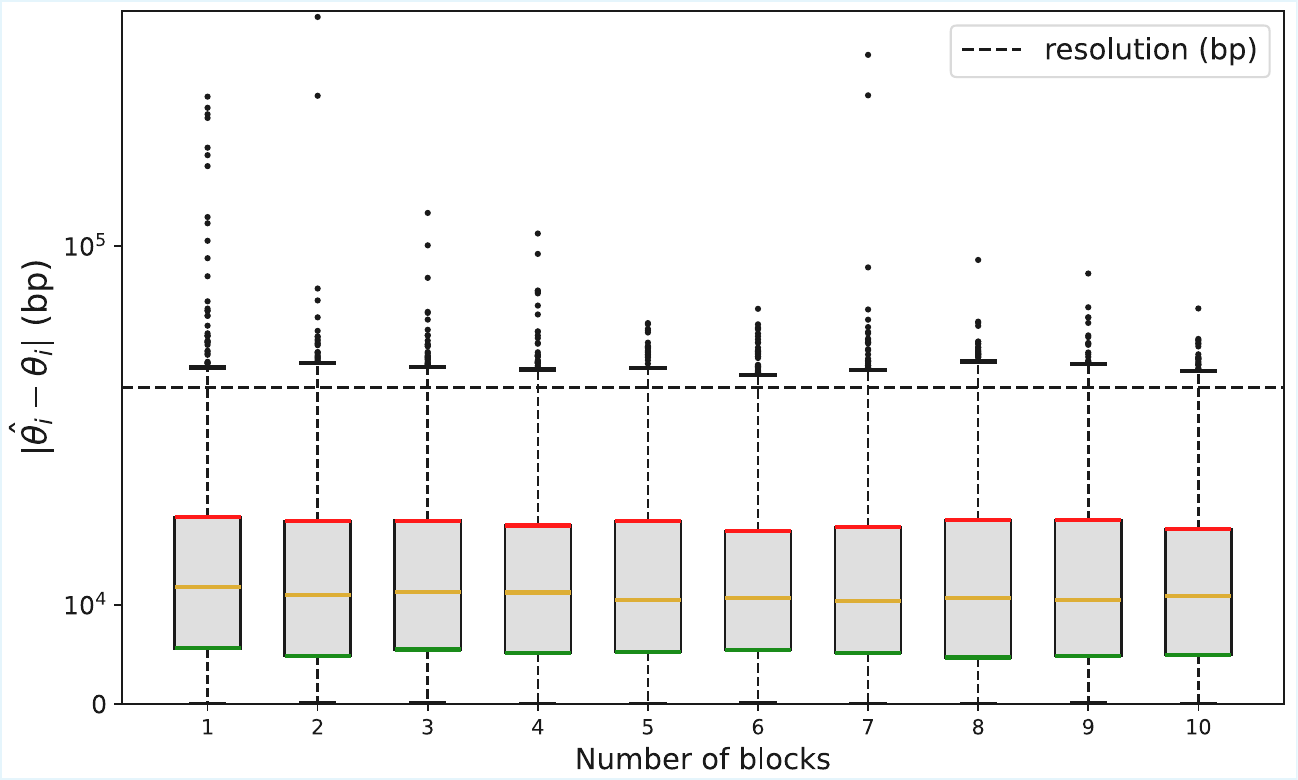}
  
  \caption{\small Boxplot of the absolute error between estimated $\hat{\theta}_i$ and ground truth $\theta_i$ averaged over $1\ 000$ synthetic contact maps per number of trans-blocks $k$, where $k$ varies from $1$ to $10$ .The maps are generated at resolution $32$~kb  from a synthetic genome of $k+1$ chromosomes, whose sizes vary from $2 \times 10^{5}$ bp to $2$ Mbp. At this resolution, each trans-block has a size varying between $6$ and $62$ pixels. In all the cases, more than $75\%$ of the parameter estimation errors lie under the resolution.}
  \label{fig:simu_variable}
\end{figure}

\subsection{A model flexible to various sequencing depths
\label{sec:seq_depth_appendix}}
We test the accuracy of \textit{BlockFormer} on $100$ synthetic maps generated from the reference genome of S.C. and also on the reference map of S.C., always keeping $10\%$ or $50\%$ of the sequencing depth. For each number of blocks $k$ and each chromosome $i$, we report in Figure~\ref{fig:seq_depth_synth_ref} the absolute error $err_i^k$, defined as follows: per number of blocks $k$, we make 10 random choices ($l=1, ..., 10$) of $k$ blocks of various sizes from the map (e.g. for $k=2$, we choose blocks $(2,9)$ (choice $l=1$), blocks $(1,7)$ (choice $l=2$), ...). Then, for target chromosome $i$, $k$ blocks, and map $n$, we compute $e^{k,n}_i = \frac{1}{10}\sum_{l=1}^{10} |\hat{\theta}_i^{n,k,l} - \theta_i|$. We report $err^k_i = \frac{1}{100} \sum_{n=1}^{100} e^{k,n}_i$.\\
Figure~\ref{fig:seq_depth_synth_ref} shows that the model generalizes well to different sequencing depths: for synthetic data, the accuracy does not change from $10\%$ to $50\%$ of the sequencing depth and stays under the resolution in most cases. Concerning the reference matrix, $50\%$ of sequencing depth does not change drastically the accuracy but at $10\%$ we start to see slight deterioration.

\newpage 

% --- First figure (Page 1) ---
\begin{figure}[H]
    \centering

    \includegraphics[width=0.8\textwidth, keepaspectratio]{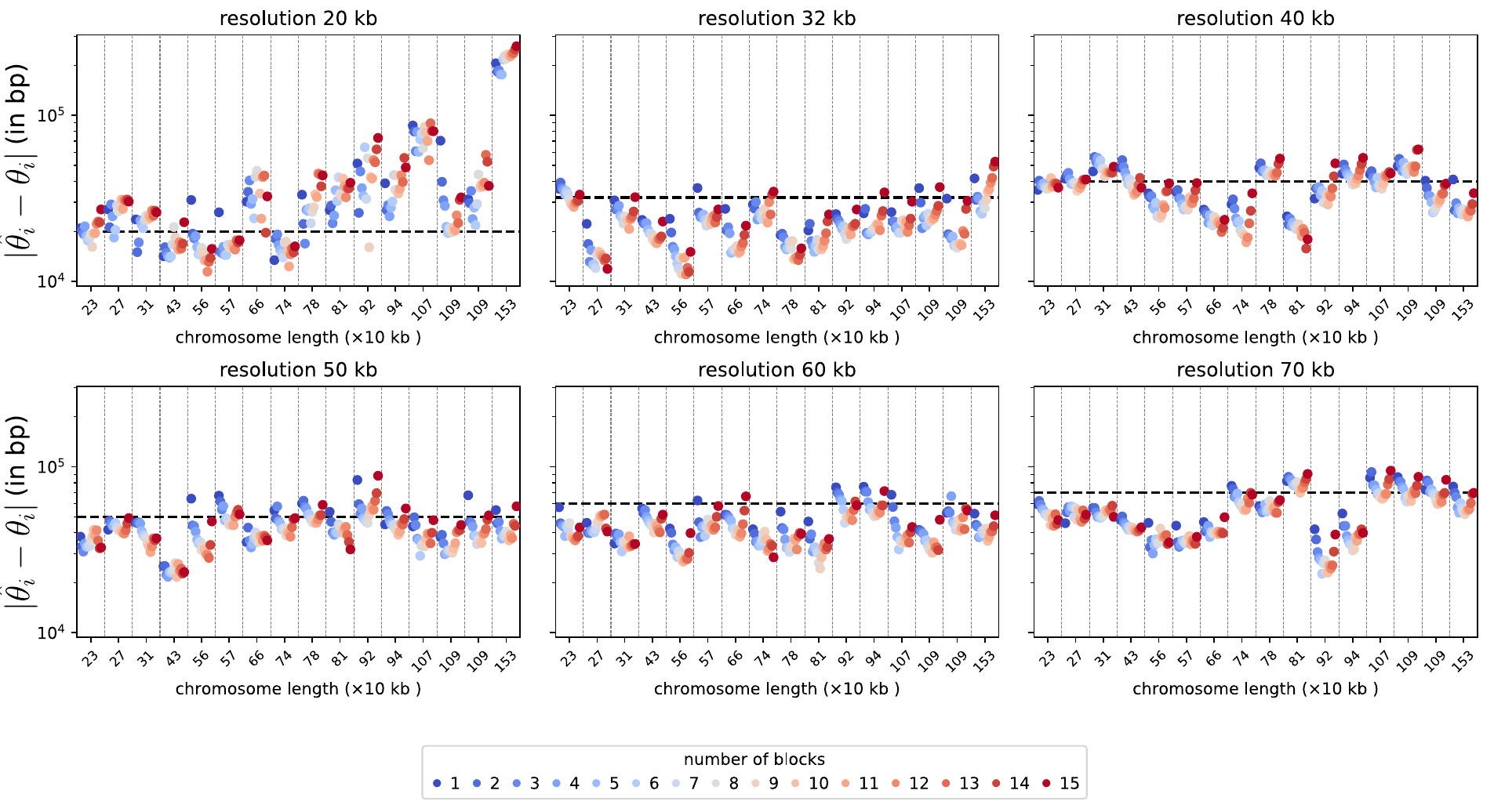}
    \caption*{(a) Synthetic maps, 10\% sequencing depth.}

    \vspace{0.3cm}

    \includegraphics[width=0.8\textwidth, keepaspectratio]{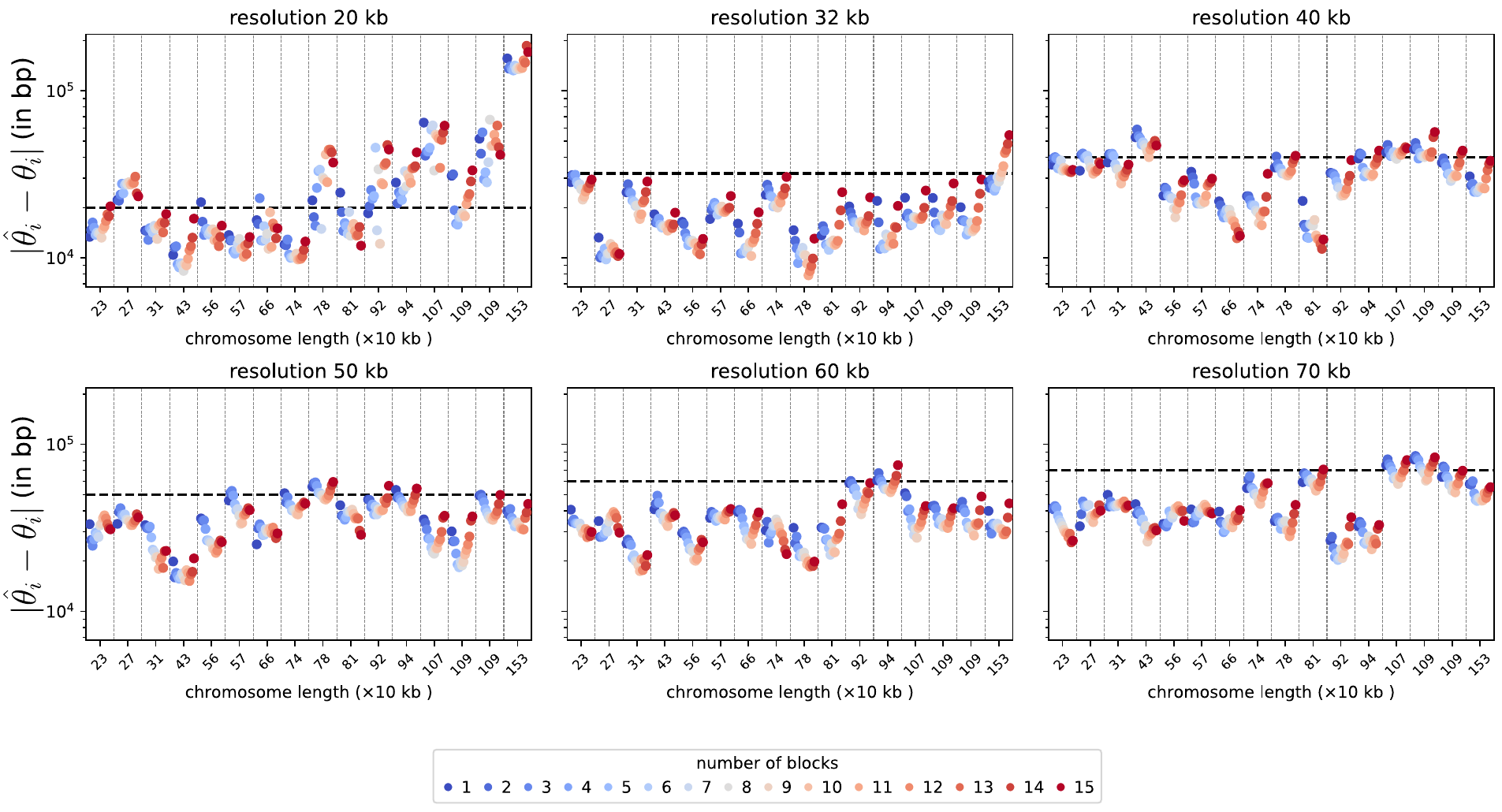}
    \caption*{(b) Synthetic maps, 50\% sequencing depth.}

\end{figure}

\clearpage % Force the next figure to start on a new page

% --- Second figure (Page 2) ---
\begin{figure}[H]
    \centering

    \includegraphics[width=0.8\textwidth, keepaspectratio]{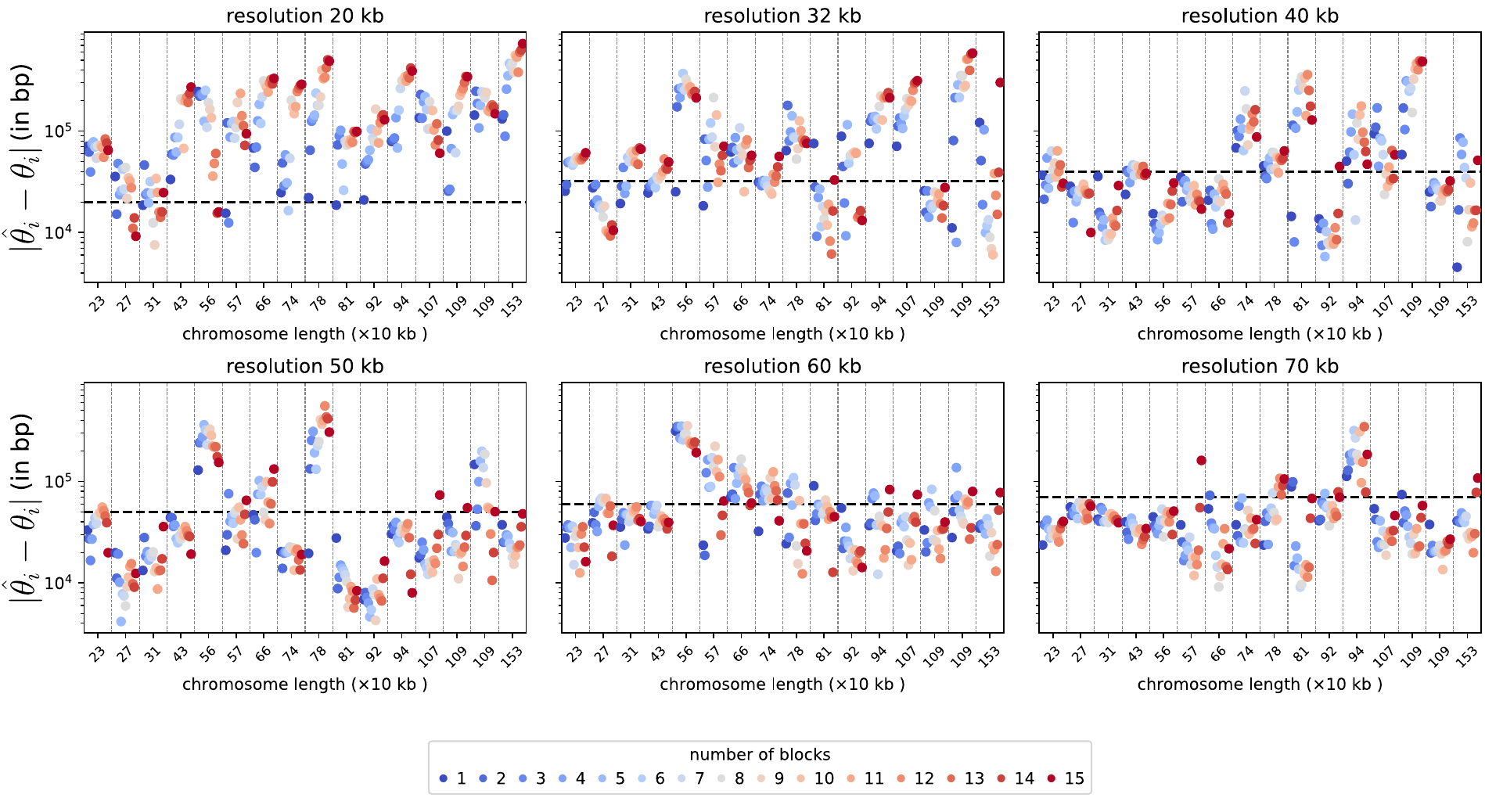}
    \caption*{(c) Reference map, 10\% sequencing depth.}

    \vspace{0.3cm}

    \includegraphics[width=0.8\textwidth, keepaspectratio]{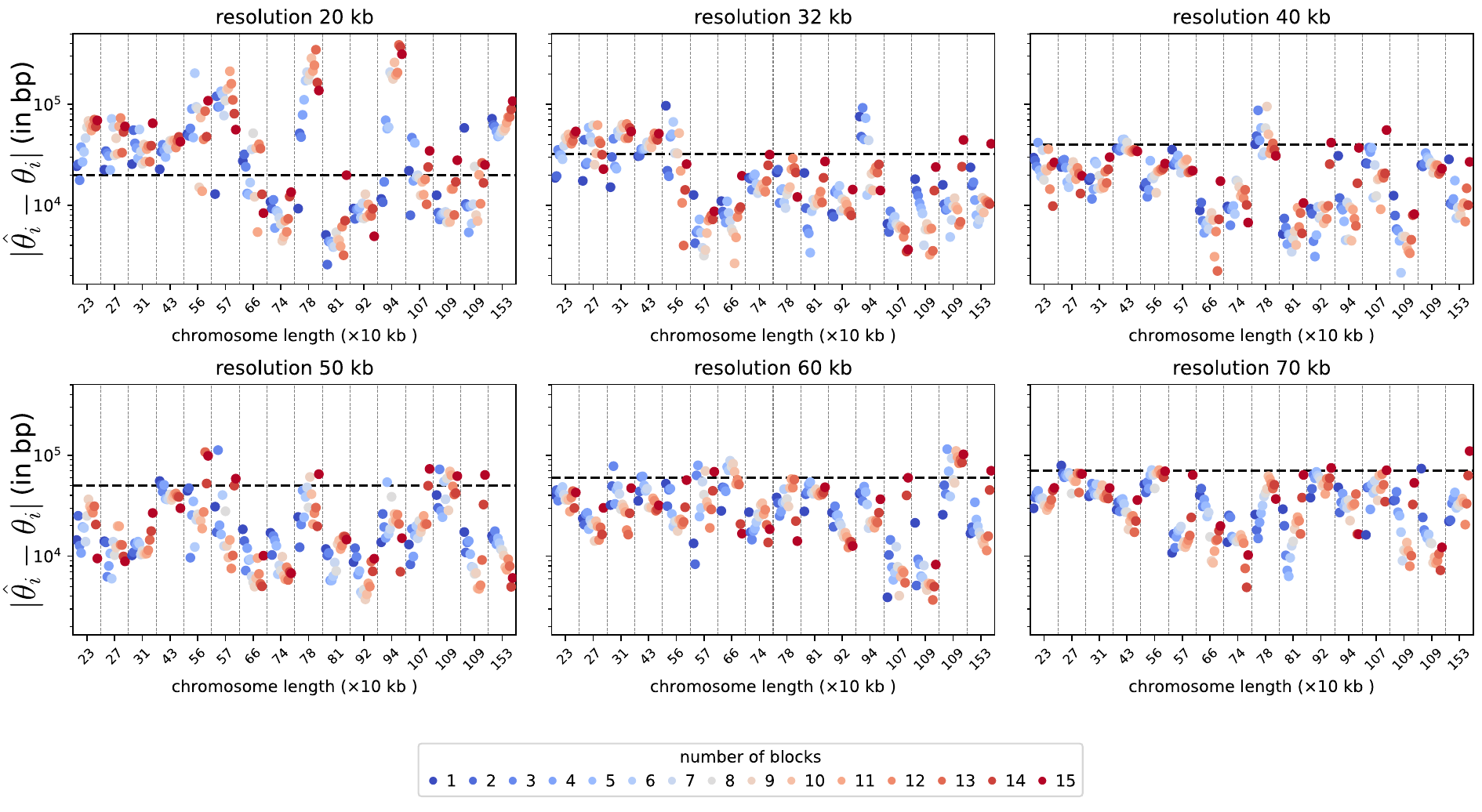}
    \caption*{(d) Reference map, 50\% sequencing depth.}

    \caption{\small Absolute error per centromere over $100$ synthetic contact maps generated from the S.C. genome \textbf{(a)} and \textbf{(b)} and one reference map \textbf{(c)} and \textbf{(d)} with $10\%$ or $50\%$ of sequencing depth. Chromosomes on the x-axis are sorted by length (bp). Color shades range from blue to red as the number of blocks $k$ increases from $1$ to $15$.}
    
    \label{fig:seq_depth_synth_ref}
\end{figure}

\newpage
\subsection{A model flexible to map normalization \label{sec:hic_norm_appendix}}
Hi-C contact maps are subject to various biases arising from experimental protocols as well as sequencing and mapping errors. To mitigate these effects, all real contact maps considered in the study are normalized using ICE, an iterative procedure that forces all rows and columns of the map to sum up to one \cite{hicnorm}. To normalize the reference contact maps, we use the ICE normalization via the Python library \texttt{iced} with the function \texttt{ICE\_normalization} from the module \texttt{normalization}.\\
In contrast, synthetic maps are free from experimental bias and therefore do not require normalization. 
We show that  \textit{BlockFormer} remains robust this difference in preprocessing by applying it to both normalized and unnormalized real maps. The results on several real-world cases are reported in the table below, with performances averaged over five random seeds.

\begin{table}[H]
\centering
\small
\caption{\small Raw maps (without ICE normalization).  Comparison of normalized error (the smaller the better) and time across species. (5 random seeds)}
\label{tab:raw_map_small}
\renewcommand{\arraystretch}{1.1}
\setlength{\tabcolsep}{2.5pt}
\resizebox{\linewidth}{!}{%
\begin{tabular}{cc cccccc}
\toprule
\multicolumn{2}{c}{\multirow{2}{*}{\textbf{Method}}} 
& \multicolumn{2}{c}{\textbf{S.C.}} 
& \multicolumn{2}{c}{\textbf{L.K.}} 
& \multicolumn{2}{c}{\textbf{L.T.}} \\
\cmidrule(r){3-4} \cmidrule(r){5-6} \cmidrule(r){7-8}
& & Err & Time & Err & Time & Err & Time \\
\midrule

\multirow{2}{*}{\rotatebox[origin=c]{90}{Init.}} 
& \textit{BlockFormer}
& 0.69 $\pm$ 0.01 & \textbf{0.41 $\pm$ 0.02}
& \textbf{1.54 $\pm$ 0.15} & \textbf{0.27 $\pm$ 0.08}
& 1.08 $\pm$ 0.07 & \textbf{0.26 $\pm$ 0.04} \\

& \textit{Centurion}
& 1.07 $\pm$ 0.00 & 2.28 $\pm$ 0.11
& 1.93 $\pm$ 0.00 & 0.73 $\pm$ 0.09
& \textbf{0.67 $\pm$ 0.00} & 1.05 $\pm$ 0.09 \\

\midrule

\multirow{2}{*}{\rotatebox[origin=c]{90}{Fitting}} 
& \textit{BlockFormer}
& \textbf{0.51 $\pm$ 0.00} & \textbf{2.60 $\pm$ 0.17}
& \textbf{0.17 $\pm$ 0.00} & \textbf{1.83 $\pm$ 0.14}
& \textbf{0.35 $\pm$ 0.00} & \textbf{1.73 $\pm$ 0.12} \\

& \textit{Centurion}
& \textbf{0.51 $\pm$ 0.00} & 133.57 $\pm$ 7.06
& \textbf{0.17 $\pm$ 0.00} & 10.59 $\pm$ 0.41
& \textbf{0.35 $\pm$ 0.00} & 3.70 $\pm$ 0.24 \\

\bottomrule
\end{tabular}%
}
\end{table}

\begin{table}[H]
\centering
\small
\caption{\small Raw maps (with some out-of-training species) (without ICE normalization).  Comparison of normalized error (the smaller the better) and time across species (5 random seeds)}
\label{tab:raw_map_big}
\renewcommand{\arraystretch}{1.1}
\setlength{\tabcolsep}{2.5pt}
\resizebox{\linewidth}{!}{%
\begin{tabular}{cc cccccccc}
\toprule
\multicolumn{2}{c}{\multirow{2}{*}{\textbf{Method}}} 
& \multicolumn{2}{c}{\textbf{S.M.}} 
& \multicolumn{2}{c}{\textbf{K.L.}} 
& \multicolumn{2}{c}{\textbf{S.P.}} 
& \multicolumn{2}{c}{\textbf{Z.R.}} \\
\cmidrule(r){3-4} \cmidrule(r){5-6} 
\cmidrule(r){7-8} \cmidrule(r){9-10}
& & Err & Time & Err & Time & Err & Time & Err & Time \\
\midrule

\multirow{2}{*}{\rotatebox[origin=c]{90}{Init.}} 
& \textit{BlockFormer}
& \textbf{0.68 $\pm$ 0.06} & \textbf{0.31 $\pm$ 0.01}
& 0.83 $\pm$ 0.02 & \textbf{0.31 $\pm$ 0.10}
& \textbf{3.27 $\pm$ 0.01} & 0.17 $\pm$ 0.04 & \textbf{0.83 $\pm$ 0.12} & \textbf{0.33 $\pm$ 0.07} \\

& \textit{Centurion}
& 0.87 $\pm$ 0.00 & 2.78 $\pm$ 0.02
& \textbf{0.67 $\pm$ 0.00} & 0.70 $\pm$ 0.12
& 23.47 $\pm$ 0.00 & \textbf{0.05 $\pm$ 0.01} & 5.89 $\pm$ 0.0 & 0.5 $\pm$ 0.05 \\

\midrule

\multirow{2}{*}{\rotatebox[origin=c]{90}{Fitting}} 
& \textit{BlockFormer}
& \textbf{0.19 $\pm$ 0.00} & \textbf{3.62 $\pm$ 0.19}
& \textbf{0.14 $\pm$ 0.00} & \textbf{2.29 $\pm$ 0.57}
& \textbf{1.15 $\pm$ 0.00} & \textbf{1.18 $\pm$ 0.06} & \textbf{0.35 $\pm$ 0.0} & \textbf{2.15 $\pm$ 0.26} \\

& \textit{Centurion}
& \textbf{0.19 $\pm$ 0.00} & 9.18 $\pm$ 0.08
& \textbf{0.14 $\pm$ 0.00} & 4.36 $\pm$ 0.84
& 32.25 $\pm$ 16.18 & 380.92 $\pm$ 176.41 & 2.17 $\pm$ 1.49 & 12.91 $\pm$ 1.28 \\

\bottomrule
\end{tabular}%
}
\end{table}

These results together with Tables \ref{tab:other_species_small} and \ref{tab:other_species_big}} demonstrate that \textit{BlockFormer} is robust to the presence or absence of normalization, maintaining consistent performance across both normalized and unnormalized real contact maps.

\newpage
\subsection{A model flexible to various spot patterns settings \label{sec:other_settings_appendix}}
We compare \textit{BlockFormer} with \textit{Centurion} in slightly different settings from the centromeres inference task. The spots have different shapes that can appear in realistic biological cases.\\
%In all the experiments, $\hat{\theta}$ is estimated in parallel component per component. For the transformer-based network, we consider $k=1$ to $15$ trans-blocks. Per number of blocks $k$, $10$ random subsets of $k$ trans-blocks are sampled (choices $l=1,...,10$). Parameter estimation $\hat{\theta}_i^k$ is performed in parallel for each chromosome $i$ with $\hat{\theta}_i^k = \frac{1}{10} \sum_{l=1}^{10} \hat{\theta}_i^{l, k}$. 
 In each setting, we generate $100$ synthetic maps at resolution $30$~kb based on the reference genome of S.C.. We report the mean absolute error between $\hat{\theta}$ and $\theta$ in bp. $\hat{\theta}$ is constructed as in Appendix~\ref{sec:parameter_construction}.

\subsubsection{Gaussian spots \label{sec:gaussian_appendix}}
\textit{Centurion} is fine-tuned to the centromere inference problem based on Hi-C maps. We first compare \textit{BlockFormer} with it in a synthetic setting with Gaussian spots of different sizes and locations. In this setting favorable to both methods, our network is much faster than \textit{Centurion} and leads to slightly similar accuracy: around $10$~kb when working at resolution $30$~kb.

\begin{figure}[H]
    \centering
    \includegraphics[width=0.7\linewidth]{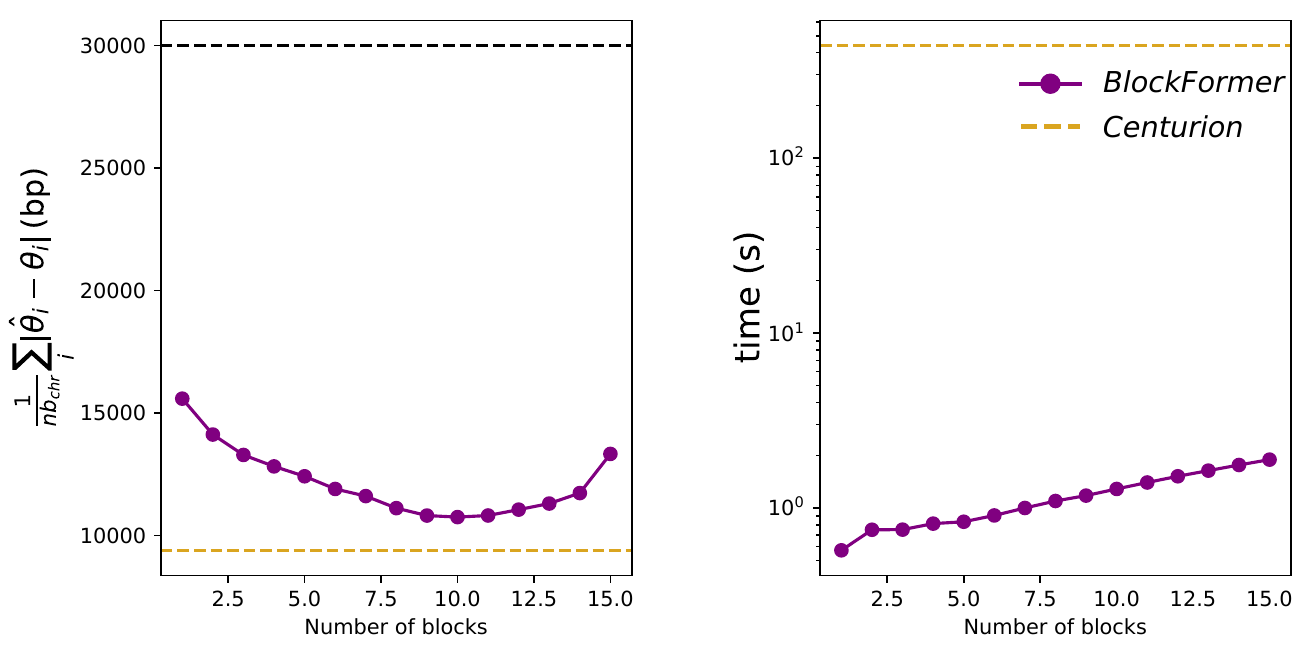}
    \caption{\small Mean absolute error (bp, left) and runtime (s, right) over $100$ synthetic maps generated from the S.C. genome at resolution $30$~kb. The spots in each trans-block are Gaussian. The black dotted line in the left plot represents the resolution of the maps.\label{fig:comparison_nelle_simu_gaussian}
    % 1 candidate time limmit 1000 time ptim 1 sigma_4 for centurion
    }
\end{figure}

\subsubsection{Square spots \label{sec:square_appendix}}
\begin{figure}[H]
    \centering
    \includegraphics[width=0.5\linewidth]{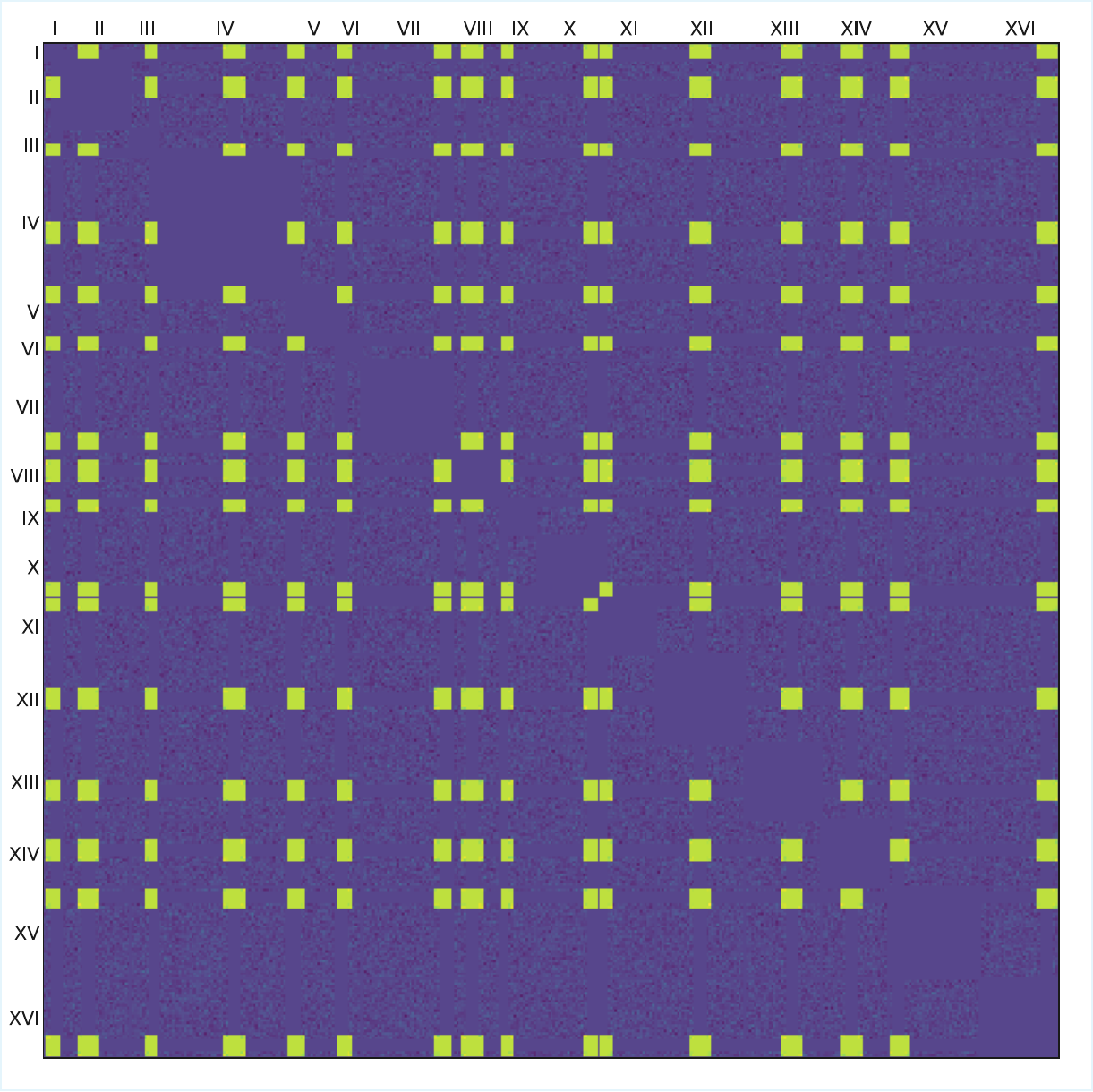}
    \caption{\small One synthetic map simulated from the S.C. reference genome with square spots in each trans-block.\label{fig:simu_carre}}
\end{figure}

\newpage
\subsubsection{Elliptical spots \label{sec:ellipse_appendix}}
In gene-dense regions, the chromatin is more dynamic leading to interactions fluctuation and anisotropic spots. To represent such situation, we generate maps with elliptical spots. 
%(see Figure~\ref{fig:simu_ellipse} and Figure~\ref{fig:comparison_nelle_simu_ellipse}).
In this setting, \textit{Centurion} is slightly more accurate than our model but remains slower.
\begin{figure}[H]
    \centering
    \includegraphics[width=0.5\linewidth]{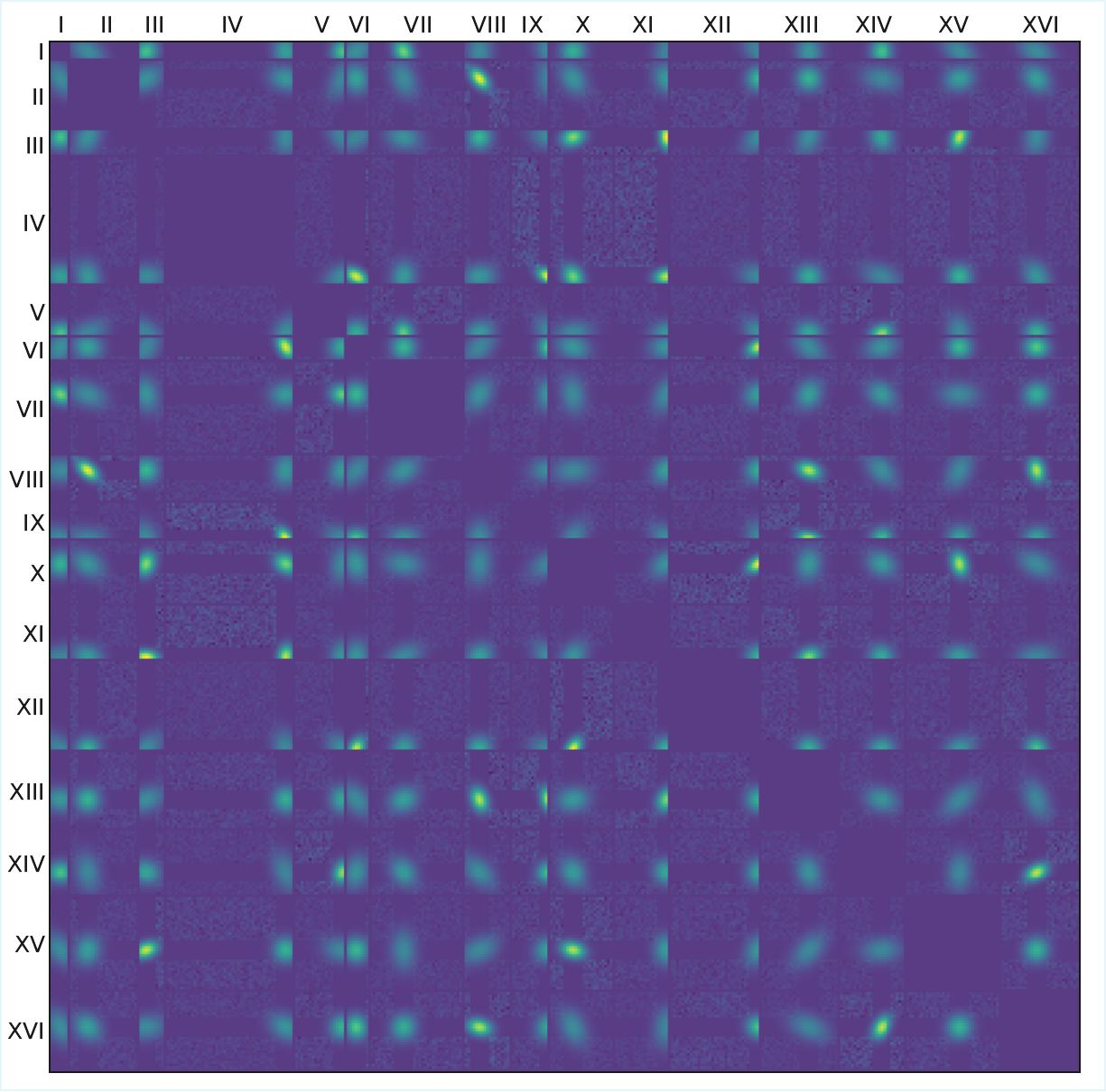}
    \caption{\small One synthetic map simulated from the S.C. reference genome with elliptical spots in each trans-block.\label{fig:simu_ellipse}}
\end{figure}
When the spots are elliptical, both methods output under-resolution parameter estimation. \textit{Centurion} is $1.5$ more accurate but more than $10$ times slower than \textit{BlockFormer}. 
\begin{figure}[H]
    \centering
    \includegraphics[width=0.7\linewidth]{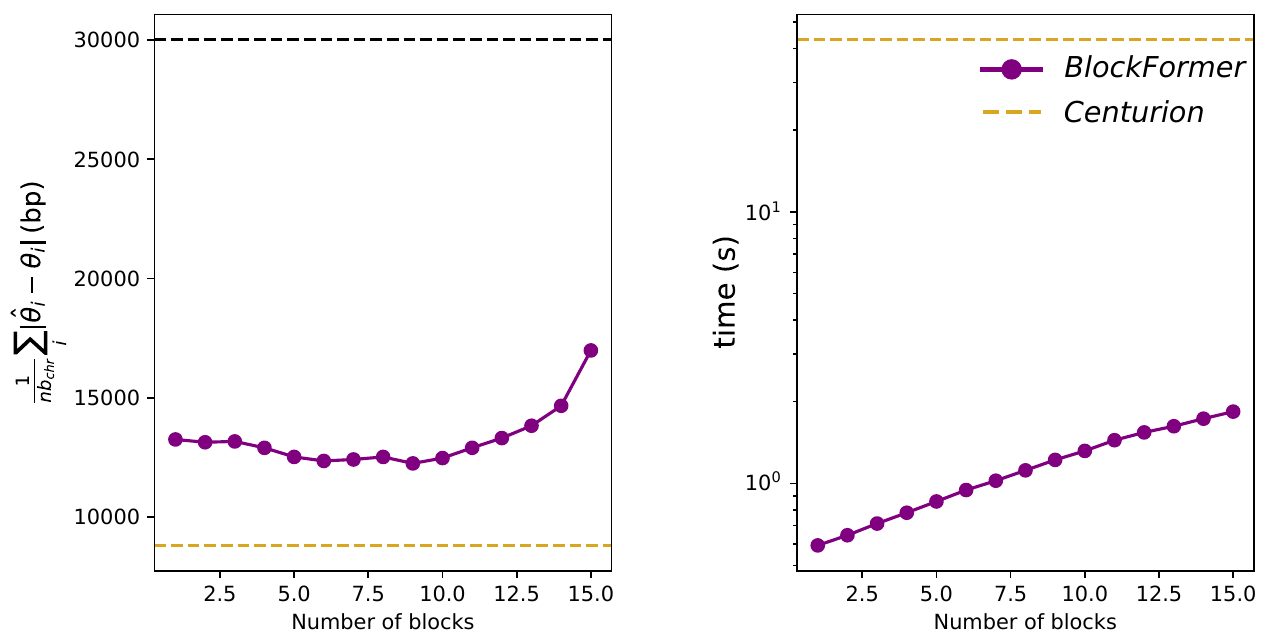}
    \caption{\small Mean absolute error (bp, left) and runtime (s, right) over $100$ synthetic maps generated from the S.C. genome at resolution $30$~kb. The spots in each trans-block are ellipse. The black dotted line in the left plot represents the resolution of the maps.\label{fig:comparison_nelle_simu_ellipse}
    % 1 candidate time limmit 1000 time ptim 1 sigma_4 for centurion
    }
\end{figure}

\newpage
\subsubsection{Ring spots \label{sec:ring_appendix}}
Interactions between entities can also happen at a preferred distance with an exclusion at center likely due to structural constraint (e.g. with protein complexes blocking direct contact). To mimic this scenario, we generate maps with ring-shaped spots. 
%(see Figure~\ref{fig:simu_ring} and Figure~\ref{fig:comparison_nelle_simu_ring}). 
When the spots are ring-shaped, \textit{Centurion} is more accurate than our method but remains too slow.
\begin{figure}[H]
    \centering
    \includegraphics[width=0.5\linewidth]{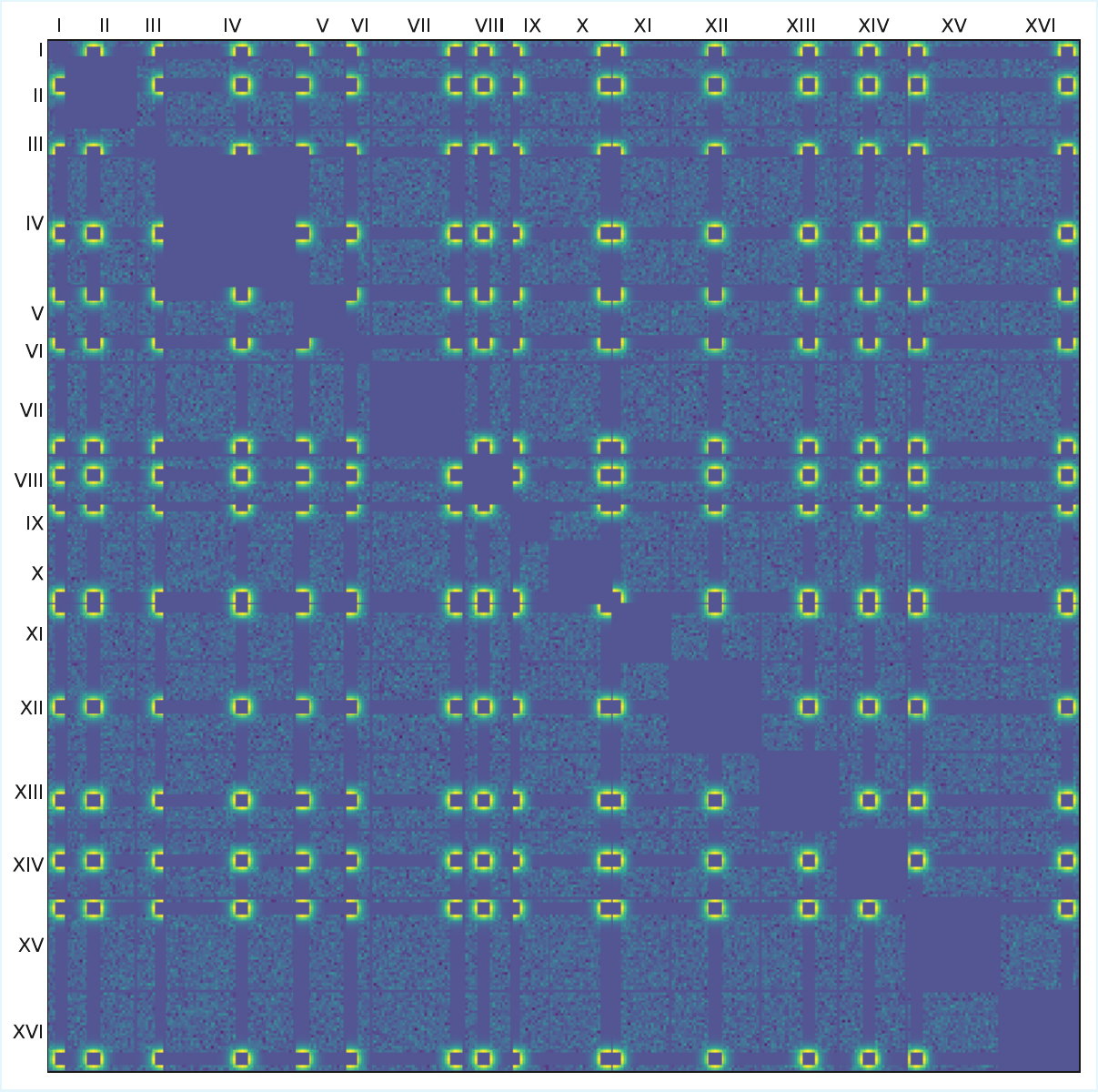}
    \caption{\small One synthetic map simulated from the S.C. reference genome with ring spots in each trans-block.\label{fig:simu_ring}}
\end{figure}
When the spots in the maps are rings instead of Gaussian spots, \textit{Centurion} has difficulties to output accurate results in reasonable time. Our approach is $1.2$ times less accurate than \textit{Centurion} since the accuracy is around twice the resolution but the runtime is nearly $10^2$ smaller.
\begin{figure}[H]
    \centering
    \includegraphics[width=0.7\linewidth]{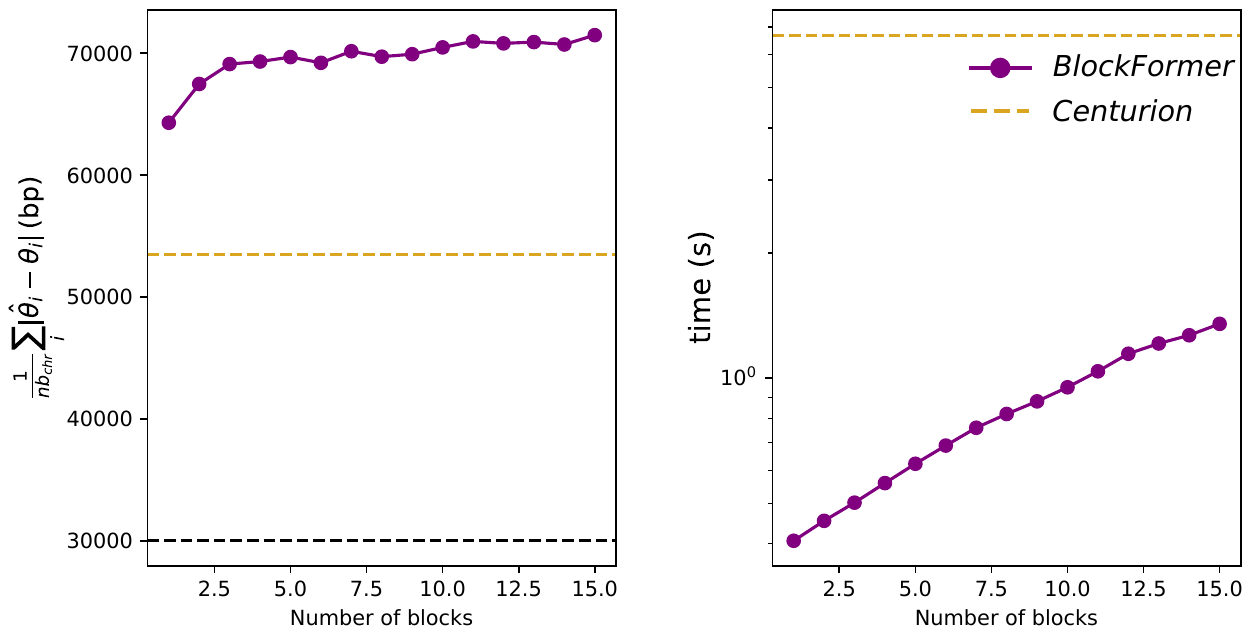}
    \caption{\small Mean absolute error (bp, left) and runtime (s, right) over $100$ synthetic maps generated from the S.C. genome at resolution $30$~kb. The spots in each trans-block are rings. The black dotted line in the left plot represents the resolution of the maps.\label{fig:comparison_nelle_simu_ring}
    % old results for the model + 1 candidate time limmit 1000 time ptim 10 sigma_4 for centurion
    }
\end{figure}
\newpage
\subsubsection{Multiple spots \label{sec:multiple_spots_appendix}}
\begin{figure}[H]
    \centering
    \includegraphics[width=0.5\linewidth]{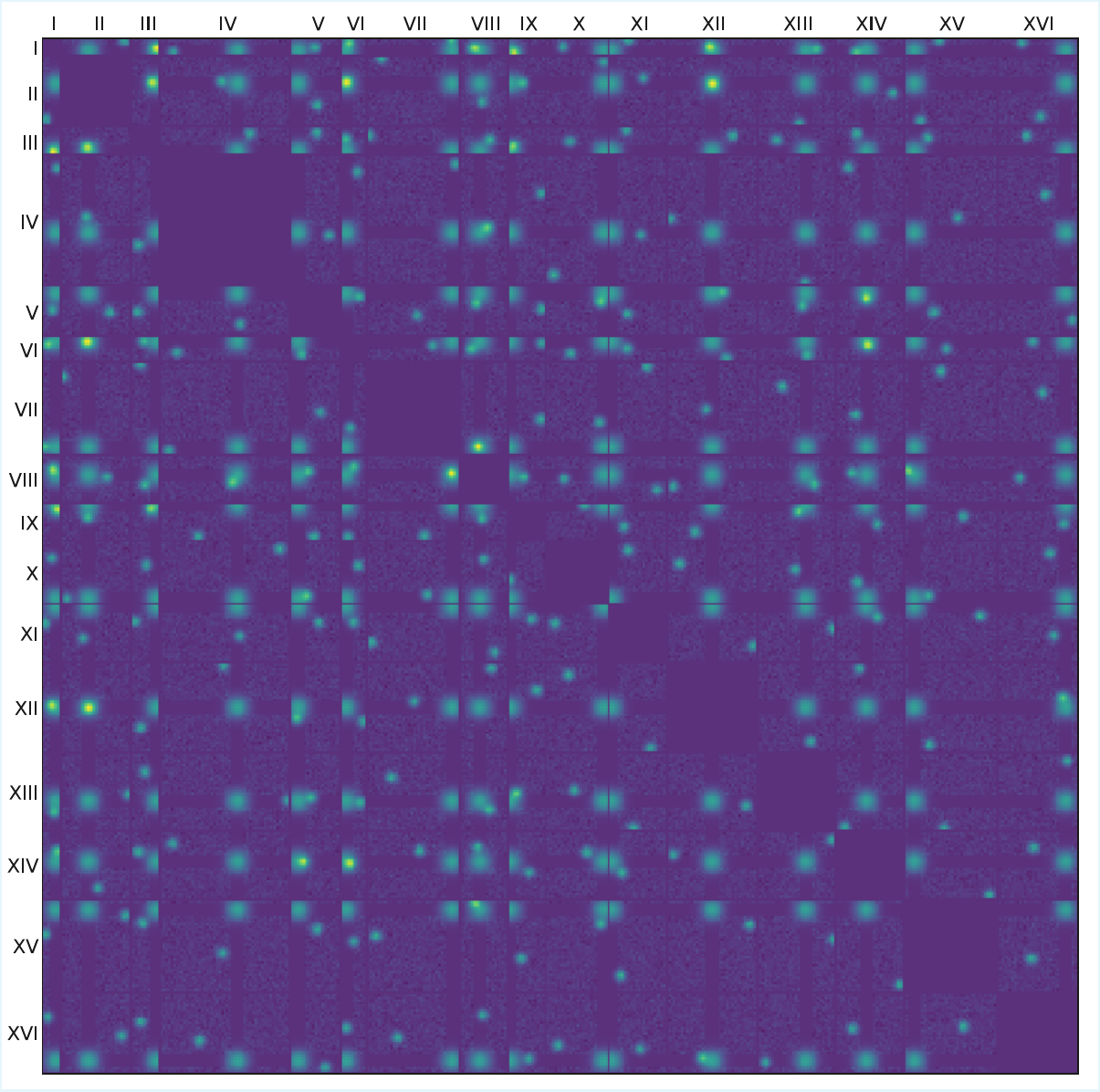}
    \caption{\small One synthetic map simulated from the S.C. genome with two Gaussian spots per trans-block.\label{fig:simu_multiple_spots}}
\end{figure}

When the contact maps are noisy: with $2$ Gaussian spots per trans-blocks, one major and one auxiliary smaller and less bright, our model outperforms \textit{Centurion} in both speed and accuracy, achieving 1.8 times better accuracy while running 10 times faster.
%\textit{Centurion} is $3$ times more accurate but the runtime is nearly $10^3$ slower.
Moreover, our method manages to estimate $\theta$ at a precision under the resolution. 
%To speed up \textit{Centurion}, we force only one candidate for the initialization, and in this setting, we manage to be more accurate than \textit{Centurion} in less time.

%\begin{figure}[H]
%    \centering

%    \begin{subfigure}[b]{0.48\linewidth}
%        \centering
 %       \includegraphics[width=\linewidth]{comparison_nelle_multiple_spots_cropped.pdf}
%        \caption{Multiple candidate pre-localizations}
%        \label{fig:comparison_nelle_simu_multiple_spots}
%    \end{subfigure}
%    \hfill
%    \begin{subfigure}[b]{0.48\linewidth}
 %       \centering
%        \includegraphics[width=\linewidth]{comparison_simu_nelle_multiple_spots_init_1_cropped.pdf}
%        \caption{Single candidate pre-localization}
%        \label{fig:comparison_nelle_simu_multiple_spots_1_init}
%    \end{subfigure}

 %   \caption{\small Absolute error (bp, left) and runtime (s, right) over $100$ synthetic maps generated from the S.C. genome at resolution $30$~kb. \textbf{(a)} $2$ candidates per centromere for pre-localization of \textit{Centurion}, \textbf{(b)} $1$ candidate per centromere for pre-localization of \textit{Centurion}.\\
 %   Per trans-block, one major Gaussian spot and one auxiliary Gaussian spot, smaller and less bright. The black dotted line in the left plots represents the resolution of the maps.}
 %   \label{fig:comparison_nelle_simu_multiple_spots_combined}
%\end{figure}

\subsubsection{Noisy map \label{sec:pirate_appendix}}
\begin{figure}[H]
    \centering
    \includegraphics[width=0.5\linewidth]{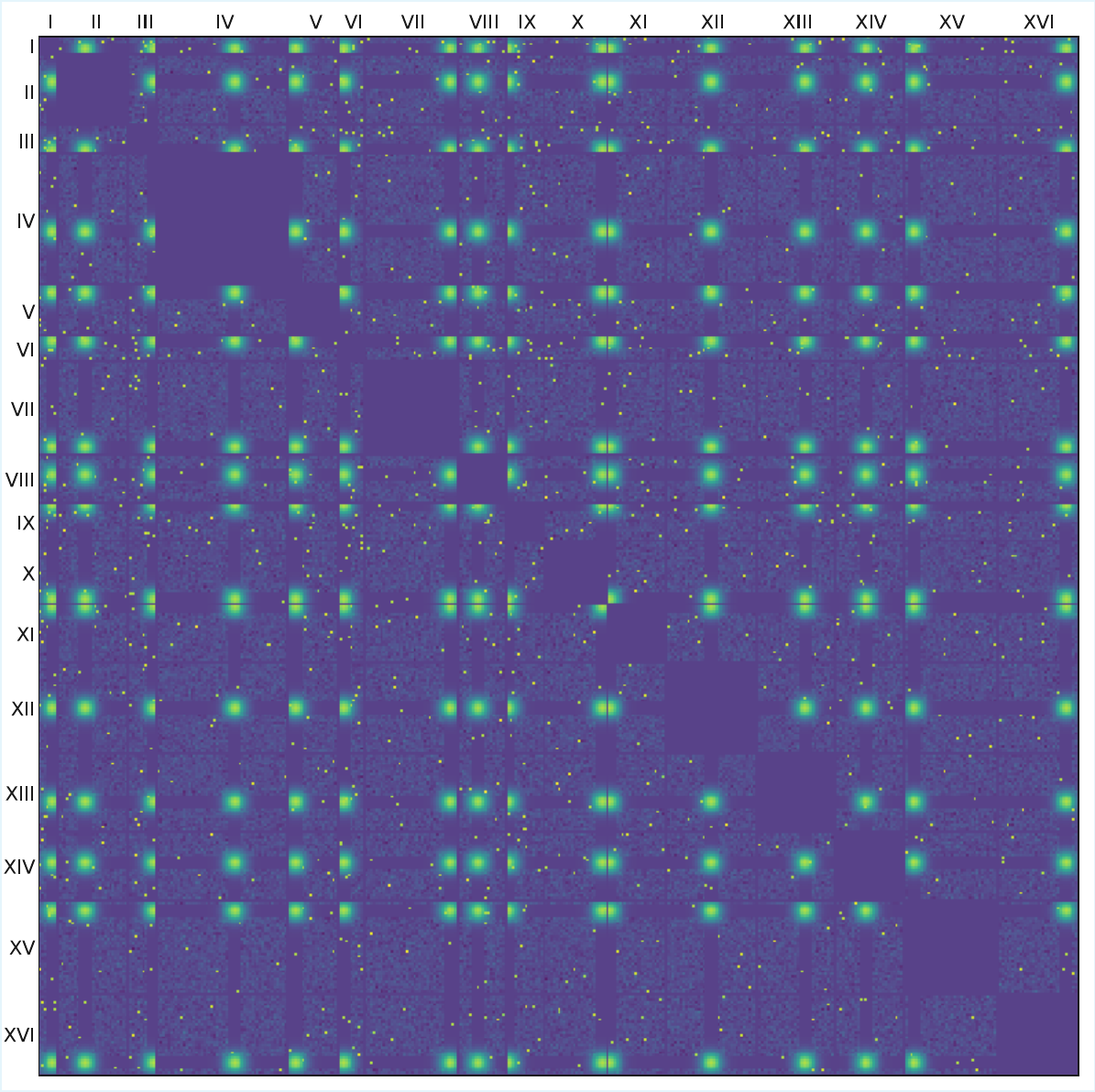}
    \caption{\small One synthetic map simulated from the S.C. reference genome with noise.\label{fig:simu_trap}}
\end{figure}
When the map is noisy: per trans-block, we add traps consisting in $5$ random pixels with intensity equal the maximum of the block, \textit{Centurion} produces estimates with a precision exceeding the map resolution, whereas our method is able to estimate $\theta$ with sub-resolution accuracy. In this setting \textit{Centurion} is $2.5$ times less accurate than our method and its runtime is nearly $10$ times slower. 
\begin{figure}[H]
    \centering
    \includegraphics[width=0.7\linewidth]{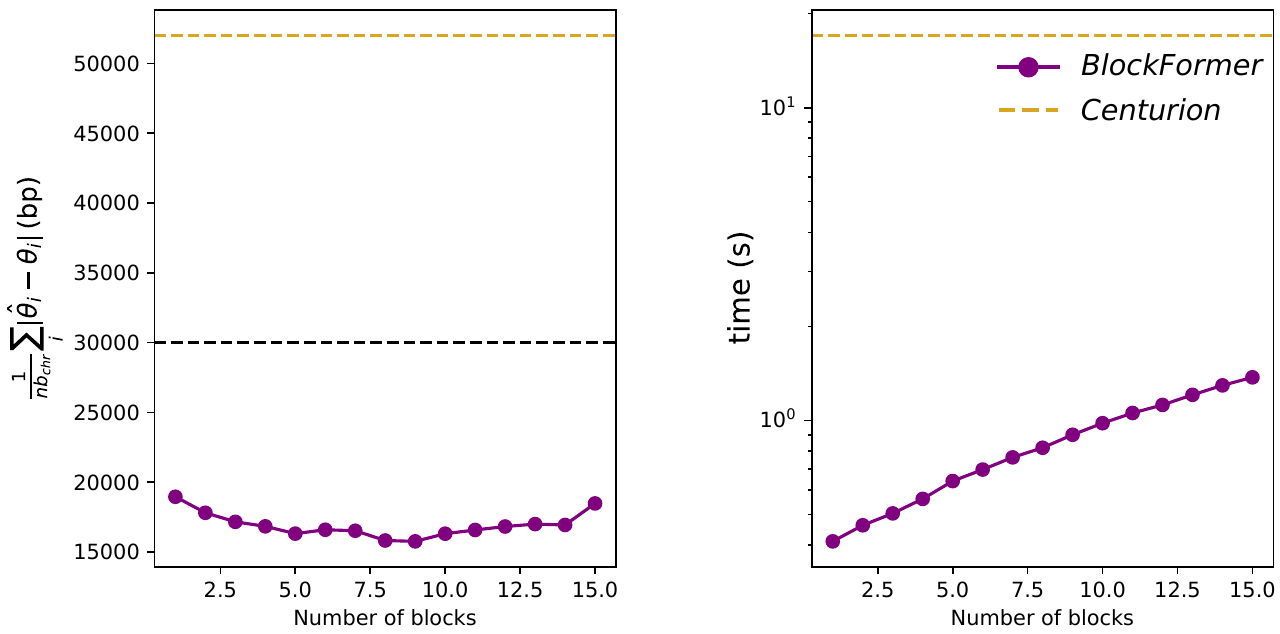}
    \caption{\small Absolute error (bp, left) and runtime (s, right) over $100$ synthetic maps generated from the S.C. genome at resolution $30$~kb. Per trans-block, we add $5$ random pixels with intensity the maximum of the bloc. The black dotted line in the left plot represents the resolution of the maps.\label{fig:comparison_nelle_simu_trap}
    % results time limit 1000 time optim 10 sigma 4 1 candidate centurion
    }
\end{figure}
\newpage
\subsection{Inference from real-world contact maps}
\subsubsection{Strategy to construct the parameter estimation $\hat{\theta}$ \label{sec:parameter_construction}}
In all experiments, $\hat{\theta}$ is estimated independently for each component.\\
In contact maps, some blocks may be large or noisy which can corrupt the centromere signal and make inference difficult. This can lead to inaccurate estimates depending on the selected blocks. We exploit the flexibility of our model to block sizes to bypass this issue. If we want to use $k$ trans-blocks, for each parameter component $i$, we randomly sample $10$ sequences of $k$ blocks and provide them to the network, producing $10$ candidate predictions. The final estimate $\hat{\theta_i}$ is computed as the mean over these candidates.   
\subsubsection{Centromere identification
\label{sec:other_species_appendix}}
We first provide genomic information of each studied specie.
\begin{table}[H]
\centering
\small
\begin{tabular}{lccc}
\toprule
\textbf{Species} & \textbf{Abbreviation} & \textbf{Number of chromosomes} & \textbf{Genome range (bp)} \\
\midrule
\multicolumn{4}{l}{Yeasts} \\
\textit{Kluyveromyces lactis} & K.L. & 6 & 1 062 590 – 2 602 197 \\
\textit{Lachancea kluyveri} & L.K. & 8 & 951 467 – 2 314 951 \\
\textit{Lachancea thermotolerans} & L.T. & 8 & 687 718 – 1 720 065\\
\textit{Saccharomyces cerevisiae} & S.C. & 16 & 230 209 – 1 533 918\\
\textit{Saccharomyces kudriavzevii} & S.K. & 16 & 170 892 – 1 436 357 \\
\textit{Saccharomyces mikatae} & S.M. & 16 & 188 471 – 1 444 590\\
\textit{Schizosaccharomyces pombe} & S.P. & 3 & 2 452 883 – 5 579 133\\
\textit{Zygosaccharomyces rouxii} & Z.R. & 7 & 881 646 - 1 865 392\\
\midrule
\multicolumn{4}{l}{Parasite} \\
\textit{Plasmodium Falciparum} & P.F. & 14 & 640 851 – 3 291 936\\
\midrule
\multicolumn{4}{l}{Plant} \\
\textit{Arabidopsis thaliana} & A.T. & 5 & 18 585 056 – 30 427 671\\
\bottomrule
\end{tabular}
\caption{Species used in this study with abbreviations and chromosome counts.}
\label{tab:various_species}
\end{table}

We show performances of \textit{BlockFormer} on  out-of-training chromosomes sizes. 

\begin{table}[H]
\centering
\small
\caption{\small Species with out-of-training chromosomes sizes, too noisy map or high resolution: we apply the refinement step as in Appendix ~\ref{sec:refine}. P.F.s. and P.F.t. stand for the $2$ stages schizonts and trophozoites of the parasite.  Comparison of normalized error (the smaller the better) and time. (5 random seeds)}
\label{tab:other_species_big_1}
\renewcommand{\arraystretch}{1.2}
\setlength{\tabcolsep}{3pt}
\resizebox{\linewidth}{!}{%
\begin{tabular}{cc cccccccccc}
\toprule
\multicolumn{2}{c}{\multirow{2}{*}{\textbf{Method}}} 
& \multicolumn{2}{c}{\textbf{S.K.}} 
& \multicolumn{2}{c}{\textbf{S.C.}} 
& \multicolumn{2}{c}{\textbf{K.L.}} 
& \multicolumn{2}{c}{\textbf{P.F.s.}}
& \multicolumn{2}{c}{\textbf{P.F.t.}} \\
\cmidrule(r){3-4}
\cmidrule(r){5-6}
\cmidrule(r){7-8}
\cmidrule(r){9-10}
\cmidrule(r){11-12}
& & Err & Time & Err & Time & Err & Time & Err & Time & Err & Time \\
\midrule

\multirow{2}{*}{\rotatebox[origin=c]{90}{Init.}} 
& \textit{BlockFormer}
& 0.83 $\pm$ 0.08 & \textbf{0.81 $\pm$ 0.22}
& \textbf{0.28 $\pm$ 0.02} & \textbf{0.72 $\pm$ 0.06}
& 0.83 $\pm$ 0.02 & \textbf{0.37 $\pm$ 0.01}
& \textbf{0.71 $\pm$ 0.03} & \textbf{1.11 $\pm$ 0.31}
&  \textbf{0.74 $\pm$ 0.01} & \textbf{0.97 $\pm$ 0.02} \\

& \textit{Centurion}
& \textbf{0.60 $\pm$ 0.0} & 2.50 $\pm$ 0.42
& 8.12 $\pm$ 0.0 & 1.24 $\pm$ 0.09
& \textbf{0.73 $\pm$ 0.0 } & 0.76 $\pm$ 0.01
& 1.06 $\pm$ 0.0 & 15.36 $\pm$ 1.74
& 1.07 $\pm$ 0.0 & 13.47 $\pm$ 0.07 \\

\midrule

\multirow{2}{*}{\rotatebox[origin=c]{90}{Fitting}} 
& \textit{BlockFormer}
& 0.53 $\pm$ 0.17 & \textbf{589.98 $\pm$ 92.07}
& \textbf{0.20 $\pm$ 0.0} & \textbf{33.0 $\pm$ 3.34}
& \textbf{0.18 $\pm$ 0.0} & \textbf{2.34 $\pm$ 0.19}
& \textbf{0.26 $\pm$ 0.0} & \textbf{41.34 $\pm$ 8.41}
& \textbf{0.28 $\pm$ 0.0} & \textbf{28.79 $\pm$ 0.63} \\

& \textit{Centurion}
& \textbf{0.44 $\pm$ 0.0} & 1507.59 $\pm$ 1784.14
& \textbf{0.20 $\pm$ 0.0} & 137.29 $\pm$ 5.86
& \textbf{0.18 $\pm$ 0.0} & 2.63 $\pm$ 0.13
& \textbf{0.26 $\pm$ 0.0} & 280.96 $\pm$ 18.70
& \textbf{0.28 $\pm$ 0.0} & 68.33 $\pm$ 4.0 \\

\bottomrule
\end{tabular}%
}
\end{table}

\begin{figure}[H]
    \centering
    \includegraphics[width=14cm, height=8cm]{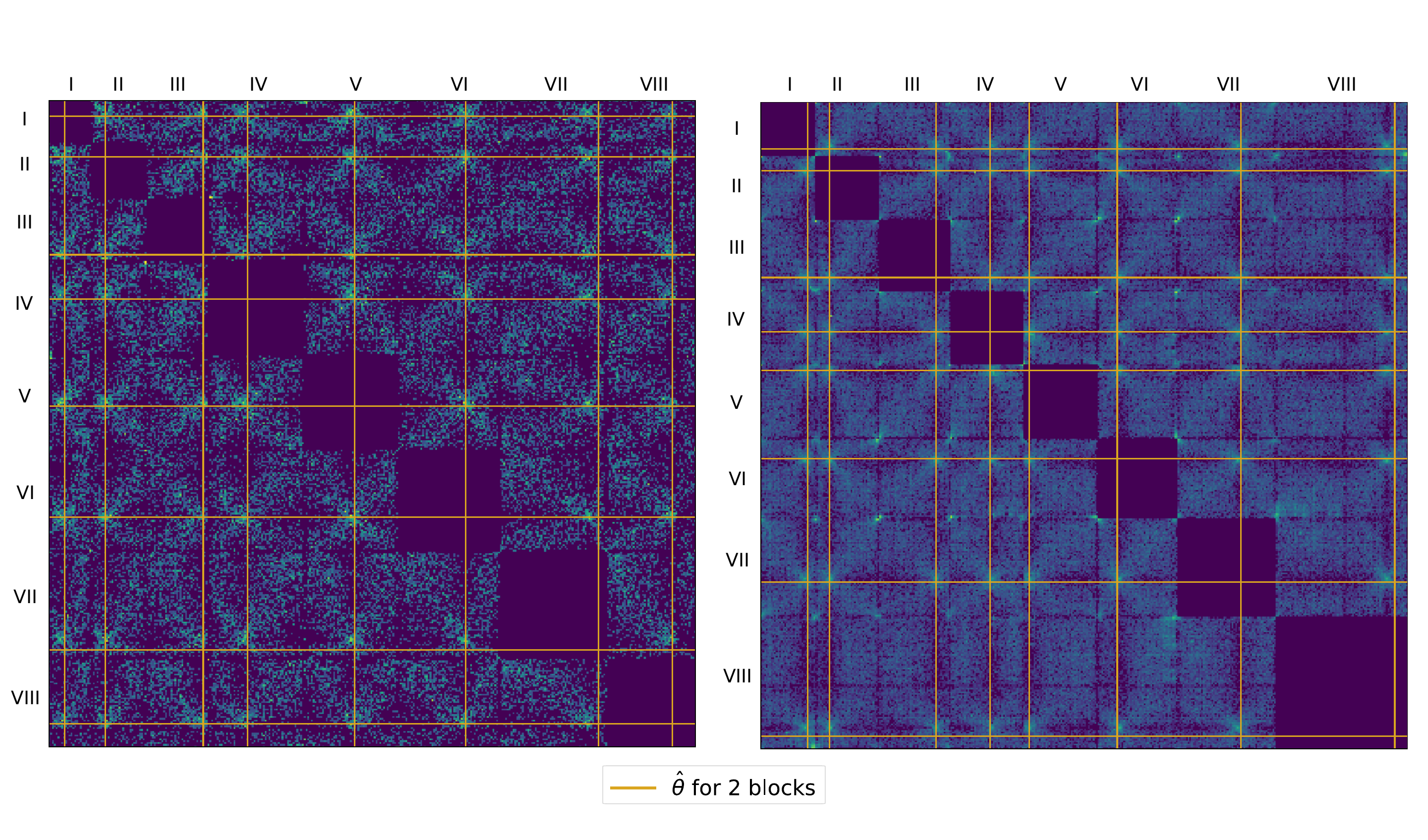}
    \caption{\small Centromere estimation for the yeasts L.T. (left) and L.K. (right), resolution $30$~kb.}
    \label{fig:lk_lt}
\end{figure}

We show the parameter estimation using our network with $2$ trans-blocks for the yeasts L.T. and L.K. at resolution $30$~kb (see Figure~\ref{fig:lk_lt}).
In almost all dimensions except for chromosome $7$ in L.T. and for chromosome $8$ in L.K., the network achieves to estimate $\theta_i$ with a precision close to the resolution: graphically, the positions are aligned with the spots of interactions on the reference maps.

\newpage
\subsubsection{Loop localization \label{sec:loop_appendix}}
We evaluate the capacity of \textit{BlockFormer} to estimate some loops given Hi-C data from the human IMR90 cell line. Since full contact maps are significantly larger than those used during training, we restrict our analysis to genomic regions around the main diagonal considered as local cis-blocks.\\
Several structural characteristics require adaptations to fit to \textit{BlockFormer}. \begin{itemize}
    \item loops are intra-chromosomal features, which necessitates to work on individual cis-blocks instead of multiple trans-blocks used for the centromeres identification. 
    \item cis-blocks are symmetric causing loops to appear twice in the map. We thus provide either the upper or lower triangular part of the map to the model or localized patches around each loop.
    \item the strong signal along the main diagonal in cis-block can bias the estimation, motivating the use of an observed-over-expected normalization for the map (see Algorithm~\ref{alg:oe}). 
    \item the number and locations of loops within a given cis-block are unknown whereas \textit{BlockFormer} was trained on sequences of trans-blocks containing a single spot per block. We thus restrict our analysis to genomic regions where only one loop is visible.
    \item each loop position is defined by two coordinates whereas \textit{BlockFormer} produces only a single scalar parameter per entity. We therefore perform two passes through the model to separately infer the x- and y-coordinates of each loop.
\end{itemize}
Since no experimentally validated ground-truth loop annotations are available, we use qualitative visual inspection and an external loop-calling method for quantitative evaluation. Specifically, we use the algorithm \textit{Chromosight} \cite{loop_ground_truth} a pattern-based algorithm for detecting structural features in Hi-C maps  such as loops or TADs as a reference for loop localization.
\begin{algorithm}[H]
\caption{Observed-over-expected normalization \label{alg:oe}}
\begin{algorithmic}
\State \textbf{Input}: contact map $C$ of size $(N \times N)$
\State \textbf{Return}: observed-over-expected map $OE$ of size $(N \times N)$ 
\State
\State compute the matrix of distance to the diagonal $D$ $(N \times N)$ such that $D_{i,j} = |i-j|$
\State
\State compute the expected maps $E$ $(N \times N)$ such that $E_{i,j} = \underset{|k-l| =D_{i,j}}{\mathrm{mean}} (C_{k,l})$
\State construct the observed-over-expected map $OE$ of size $(N \times N)$ such that $OE_{i,j} = \frac{C_{i,j}}{E_{i,j}}$
\State \textbf{return} the observed-over-expected map $OE$ of size $(N \times N)$.
\end{algorithmic}
\end{algorithm}

We first report quantitative performance results by treating \textit{Chromosight} detections as reference loop positions in an automatized procedure. We compare \textit{BlockFormer}’s performance with \textit{Centurion}. As \textit{Centurion} is fine-tuned for centromere identification using genome-wide contact maps, it faces the same limitations as our approach (single block, unknown number of spots per block, pair coordinates estimation per spot). 
%To address this, we provide it with the upper then the lower triangular part of the observed-over-expected map to have the loop estimation.
Therefore, \textit{BlockFormer} and \textit{Centurion} require to operate on regions of the Hi-C map that are clean and contain only a single loop. A pre-localization step is necessary to identify such regions. To address this, we employ a simple peak detection algorithm on the Hi-C map presented in Algorithm~\ref{alg:loop_preloc}. 

\newpage

\begin{algorithm}[H]
\caption{Loop pre-localization \label{alg:loop_preloc}}
\begin{algorithmic}
\State \textbf{Input}: observed-over-expected map $OE$ of size $(N \times N)$
\State \textbf{Return}: pre-located loops 
\State
\State smooth the map via Gaussian blur with $\sigma=1$ to denoise it.
\State construct a threshold to define what is a peak in the map via a percentile (e.g. $92\%$).  
\State find local maxima in the map: compare each pixels to their neighborhood region of a given size (e.g. $5$).
\State filter candidates: a peak is a local max with intensity value above the threshold.
\State filter peaks that are too close from the diagonal with a given distance (e.g. $3$) because they correspond to self-interactions.
\State \textbf{return} a set of loops pre-localization.
\end{algorithmic}
\end{algorithm}
The full process for loop localization using \textit{BlockFormer} or \textit{Centurion} is described in Algorithm~\ref{alg:loop_detection}. 
\begin{algorithm}[H]
\caption{Loop detection with \textit{BlockFormer} \label{alg:loop_detection}}
\begin{algorithmic}
\State \textbf{Input}: region from Hi-C map $C$ of size $(N \times N)$
\State \textbf{Return}: loops estimation 
\State compute the observed-over-expected map $OE$ from $C$ via Algorithm~\ref{alg:oe}.
\State find loops pre-localization via Algorithm~\ref{alg:loop_preloc} (referred as pre-loc loops).
\State find loops reference positions via \textit{Chromosight} (referred as true loops).
\State filter pre-loc loops to ensure that each of them are close to one true loop. 
\For{ each pre-loc loop}

\State define a region of size $30 \times 30$ in the map $OE$ around the pre-loc loop (referred as $C_\text{loop}$).
\State $0-$pad $C_\text{loop}$ to a multiple of the patch size and norm it between $0-1$
\State pass $C_\text{loop}$ to \textit{BlockFormer} to find the $i$-coordinate of the loop: $\theta_i$.
\State pass $C_\text{loop}^T$ to \textit{BlockFormer} to find the $j$-coordinate of the loop: $\theta_j$. 
\EndFor
\State \textbf{return} a set of loops localization estimations $(\theta_i, \theta_j)$.
\end{algorithmic}
\end{algorithm}

\newpage

We report performance of both methods across multiple chromosomes. The metric used is the absolute error normalized by the resolution. In each chromosome, more than $200$ loops are identified.
\begin{figure}[H]
    \centering
    \includegraphics[width=0.8\linewidth]{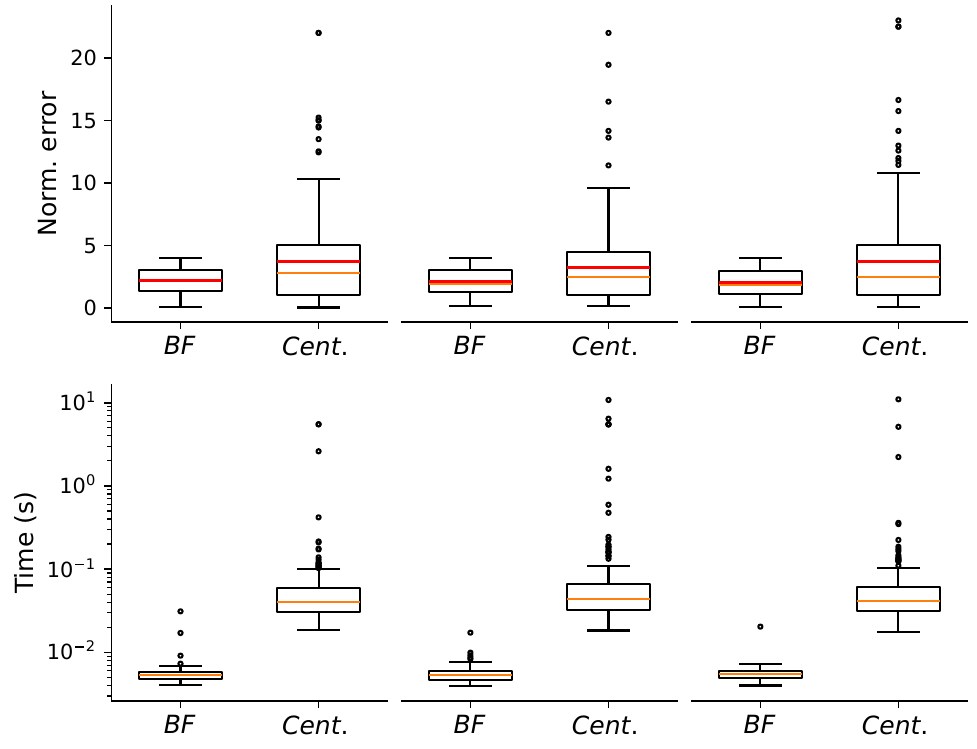}
    \caption{\small Loops detection at resolution $5$ kb in automated selected genomic regions of chromosomes $1$ (left), $3$ (middle) and $7$ (right). \textit{BlockFormer} (referred as \textit{BF}) or \textit{Centurion} (referred as \textit{Cent.}) take as input pre-localized regions in the observed-over-expected maps. Red lines in the top plots indicate the mean error. Depending on the region, \textit{BlockFormer} does not always estimate loops at under-resolution precision but remains however more precise and faster than \textit{Centurion}.} 
    \label{fig:auto_loops}
\end{figure}

\begin{table}[H]
    \centering
    \small
    \caption{Performance comparison (mean $\pm$ std) across chromosomes.}
    \label{tab:loop_comparison}
    \begin{tabular}{lccc}
    \toprule
    Method & chr. 1 & chr. 3 & chr. 7 \\
    \midrule
    \textit{BlockFormer}& \textbf{2.21 $\pm$ 1.03} & \textbf{2.12 $\pm$ 1.08} & \textbf{2.01 $\pm$ 1.09} \\
    \textit{Centurion}& 3.73 $\pm$ 3.68 & 3.26 $\pm$ 3.27 & 3.70 $\pm$ 3.93 \\
    \bottomrule
    \end{tabular}
    \end{table}
        
\textit{BlockFormer} can be adapted to loop localization, requiring however a pre-localization step. \textit{Centurion} is not well suited for single block inference because its optimization process is based on the alignment of multiple spots across blocks in the map. On the contrary, thanks to its training strategy and its flexible architecture, \textit{BlockFormer} can adapt to this setup and outperforms \text{Centurion} both in accuracy and time.

\newpage

We further present qualitative results where genomic regions have been manually selected across multiple chromosomes.\\

\begin{figure}[H]
    \centering
     \includegraphics[width=0.5\textwidth]{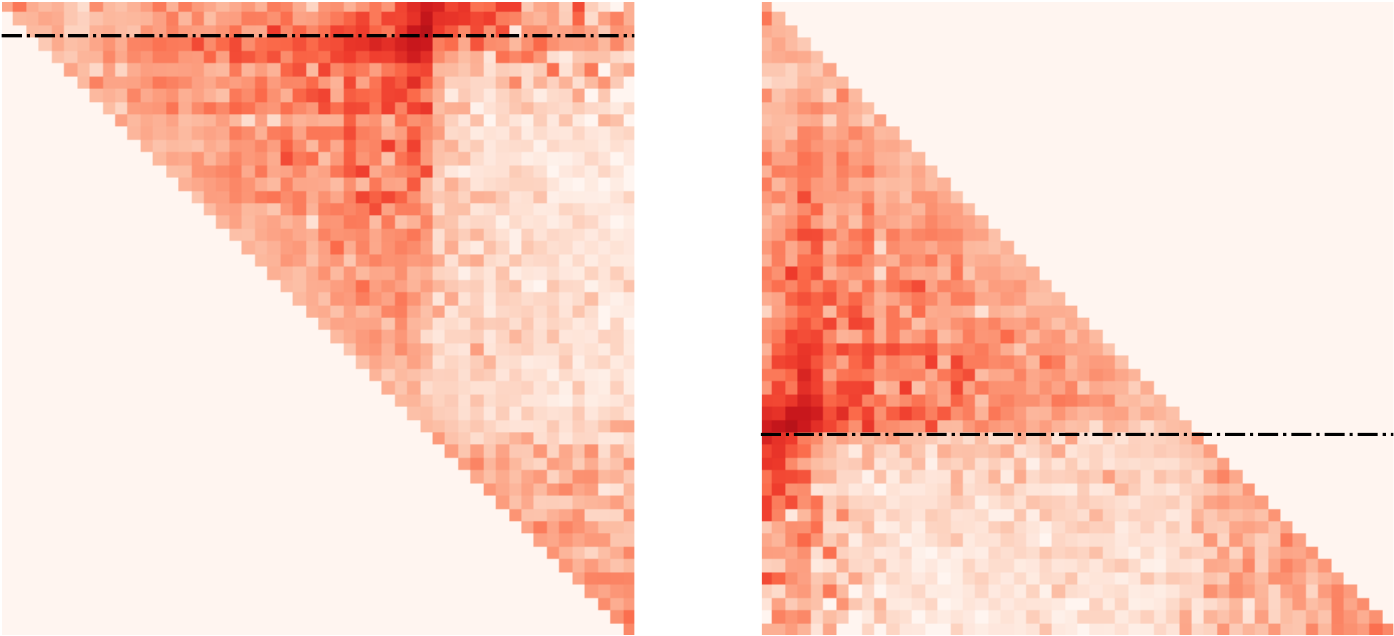}

    \vspace{0.8cm}

    \includegraphics[width=0.5\textwidth]{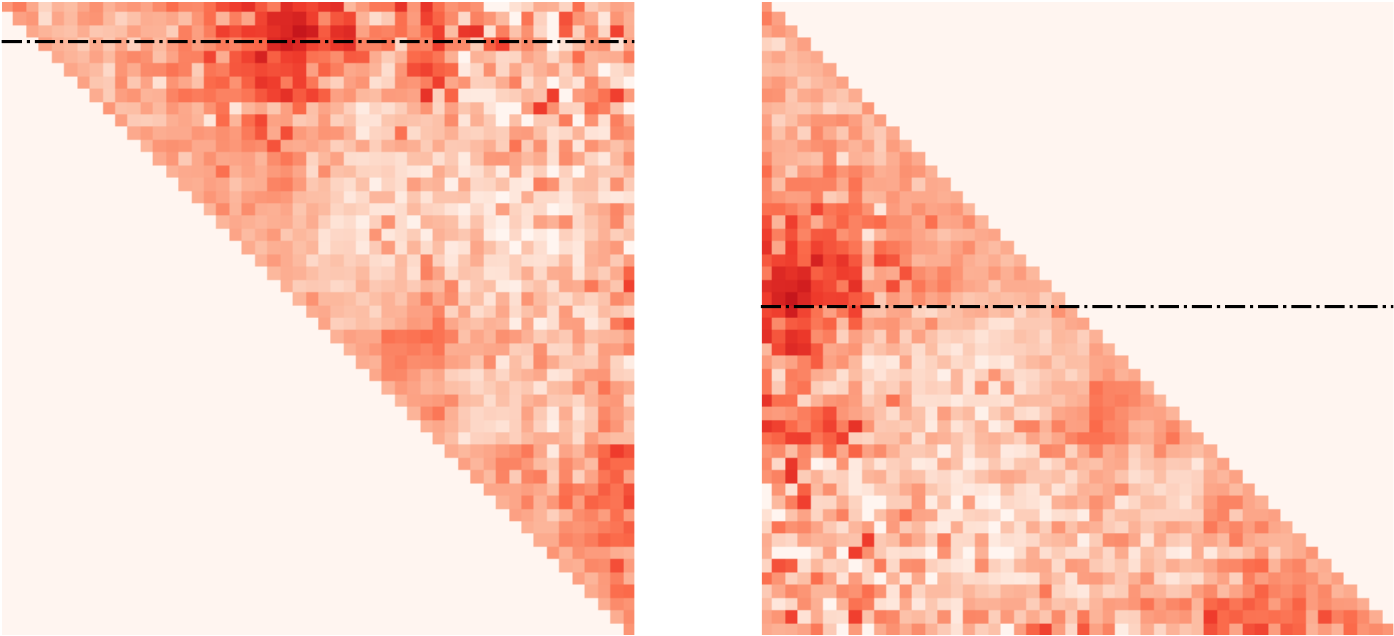}
        
    \caption{\small Example of observed-over-expected maps used as input to \textit{BlockFormer} for loop position estimation. Only the upper (left) or lower (right) triangular part is considered for respectively the y- or x- coordinate estimations. Dashed lines represent the model's predictions of the loops.}
    \label{fig:oe_loops}
\end{figure}

\begin{figure}[H]
    \centering
     \begin{minipage}{0.32\textwidth}
        \centering
        \includegraphics[width=\linewidth]{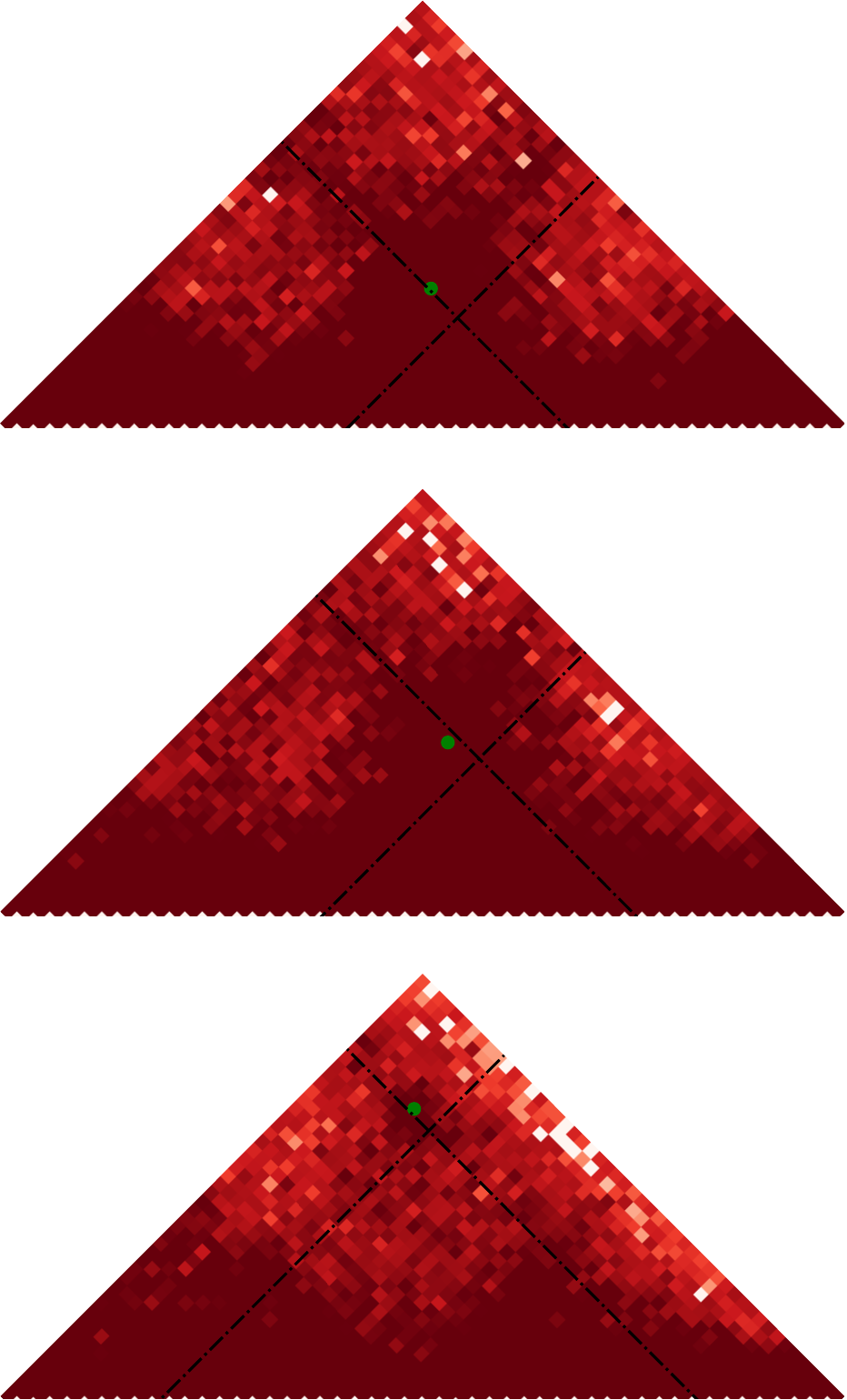}
        
    \end{minipage}
    \hfill
    \begin{minipage}{0.32\textwidth}
        \centering
        \includegraphics[width=\linewidth]{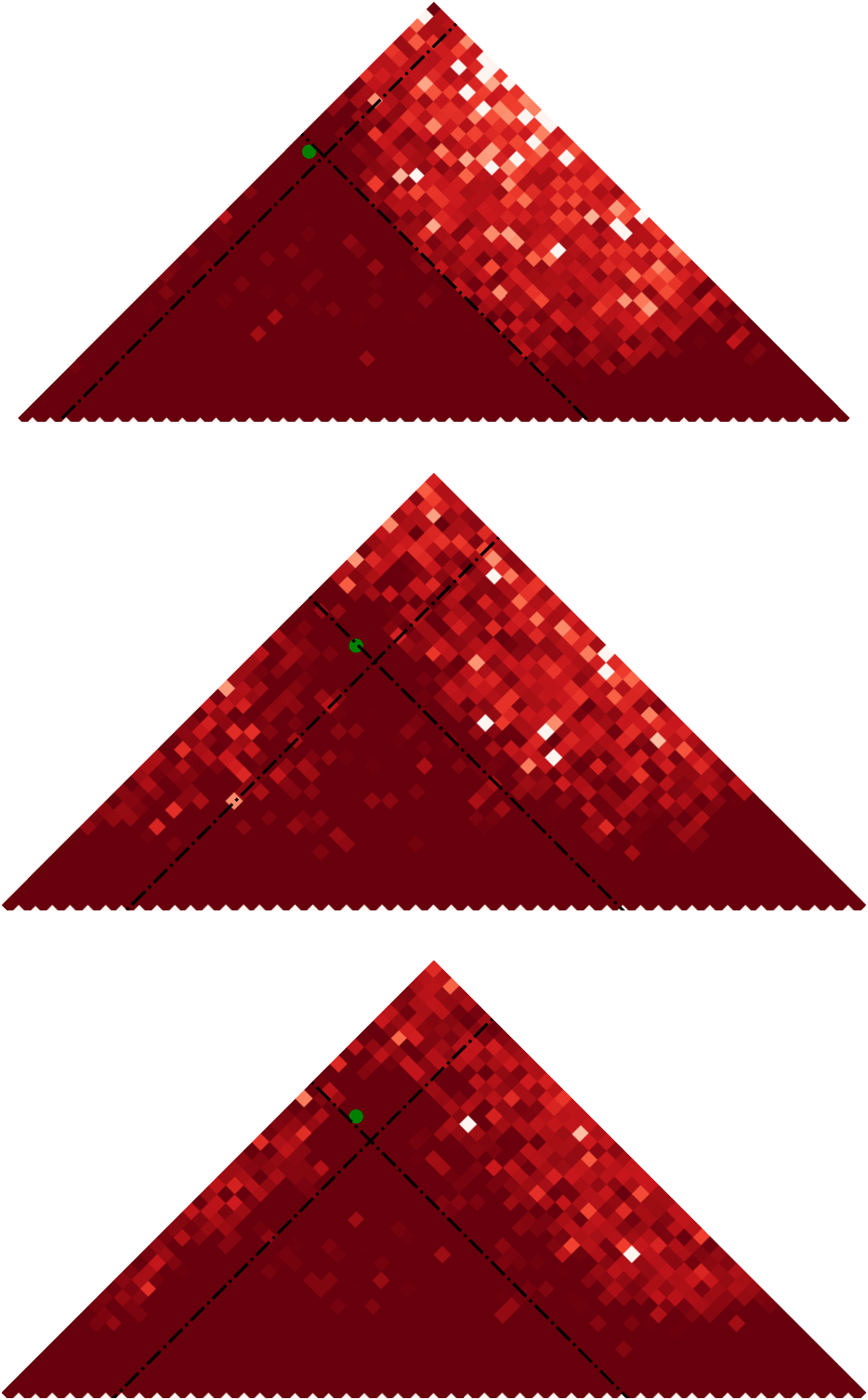}
        
    \end{minipage}
    \hfill
    \begin{minipage}{0.32\textwidth}
        \centering
        \includegraphics[width=\linewidth]{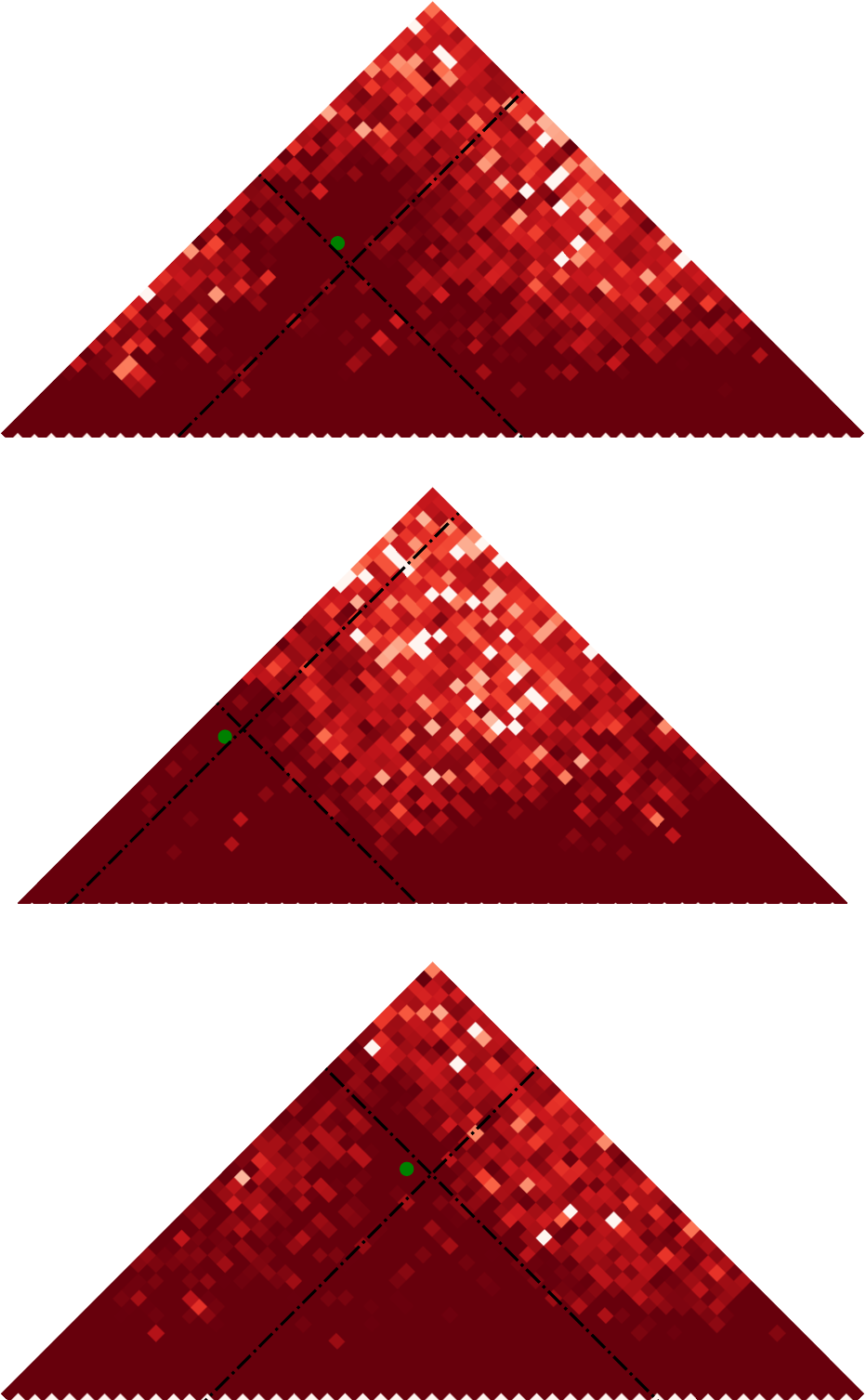}
        
    \end{minipage}
    \caption{\small Loops detection at resolution $5$ kb in selected genomic regions of chromosomes $1$ (left), $3$ (middle) and $7$ (right). Loops appear as bright off-diagonal enrichment spots. \textit{BlockFormer} takes as input the upper or lower triangular part of the observed-over-expected maps. \textit{Chromosight} takes as input the entire raw map but considers only the upper triangular part. Dashed lines represent model's loop predictions, while green dots indicate \textit{Chromosight}'s loop positions. Agreement between both methods indicates that when the map is sufficiently clear, the model accurately identifies loops.}
    \label{fig:estim_loops}
\end{figure}
%%%%%%%%%%%%%%%%%%%%%%%%%%%%%%%%%%%%%%%%%%%%%%%%%%%%%%%%%%%%
%\end{document}
\newpage
 \section{Simulation-based inference for parameter uncertainty quantification}
\subsection{Generic framework \label{sec:sbi_presentation}}
 Interaction maps are summary statistics over a population of entities and thus inherently contain bias and noise. This is particularly visible in Hi-C data where DNA folding is highly variable. A Bayesian approach offers a clear advantage over simple point estimation by explicitly modeling this uncertainty (posterior width reflects confidence), allowing variability and noise to be incorporated into the analysis. This stochastic framework provides a more faithful representation of the underlying biological heterogeneity and yields more robust, interpretable inferences.\\ 
 The goal is to infer a set of per-entity parameters $\theta$ from an interaction map $C_\text{ref}$ using a probabilistic framework based on simulations. The usual way for doing so would be to search for the most appropriate $\theta$ for a given $C_\text{ref}$ by maximizing the likelihood: 
\begin{equation*}
    \hat{\theta} = \underset{\theta \in \Omega}{\text{argmax}}\ \log p(C_\text{ref}|\theta)~.
\end{equation*} 
However, as the simulator is often very complex (e.g. biological simulators), the {likelihood} $p(C|\theta)$ may be {intractable}. As such, we directly {target the posterior density} $p(\theta|C_\text{ref})$ using data from the simulator, either via {approximate Bayesian computation} (Sequential Monte-Carlo ABC: SMC-ABC) or by estimating the posterior density with a conditional normalizing flow (Sequential neural posterior estimation: SNPE) \cite{snpea, snpec}.

\subsection{Inference of variable size parameter \label{sec:indep}}
As the dimensionality of the parameter to infer varies with the number of entities, directly inferring any $\theta$ from any interaction map $C$ becomes challenging. Moreover, if we consider the entire interaction maps and the corresponding parameters in the Bayesian inference methods, we face the curse of dimensionality: the space of parameters $\theta$ becomes too large to cover with few simulations, and the maps $C$ are too big. To bypass those issues, we decompose the problem into multiple sub-problems: we perform in parallel one inference per entity, estimating $\theta_i$ from $C_i$, the parameter-related part of $C$. The space of $\theta$ is thus cut into several entity-length 1D intervals, reducing the train set size. Assuming $I$ entities, the actual targeted posterior is 
\begin{equation*}
    p^\otimes(\theta | C_\text{ref}):= \prod_{i=1}^I p(\theta_i | C_{\text{ref}, i})
\end{equation*}
 $p^\otimes(\theta | C_\text{ref})$ is not the true joint posterior $p(\theta | C_\text{ref})$.
To justify it, we represent $C$ as a sequence of trans-blocks, $C = (C_{k,l})_{1 \leq k \neq l \leq I}$ and draw a part of the causal graph between $\theta$ and $C$: 
\begin{center}
\begin{tikzpicture}[->, node distance=1cm]

% Nodes (top)
\node (cik) at (0,0.5) {$C_{i,k}$};

% Middle layer (thetas)
\node (t1) at (-2,0) {$\theta_1$};
\node (ti) at (-1,0) {$\theta_i$};
\node (tj) at (0,0) {$\theta_j$};
\node (tk) at (1,0) {$\theta_k$};
\node (tn) at (2,0) {$\theta_n$};
\node at(-1.5,0) {$\cdots$};
\node at(1.5,0) {$\cdots$};

% Bottom layer (blocks)
\node (cij) at (-0.5,-1) {$C_{i,j}$};
\node (cjk) at (0.5,-1) {$C_{j,k}$};

% Edges (bottom colliders)
\draw[red] (ti) -- (cij);
\draw[red] (tj) -- (cij);

\draw (tj) -- (cjk);
\draw[green] (tk) -- (cjk);

% Edges (top collider)
\draw[green] (ti) -- (cik);
\draw[green] (tk) -- (cik);

\end{tikzpicture}
\end{center}
We first introduce a \textit{mean-field approximation} by assuming the conditional independence across chromosomes: $p(\theta|C) \approx \prod_{i \in I} p(\theta_i | C)$.\\
Indeed, since $C$ contains the block $C_{i,j}$ that is jointly generated from $(\theta_i, \theta_j)$, $C_{i,j}$ acts as a collider in the causal graph (in red) and $\theta_i \not\!\perp\!\!\!\perp \theta_j | C$.\\
We furthermore introduce a \textit{locality assumption} by stating that $p(\theta_i | C) \approx p(\theta_i | C_i)$.\\
$\theta_i$ appears directly in $C$ only through blocks $C_{i,k}$ or $C_{k,i}$ and is indirectly influenced by other blocks $C_{j,k}$. Indeed, in the causal graph, the green path is open since it contains a collider and $\theta_i \not\!\perp\!\!\!\perp C_{j,k} | C_{i,k}$. However, we only consider first-order dependencies (direct causal link).\\
So $p(\theta_i |C) = p(\theta_i | (C_{k,l}, {1 \leq k \neq l \leq I}) \approx p(\theta_i | [C_{i,k}, C_{k,i}], k \neq i)$.\\
Since $C_{k,i} = C_{i,k}^T$, it does not contain additional information and $p(\theta_i | [C_{i,k}, C_{k,i}], k\neq i) = p(\theta_i | C_{i,k}, k\neq i):= p(\theta_i | C_i)$ such that $p(\theta_i | C) \approx p(\theta_i | C_i)$. To summarize:\begin{equation*}
    p(\theta |C_\text{ref}) \underset{\text{\textit{mean-field approx.}}}{\approx} \prod_{1 \leq i \leq I} p(\theta_i | C_\text{ref}) \underset{\text{\textit{locality}}}{\approx} \prod_{1 \leq i \leq I} p(\theta_i | C_\text{ref, i}) = p^\otimes(\theta | C_\text{ref}) 
\end{equation*}
Although this approximation ignores higher-order dependencies induced by collider paths in the full causal graph, it yields a scalable inference procedure that can be independently applied across entities of varying dimensionality.\\
In all the presented inference methods in Appendices~\ref{sec:ABC_appendix} and~\ref{sec:snpe_appendix}, we target in parallel, dimension per dimension, each marginal $p(\theta_i |C_i)$. \textbf{To avoid notational overhead in the algorithms, $\theta$ will represent any $\theta_i$ and $C$, any $C_i$}. 

\newpage
\section{SMC-ABC \label{sec:ABC_appendix}}
We use a variant of ABC coupled with sequential Monte-Carlo (SMC)~\cite{smcabc}. It consists of multiple rounds of ABC where, at each round, relevant $\{\theta^{k,*}\}_k$ are selected from the training set $\{(\theta^n, C^n)\}_n$ depending on a closeness criterion between $C$ and $C_\text{ref}$. We then associate weights $\{w^k\}_k$ to those selected $\{\theta^{k,*}\}_k$, and use the set $\{(\theta^{k,*}, w^k)\}_k$ to create the next population of $\{\theta^n\}_n$ for the next round of ABC. This sequential approach enables us to refine the relevant $\theta$ at each round.
However, we need to {define a metric} for discriminating $(\theta^n, \theta^m)$ based on their associated observations $(C^n, C^m)$.
\subsection{With the metric Pearson correlation -- \ABCPearson \label{sec:SMCABCPearson}}
To measure the closeness between $C$ and $C_\text{ref}$, the Pearson correlation is commonly used~\cite{tjong, nelly, nelle}. We find that the {vector-based Pearson correlation} averaged over all trans-contacts blocks is the most discriminative metric: each trans-contacts block of $C$ and $C_\text{ref}$ is vectorized and the Pearson correlation is computed between both. We then average all the correlations over the trans-contacts blocks (see Algorithm ~\ref{alg:smcabc}). However, this metric is fine-tuned to this specific inference task.

\begin{algorithm}[H]
\caption{SMC-ABC based on Pearson correlation inspired from \cite{smcabc} }\label{alg:smcabc}
\begin{algorithmic}
\State \textbf{Input}: $T$ rounds, prior $p$, train set of size $N$, acceptance rate $5\%$, perturbation kernel $K = \mathcal{N}(., \sigma^2\text{Id})$ ($\sigma = $ resolution (bp))
\State \textbf{Return}: $\theta \sim p(\theta|\text{corr}(C,C_\text{ref}) \geq \epsilon_{\text{corr}})$
\State \textbf{round $t=0$}
\State - sample $\theta^n \sim p$, and $C^n \sim p(.|\theta^n) , n \in \llbracket 1, N \rrbracket$
\State - compute $\text{corr}(C^n, C_\text{ref})$ and keep the top $5\%$ of $\{\theta^n\}_n$ in terms of the highest correlation: $\{\theta^{m,0}, m \in \llbracket 1, M \rrbracket\}$ with $M = 5\% \times N$
\State - compute weights $\{w^{m,0} = \frac{1}{M} , m \in \llbracket1, M\rrbracket\}$
\State \textbf{output round $t=0$}: $\{ (\theta^{m,0}, w^{m,0}) \}_{m \in \llbracket 1, M\rrbracket}$
\For{$0 < t < T$}
\State \textbf{round $t$}
\State - from the previous accepted $\{\theta^{m,t-1}\}_{m \in \llbracket 1, M \rrbracket }$, sample $\{ \bar{\theta}^k, k \in \llbracket 1, M \rrbracket\}$ from multinomial $\mathcal{M}(\{\theta^{m,t-1}\}_m, \{w^{m,t-1}\}_m)$ with replacement
\State - perturb $\frac{N}{M}$ times the $M$ samples $\bar{\theta}^k$ to have $N$ samples $\theta^n$ 
\begin{equation*}
    \theta^n \gets \bar{\theta}^k + \epsilon \text{ with } \epsilon \sim \mathcal{N}(0, \sigma^2\text{Id}) \text{ for } k=n \text{ mod } M \text{ and } n=1,...,N
\end{equation*}
\State - check that $\theta^n$ is in the prior bound otherwise, set $\theta^n \gets \bar{\theta}^k$
\State - from this set $\{\theta^n\}_{n \in \llbracket 1,N\rrbracket}$, sample $C^n \sim p(.|\theta^n) , n \in \llbracket1, N\rrbracket$
\State - compute $\text{corr}(C^n, C_\text{ref})$ and keep the top $5\%$ of $\{\theta^n\}_n$ in terms of the highest correlation: \{$\theta^{m,t}, m \in \llbracket1, M\rrbracket\}$
\State - compute corresponding weights 
\begin{equation*}
     w^{m,t} = \frac{p(\theta^{m,t})}{\sum_{k=1}^M w^{k, t-1} K(\theta^{m,t}; \theta^{k, t-1})}
\end{equation*}
\State \textbf{output round $t$}: $\{ (\theta^{m,t}, w^{m, t}) \}_{m \in \llbracket 1, M \rrbracket}$
\EndFor
\State \textbf{return} accepted samples $\theta^n \sim p(\theta|\text{corr}(C^n,C_\text{ref}) \geq \epsilon_{\text{corr}})$
\end{algorithmic}
\end{algorithm}

When $\epsilon _{\text{corr}}\to 1$, $p(\theta|\text{corr}(C,C_\text{ref}) \geq \epsilon_{\text{corr}}) \to p(\theta | C_\text{ref})$ \cite{smcabc}.

\newpage
\subsection{With a summary statistic and the classical $l^2$-norm -- \ABCTransf or \ABCCNN \label{sec:ABCCNN_appendix}}
Instead of looking for a specific metric to compare $C$ to $C_\text{ref}$, we choose to use the classical $l^2$-norm. For this, we need a summary statistic $S$ that will extract the main features of $C$ and project it into a low-dimensional vector. We employ \textit{BlockFormer} as a \emph{data-driven summary statistic} (method referred as \ABCTransf) motivated by the fact that $\mathcal{BF}_\phi(C)$ pre-trained with Equation \ref{eq:loss} approximates the conditional expectation $\mathbb{E}\left[\theta|C\right]$,  which is, by definition, the solution to the regression of $\theta$ from $C$:
\begin{equation*}
    \mathbb{E}\left[\theta|C\right] = \underset{S \in \mathcal{F}}{\text{argmin}}\ \mathbb{E}\left[ \Vert S(C)-\theta \Vert ^2_2\right]~,
\end{equation*}
where $\mathcal{F}$ is the set of square integrable functions. As shown in \cite{summstatabc}, such a choice is relevant because $\mathbb{E}\left[\theta|C\right]$ preserves first-order information when summarizing $C$ as the threshold $\epsilon$ approaches $0$.\\
We also compare the approach with another architecture for $S$ composed of a CNN shared across chromosomes followed by a per-chromosome set of MLPs (method referred as \ABCCNN). 

\begin{algorithm}[H]
\caption{ABC with learned summary statistic inspired from \cite{summstatabc} \label{alg:abc-cnn}}
\begin{algorithmic}
\State \textbf{Input}:  (deep) neural network (DNN) $S_{\phi}$, threshold $\epsilon$, Euclidean norm in $\mathbb{R}^n$, simulator, prior $p$
\State \textbf{Return}:  Samples $\theta$ from the estimated posterior density $p(. \mid \Vert S_{\phi}(C)-S_{\phi}(C_\text{ref}) \Vert_2  \leq \epsilon)$\\ 
\State \textbf{Stage 1: learn the summary statistic} $S_{\phi}(.)$ s.t. $S_{\phi}(C) \approx \mathbb{E}\left[\theta|C\right]$
\State generate a train set $(\theta^n, C^n)$ from $p(\theta)p(C|\theta)$
\State train a DNN $S_\phi$ on this train set with the loss to minimize in $\phi$ \begin{equation*}
    \widehat{\mathcal{L}}_{\text{DNN}}(\phi)=\frac 1N \sum_{1\leq n\leq N}  \Vert S_{\phi}(C^n) - \theta^n \Vert ^2_ 2
\end{equation*}
\State output $S_{\phi}(.)$ s.t. $S_{\phi}(C) \approx \mathbb{E}\left[\theta|C\right]$
\State \textbf{Stage 2: run ABC with the learned summary statistic} $S_{\phi}$ \textbf{and the criterion} $ \Vert S_{\phi}(C)- S_{\phi}(C_\text{ref}) \Vert_2  \leq \epsilon$
\State \textbf{return} accepted samples $\theta^n \sim p(.  \mid \Vert S_{\phi}(C^n)-S_{\phi}(C_\text{ref}) \Vert_2  \leq \epsilon)$
\end{algorithmic}
\end{algorithm}
For $S_\phi$ informative enough and when $\epsilon \to 0$, as shown in \cite{summstatabc},
\begin{equation*}
    p(\theta  \mid \Vert S_{\phi}(C)-S_{\phi}(C_\text{ref}) \Vert  \leq \epsilon) \to p(\theta | S_{\phi}(C_\text{ref})) \approx p(\theta | C_\text{ref}).
\end{equation*}

\newpage

\section{SNPE -- \NPETransf or \NPECNN \label{sec:snpe_appendix}}
SMC-ABC yields only {samples from the target posterior distribution} $p(\theta|C_\text{ref})$, but evaluating log-probabilities can be useful for downstream tasks. In contrast, Neural Posterior Estimation (NPE) trains a conditional normalizing flow $p_{\psi}(\theta | C)$ \cite{snpea, snpec} to estimate the posterior distribution. It is then easy to sample from the posterior and return the values of its log-probabilities.
\\To ensure that $p_{\psi}(\theta| C_\text{ref})$ is close to $p(\theta |C_\text{ref})$, we minimize the Kullback--Leibler divergence ($D_\mathrm{KL}$) between both densities, averaged over the observations $C$ as per $\mathbb{E}_C \big[ D_\mathrm{KL}\big( p(\cdot | C) \Vert p_{\psi}(\cdot|C) \big) \big]$.
After simplifications and using a Monte Carlo estimator, the flow is trained to minimize, with $(\theta^n, C^n) \sim p(\theta)p(C|\theta)$:%with the following loss to be minimized in $\psi$ 
\begin{align*}
    \widehat{\mathcal{L}}_{\text{NPE}}(\psi)&=-\frac 1 N \sum_n \log(p_{\psi}(\theta^n|C^n)).
\end{align*}
Once trained, we obtain an amortized estimator of the posterior densities $p(\theta | C)$ valid for any $C$. We just have to plug in $C_\text{ref}$ to get the estimated posterior density $p_{\psi}(.|C_\text{ref})$ (see Algorithm~\ref{snpe}).\\
Since we are actually interested in the posterior at $C_\text{ref}$,  
parameters $\theta$ with very low posterior density may not be useful for learning $\psi$. Thus, we consider a sequential approach of NPE (SNPE)  with several rounds to get an iterative refinement of the posterior estimate~\cite{snpec}. From the second round, $\theta^n$ are sampled from the latest estimated posterior found instead of the prior. This way, training samples are more informative about $C_\text{ref}$, gradually improving the learning of $\psi$.\\
As the observations $C$ are high-dimensional (e.g. 2D-matrices), we encode them in a summary statistic $S_\phi$ before providing them to the normalizing flow $p_{\psi}$. This way, we actually learn  $p_\psi(.|S_\phi(C_\text{ref}))$ that should be close to $p_\psi(.|C_\text{ref})$ if $S_\phi$ is a \textit{sufficient} summary statistic. When we use the (frozen) pre-trained \textit{BlockFormer}, the method is referred as \NPETransf. For comparison, we also use the CNN-based architecture in the method referred as \NPECNN. 

\begin{algorithm}[H]
\caption{SNPE inspired from \cite{snpea} and \cite{snpec} \label{snpe}}
\begin{algorithmic}
\State \textbf{Input}:  $T$ rounds, posterior density estimator $p_{\psi}$, simulator, prior $p$, simulation budget $N$, observation $C_\text{ref}$, pre-learned summary statistic $S_\phi$
\State \textbf{Return}:  The estimated posterior density $p_\psi(.|S_\phi(C_\text{ref}))$\\ 
\For {round $t=1,...,T$}
\If{ $t=1$}
$p_t = p$
\EndIf
\For {$n=1,...,N$}
\State sample $\theta^n \sim p_t$
\State sample $C^n \sim p(.|\theta^n)$
\EndFor
\State train the posterior estimator $p_{\psi}$ on $\mathcal{D} = \{(\theta^n, C^n)\}_n$ with the loss to minimize in $\psi$ 
\begin{equation*}
    \widehat{\mathcal{L}}_{\text{NPE}}(\psi)=-\frac 1 N \sum_{1 \leq n \leq N} \log p_{\psi}(\theta^n|S_\phi(C^n))
\end{equation*}
\State use $p_{\psi}$ to construct the estimated posterior: $p_{\psi}(.|S_\phi(C_\text{ref}))$.
\State define the proposal for the next round: $p_t(\theta)  = p_{\psi}(\theta|S_\phi(C_\text{ref}))$
\EndFor
\State \textbf{return} samples $\theta^n \sim p_\psi(\theta|S_\phi(C_\text{ref}))$
\end{algorithmic}
\end{algorithm}

\newpage 

\section{Application: centromeres inference for \textit{Saccharomyces cerevisiae} -- posterior estimation. \label{sec:sc_post}}
\textit{S. cerevisiae} has a genome of $16$ chromosomes, so we look for the centromeres $\theta = (\theta_1,..., \theta_{16})$. To reduce the dimension of the problem, we carry $16$ parallel inferences: one per dimension of $\theta$. Thus, we have $16$ 1D inference problems where the parameter $\theta_i$ is drawn from a Uniform prior whose range is the size of the chromosome $i$ in bp. The simulator creates the $i^\mathrm{th}$ row of trans-blocks of a contact map $C$ (denoted $C_i)$. All the inference methods target the posterior $p(\theta_i | C_{\text{ref},i})$. The summary statistics $S_\phi$ that project each row of trans-contact blocks $C_i$ to $\theta_i$ is \textit{BlockFormer}. To construct the parameter estimate using \textit{BlockFormer}, we randomly select $10$ sets of $2$ trans-blocks from $C_i$ and pass those $10$ maps to the transformer. We obtain $10$ candidates and the final estimation is the mean over these $10$ candidates.\\
We compare with the case where $S_\phi$ is a CNN that captures the information of $C_i$ followed by an MLP to project this information into $\theta_i$. On the one hand, as the rows of trans-contact blocks $C_i$ are quite similar, we choose a shared architecture for the CNN between chromosomes. On the other hand, each MLP depends on the size of each chromosome so a chromosome-specific architecture is thus needed for this part of the network.\\
The summary statistic is used in both methods: ABC and NPE. When \textit{BlockFormer} is used, the methods are referred as \ABCTransf and \NPETransf, when the CNN is used, the methods are referred as \ABCCNN and \NPECNN.\\
For the NPE methods, since we consider sequential methods that are specific to one observation $C_\text{ref, i}$, we need to learn one normalizing flow $p_\psi$ per parameter. However, we choose the same density estimator for all inferences: a Masked Autoregressive Flow (MAF) as well as the sequential approach SNPE-C \cite{snpec} for the experiments. The MAF consist of 11 sequential transforms: 1 PointwiseAffineTransform followed by 5 blocks of [MAF + RandomPermutation], where each MAF is parameterized by a MADE network with 2 feedforward blocks (MaskedLinear 50→50) and a final MaskedLinear 50→2.\\
As the posterior ground truth $p^*(\theta | C_{\text{ref}})$ is unknown, we choose to model it via independent Gaussian distributions. We suppose conditional independence as explained in Section~\ref{sec:sbi_presentation} such that: \begin{equation*}
    p^*(\theta | C_{\text{ref}}) \approx \prod_{i=1}^{16} p(\theta_i | C_{\text{ref}, i}) = \prod_{i=1}^{16} \mathcal{N}(\theta_i;  \theta_\text{ref, i}, \sigma_\text{ref}^2)
\end{equation*}
where $\sigma_\text{ref} = 50$ based on the fact that centromeres in yeasts span genomic regions of roughly $100$ bp centered around $\theta_\text{ref}$ \cite{centrosize}.\\
The choice of this posterior is further motivated by the result in \cite{summstatabc}, which states that if the posterior distribution belongs to the exponential family (e.g. Gaussian density), then the summary statistic $S_\phi = \mathbb{E}\left[\theta|C\right]$ is a \textit{sufficient} summary statistic.
\begin{figure}[H]
    \centering
    \includegraphics[width=0.8\linewidth]{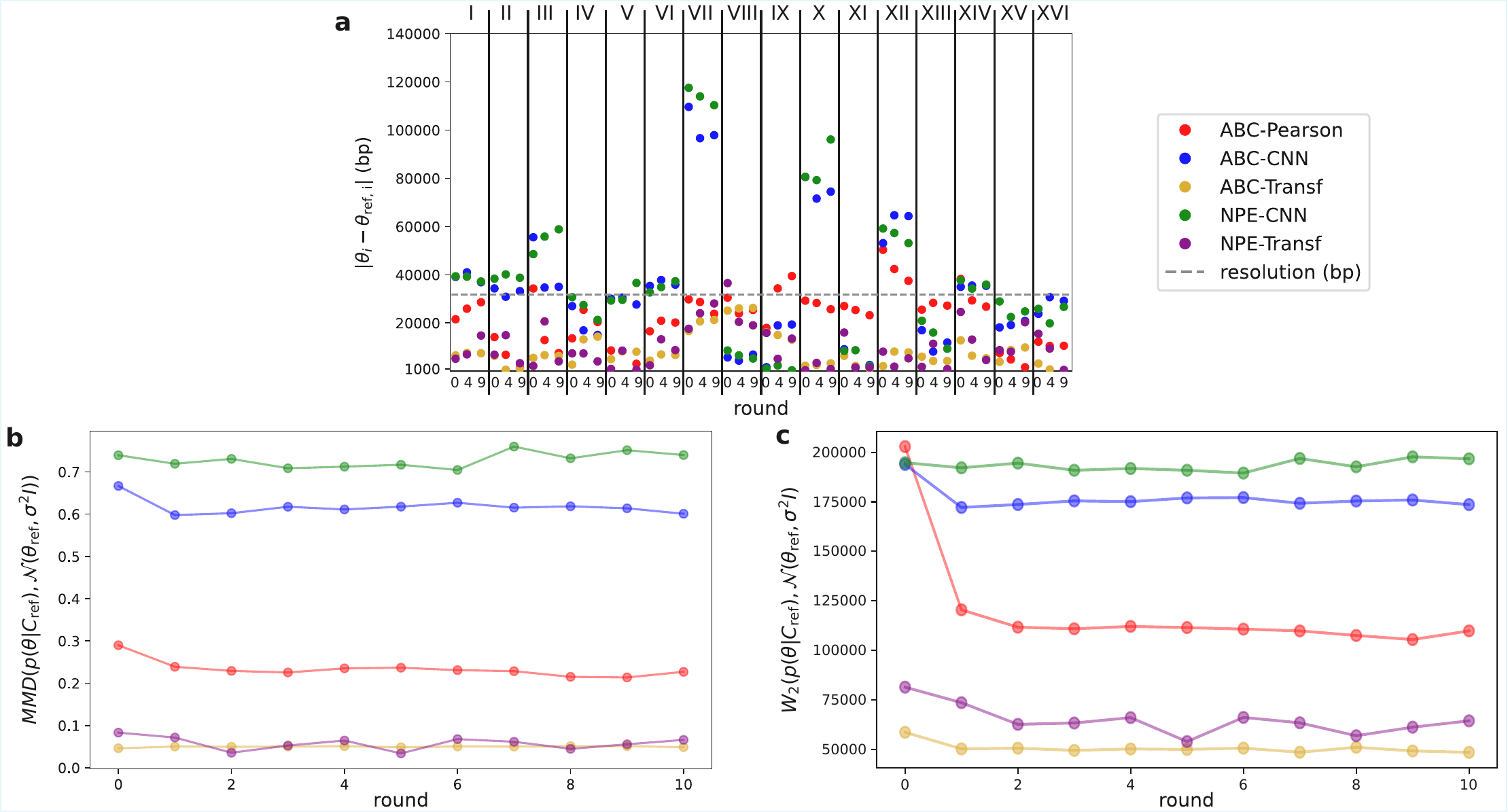}
    \caption{\label{fig:metrics_16_chr}\small We report the absolute error per dimension of $\theta$ between the mean $\theta$ computed over the $5\%$ best $\theta$ according to the ABC criterion or sampled from the flow and $\theta_\text{ref}$ (\textbf{a})  as well as the MMD (\textbf{b}) and the Wasserstein-2 distance (\textbf{c}) between $p(\theta|C_\text{ref})$ and $\delta_{\theta_\text{ref}}$. The horizontal dotted line stands for the resolution of the contact map $C_\text{ref}$ (in bp) in the top figure.}
    \label{fig:abc_sbi_metrics}
\end{figure}

\newpage

As the method \NPETransf outperforms all the others and enables access to the density, we further provide calibration diagnostics. We use the normalizing flow obtained in the final inference round to evaluate whether the predicted posterior distributions are well calibrated. Specifically, we report simulation-based calibration (SBC) ranks, which check if the true parameters are uniformly distributed within posterior samples, and the expected coverage at level $\alpha = 0.9$, which measures whether the nominal credible intervals match their empirical frequency.
\begin{figure}[H]
\centering
\begin{subfigure}{\textwidth}
    \centering
    \includegraphics[width=\textwidth]{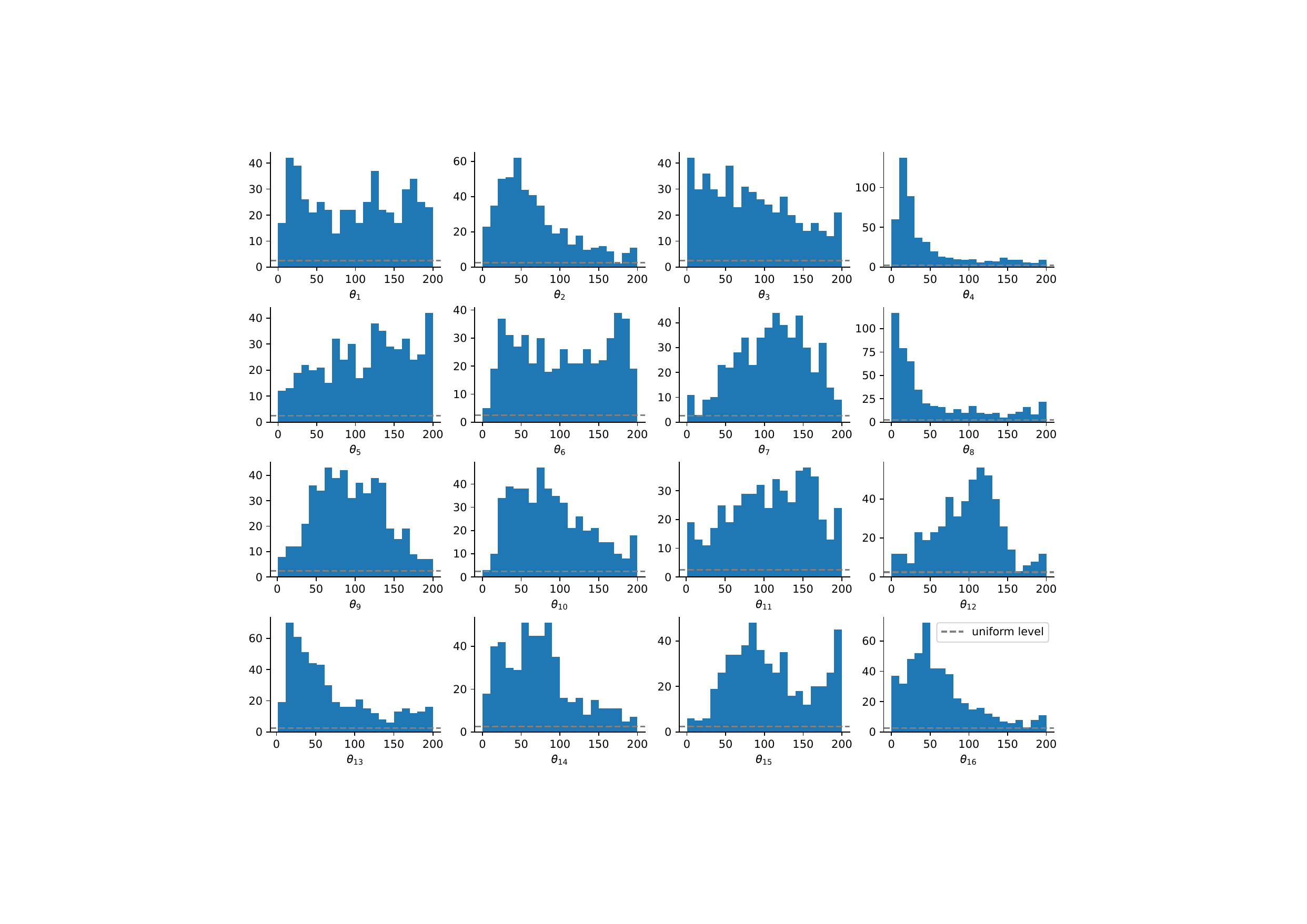}
\end{subfigure}

\vspace{0.1cm}

\begin{subfigure}{0.5\textwidth}
    \centering
    \includegraphics[width=\textwidth]{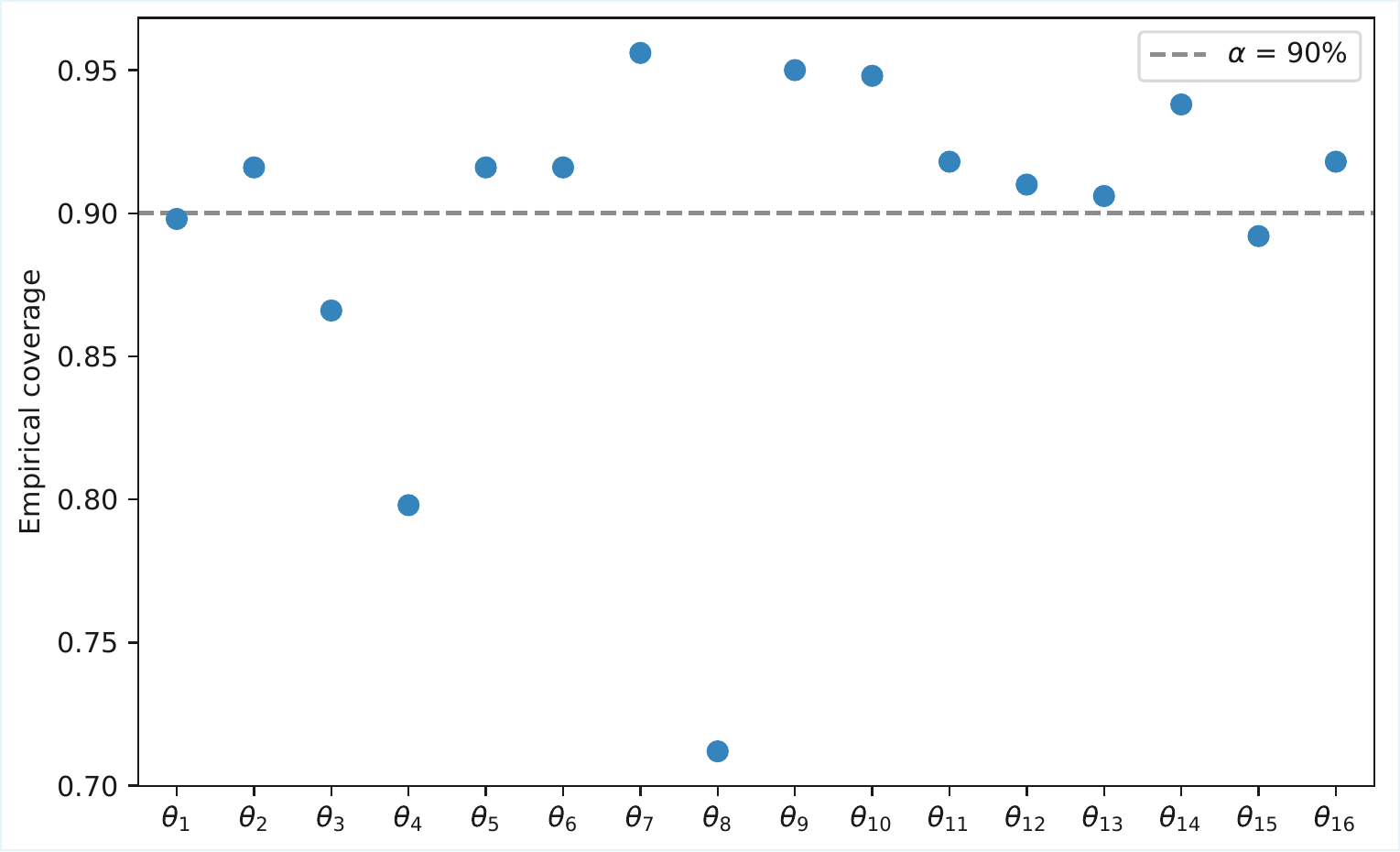}
\end{subfigure}

\caption{\small (top) Simulation-Based Calibration rank. (bottom) Expected coverage of level $\alpha = 90 \%$.}
\end{figure}
The posteriors are globally well calibrated: in many dimensions, the histograms are roughly flat and the expected coverage are close to the nominal level (e.g. for $\theta_1$, $\theta_6$ or $\theta_{11}$). However, some dimensions exhibit U-shape histograms as well as coverage under the nominal level (e.g. for $\theta_{4}$ or $\theta_8$) indicating overconfident posteriors where the true value tend to fall in the tails. Conversely, other dimensions show inverted U-shape histograms as well as coverage over the nominal level (e.g. for $\theta_7$ or $\theta_9$) reflecting overly diffuse posteriors and conservative uncertainty estimates.
Overall, these diagnostics suggest good global calibration, with noticeable variability across parameters, likely due to differences in identifiability across chromosomes.

\end{document}